\documentclass[journal]{new-aiaa}
\usepackage[utf8]{inputenc}
\usepackage{textcomp}
\usepackage{multirow}
\usepackage{graphicx}
\usepackage{amsmath}
\usepackage{color}
\usepackage{caption}
\usepackage{longtable,tabularx}
\usepackage{subcaption}
\usepackage{adjustbox}

\graphicspath{{figures/}}

\title{Overview of Gaussian process based multi-fidelity techniques with variable relationship between fidelities}

\author{Lo\"{i}c Brevault \footnote{ONERA, DTIS, Université Paris Saclay, F-91123 Palaiseau Cedex, France, loic.brevault@onera.fr}, Mathieu Balesdent \footnote{ONERA, DTIS, Université Paris Saclay, F-91123 Palaiseau Cedex, France, mathieu.balesdent@onera.fr}}
\affil{ONERA, DTIS, Université Paris Saclay, F-91123 Palaiseau Cedex, France}
\author{Ali Hebbal \footnote{ONERA, DTIS, Université Paris Saclay, F-91123 Palaiseau Cedex, France, Université de Lille, CNRS/CRIStAL, INRIA Lille, ali.hebbal@onera.fr}}
\affil{ONERA, DTIS, Université Paris Saclay, F-91123 Palaiseau Cedex, France\\ Université de Lille, CNRS/CRIStAL, INRIA Lille, France}

\begin{document}

\maketitle

\begin{abstract}
The design process of complex systems such as new configurations of aircraft or launch vehicles is usually decomposed in different phases which are characterized for instance by the depth of the analyses in terms of number of design variables and fidelity of the physical models. At each phase, the designers have to compose with accurate but computationally intensive models as well as cheap but inaccurate models. Multi-fidelity modeling is a way to merge different fidelity models to provide engineers with accurate results with a limited computational cost. Within the context of multi-fidelity modeling, approaches relying on Gaussian Processes emerge as popular techniques to fuse information between the different fidelity models. The relationship between the fidelity models is a key aspect in multi-fidelity modeling. This paper provides an overview of Gaussian process-based multi-fidelity modeling techniques for variable relationship between the fidelity models (e.g., linearity, non-linearity, variable correlation). Each technique is described within a unified framework and the links between the different techniques are highlighted. All the approaches are numerically compared on a series of analytical test cases and four aerospace related engineering problems in order to assess their benefits and disadvantages with respect to the problem characteristics.
\end{abstract}

\section{Nomenclature}

{\renewcommand\arraystretch{1.0}
\noindent\begin{longtable*}{@{}l @{\quad=\quad} l@{}}
$\mathbf{x}$  & input variable vector \\
$y$  & output scalar variable \\
$f(\cdot)$ & unknown mapping function \\
$f_t(\cdot)$ & unknown mapping function associated to fidelity $t$ \\
$\mathbf{f}(\cdot)$ & vector-valued unknown mapping function \\
$\boldsymbol{\Theta}$  & hyper-parameters \\
$k^{\boldsymbol{\Theta}}(\cdot,\cdot)$  & covariance function parameterized by hyper-parameters $\boldsymbol{\Theta}$ \\
$\mathcal{N}(\cdot,\cdot)$  & Gaussian distribution \\
$\mathbf{K}_{MM}$  & matrix of size $M \times M$\\
$\mathcal{X}_{M}$  & input data set of size $M$\\
$\mathcal{Y}_{M}$  & output data set of size $M$\\
\end{longtable*}}

\section{Introduction}
\lettrine{T}{}he design of aerospace systems such as aircraft, launch vehicles, missiles, \textit{etc.} is usually decomposed in different phases, from early-design to detailed design and manufacturing. These phases are in particular characterized by the depth of the analyses in terms of number of design variables and complexity of the phenomena to be modeled. At a given phase, the designer often has to compose with a series of physical approaches to model the phenomena or discipline of interest (structure, aerodynamics, trajectory, \textit{etc.}). These different models can be characterized by their accuracy and their computational cost. Generally, the more precise, the more computationally intensive. In the early design phase, computationally efficient (but imprecise) models are often used in order to explore a large design space with repeated calls to the models. In the detailed design phase, high-fidelity models are employed to capture complex physical phenomena and to refine the prediction uncertainty but with an intensive computational cost. Multi-fidelity methods \citep{fernandez2016review} are a way to combine the responses to different levels of fidelity in order to control the model uncertainties and to speed up the design analysis. The subject of multi-fidelity for design analysis and optimization is already profuse in contributions. Two surveys of multi-fidelity methods have been performed by Fernandez-Godino \textit{et al.} \citep{fernandez2016review} and Peherstorfer \textit{et al.} \citep{peherstorfer2018survey}. The analysis of complex systems such as uncertainty propagation, sensitivity analysis or optimization requires repeated model evaluations at different locations in the design space which typically cannot be afforded with high-fidelity models.  Multi-fidelity
methodologies perform model management that is to say, they balance the
fidelity levels to mitigate cost and ensure the accuracy in analysis. In
the review of Peherstorfer \textit{et al.} \citep{peherstorfer2018survey}, the authors classify the
multi-fidelity techniques in three categories: adaptation, fusion and
filtering. The adaptation category encompasses the methods that enhance
the low-fidelity models with results from the high-fidelity ones while
the computation proceeds. An example is given by the model correction
approach \citep{kennedy2000predicting} where an autoregressive
process is used to reflect the hierarchy between the accuracy of the
various outputs. The fusion techniques aim to build models by combining
low and high-fidelity model outputs. Two examples of fusion techniques
are the co-kriging \citep{myers1982matrix,cressie1992statistics,perdikaris2015multi} and the multilevel stochastic collocation \citep{teckentrup2015multilevel}. Finally, the filtering consists in calling low-fidelity models to decide when to use
high-fidelity models (\textit{e.g.}, multi-stage sampling).

The early multi-fidelity optimization techniques were developed to
alternate between the computationally efficient simplified models
(typically metamodels, also known as surrogate models) and the more accurate
and costly ones. Although, it would be preferred to optimize the simulator
with high accuracy, low-fidelity experiments can help ruling out some
uninteresting regions of the input space (or on the contrary help
finding interesting ones) while preserving the computational budget. A
popular example of such an alternation between low and high-fidelity models
can be found in \citep{jones1998efficient}. The use of multi-fidelity metamodels could help to decide both which input parameter and which model fidelity should be chosen within the remaining computational budget \citep{huang2006sequential}.

Then, another philosophy emerged. Instead of replacing high-fidelity
models by low-fidelity models in sequential phases, new techniques have
been proposed to synthesize all information of various fidelities by
weighting them. Bayesian statistics and in particular co-kriging are popular approaches to merge models. Multi-fidelity Bayesian optimization methods have been explored in several articles \citep{forrester2007multi,keane2012cokriging,sacher2018methodes}.

Within the context of multi-fidelity modeling, in engineering applications, approaches relying on Gaussian Processes (GP) emerge as popular techniques to fuse information between the different fidelity models \cite{liu2018remarks}. In particular, the classical Auto-Regressive (AR1) method is often employed in the aerospace design problems \cite{laurenceau2008building,kuya2011multifidelity,toal2011efficient,keane2012cokriging,toal2014multifidelity,fernandez2016review,bailly2019multifidelity}. This method defines a linear autoregressive information fusion scheme introduced by Kennedy and O'Hagan \cite{kennedy2000predicting}. It is a popular technique due to the ease of its implementation. In addition to its prediction capability, it offers an uncertainty model of the prediction that may be used for optimization purposes or surrogate model refinement. Recently, more advanced non-linear fusion schemes relying on GP have been introduced \cite{perdikaris2017nonlinear,cutajar2019deep} within the Machine Learning community (ML) to go beyond linear fusion approaches and to improve the prediction capabilities. Within the context of engineering design, these approaches can offer interesting alternatives to the standard AR1 or co-kriging methods and enable to catch non-linear relationships between the fidelity models. In some aerospace design fields, non-linear relations between the fidelity models are involved and AR1 or co-kriging may be limited to appropriately fuse information.

In this paper, it is proposed to explore advanced ML techniques (in particular Non-linear Auto-Regressive Gaussian Process \cite{perdikaris2017nonlinear} and Multi-fidelity Deep Gaussian Process \cite{cutajar2019deep}) for multi-fidelity modeling and evaluate them on a benchmark of analytical and aerospace design problems, and compare them to linear mapping between the fidelities including the classical AR1 fusion scheme. The aim of the paper is to highlight the importance of the selection of adequate multi-fidelity modeling techniques for the considered design problem to offer efficient surrogate model capabilities. A complementary important point for multi-fidelity modeling based on GP is to evaluate the accuracy of its prediction capabilities but also of its uncertainty prediction model. Indeed, GP-based techniques offer an uncertainty model for the prediction that is often used in optimization \cite{jones1998efficient,queipo2005surrogate} or uncertainty propagation \cite{girard2004approximate,lockwood2012gradient} to refine the surrogate models. Therefore, an accurate uncertainty prediction model is important within these contexts and therefore, will be assessed through the benchmark problems. 

The rest of the paper is organized as follows. In Section \ref{Section_GP}, a brief overview of Gaussian Process is carried out to introduce essential concepts and notations for multi-fidelity techniques based on GP. Section \ref{Section_MF_GP_methods} presents four GP-based multi-fidelity modeling alternative approaches. Two methods are based on a linear mapping between the fidelities (co-kriging with linear model of coregionalization and auto-regressive AR1), and two others involve a non-linear relationship between the fidelity models (non-linear auto-regressive multi-fidelity Gaussian process and multi-fidelity deep Gaussian process). The main principles for each method are presented with unified notations and links between the models are outlined. In Section \ref{Section_analytic}, analytical and aerospace benchmark problems are presented to compare these multi-fidelity techniques for different dimension problems and different fidelity mapping types. For the aerospace test cases, four different design problems are carried out: a cantilever beam problem, a launch vehicle trajectory problem, a supersonic business jet multidisciplinary test case and an aerostructural wing design problem. The prediction accuracy and the uncertainty model associated to the prediction are assessed for the multi-fidelity techniques for these benchmark problems.

\section{Gaussian Process (GP)} \label{Section_GP}

Gaussian Process (often referred as Kriging) \cite{matheron1963principles,sasena2002flexibility} is a statistical surrogate model that may be used to approximate any unknown mapping function $f(\cdot)$ on its input space $\mathbb{R}^d$ by considering $f(\cdot)$ as a realization of a Gaussian Process. A Gaussian Process (GP) describes a distribution over a set of functions. It corresponds to a collection of infinite random variables, any finite number of which has a joint Gaussian distribution. A GP is characterized by its mean and covariance functions. 
GP consists of a supervised learning problem, it is trained from a set of samples (Design of Experiment - DoE or dataset) of size $M$, $\mathcal{X}_M=\left\{ \mathbf{x}^{1}, \ldots , \mathbf{x}^{M} \right\}$ being the input data set $\left( \mathbf{x}\in \mathbb{R}^d \right)$ and the corresponding unknown function responses is noted $\mathcal{Y}^M= \left\{y^{1}=f\left(\mathbf{x}^{1}\right),\ldots, \right.$ $\left. y^{M}=f\left(\mathbf{x}^{M}\right) \right\}$. Then, this surrogate model may be used to predict the exact function response $f(\cdot)$ at a new unmapped location without evaluating it. The advantage of this approach is in terms of computational evaluation cost. Indeed, instead of evaluating the expensive black-box function $f(\cdot)$, the surrogate model is used which is much cheaper to evaluate.

In the GP regression, a GP prior is placed on the unobserved function $f(\cdot)$ using a prior covariance function $k^{\boldsymbol{\Theta}}\left(\mathbf{x},\mathbf{x}' \right)$ that depends on hyper-parameters $\boldsymbol{\Theta}$ and a mean function $m(\cdot)$. Example of usual kernel (known as p-exponential) is provided in Eq. \eqref{eq:kernel},
\begin{equation}
k^\Theta(\mathbf{x},\mathbf{x}')=\Theta_\sigma\exp\left(-\sum_{i=0}^{d}
\Theta_{\theta_i}|\mathbf{x}^{(i)}-\mathbf{x}'^{(i)}|^{\Theta_{p_i}}\right).
\label{eq:kernel}
\end{equation}

As the trend of the response is \textit{a priori} unknown a constant mean function $\mu$ is often assumed (then GP is named ordinary Kriging).
 Therefore, the GP may be written $\hat{f}(\mathbf{x}') \sim \mathcal{N}\left(m,k^{\boldsymbol{\Theta}}\left(\mathbf{x},\mathbf{x}' \right) \right)$ and has a multivariate distribution on any finite subset of variables, in particular on the DoE $\mathcal{X}^M$ (noted $\mathbf{f}^M$), $\mathbf{f}^M|\mathcal{X}^M \sim \mathcal{N}\left( \mathbf{1}\mu,\mathbf{K}^{\boldsymbol{\Theta}}_{MM} \right)$ where $\mathbf{K}^{\boldsymbol{\Theta}}_{MM}$ is the covariance matrix constructed from the parameterized covariance function $k^{\boldsymbol{\Theta}}(\cdot)$ on $\mathcal{X}^M$ (in the rest the dependence on $\boldsymbol{\Theta}$ is dropped for the sake of notation simplicity). The choice of the covariance function determines the prior assumptions of the function to be modeled. A Gaussian noise variance (also called nugget effect) can be considered, such that the relationship between the latent function values $f\left(\mathcal{X}^M\right)$  and the observed responses $\mathcal{Y}^M$ is given by: $ p\left(\mathbf{y}| \mathbf{f}^M \right) = \mathcal{N}\left(\mathbf{y}|\mathbf{f}^M, \sigma^2 \mathbf{I} \right)$. This Gaussian noise is necessary when the discipline function $f(\cdot)$ is non deterministic (or in case of noisy observations).
The marginal likelihood is obtained by integrating out the latent function $f(\cdot)$:
\begin{equation}
p \left(\mathbf{y}|\mathcal{X}^M, \boldsymbol{\Theta} \right) = \mathcal{N}\left(\mathbf{y}| m, \mathbf{K}_{MM} + \sigma^2 \mathbf{I} \right)
\end{equation}

\begin{figure}
\begin{center}
\includegraphics[width=0.9\linewidth]{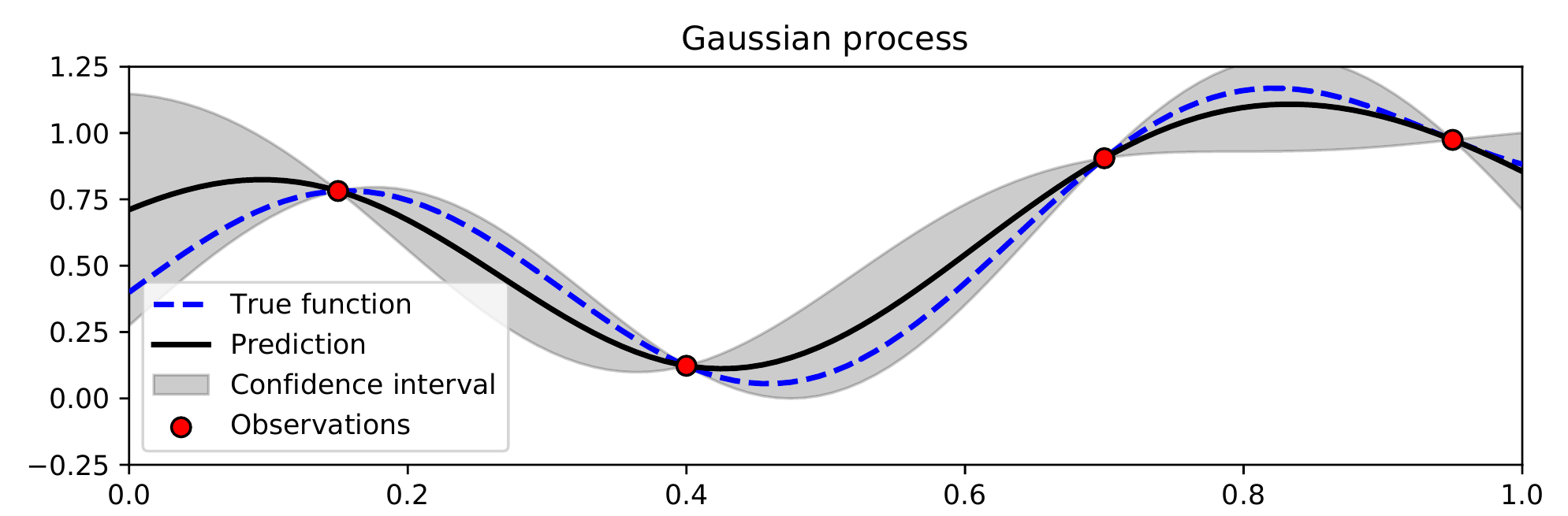}
\caption{Example of Gaussian process prediction and associated confidence interval}
\label{Kriging_example}
\end{center}
\end{figure}

In the following, we set $\mathbf{\hat{K}}_{MM} = \mathbf{K}_{MM}+\sigma^2 \mathbf{I}$. In order to train the GP, it is possible to maximize the log marginal likelihood to find the optimal values of the hyper-parameters $\boldsymbol{\Theta}, m$ and  $\sigma$. 
All the kernel matrices implicitly depend on the hyperparameters $\boldsymbol{\Theta}$ and the negative log marginal likelihood (and its derivative) is given by:
\begin{eqnarray}
L \left(\boldsymbol{\Theta}|\mathcal{X}^M,\mathcal{Y}^M \right) &=& \log \left(p \left(\mathbf{y}|\mathcal{X}^M,\mathcal{Y}^M, \boldsymbol{\Theta} \right)\right) \propto  \log \left(|\mathbf{\hat{K}}_{MM}| \right) - \mathbf{y}^T\mathbf{\hat{K}}^{-1}_{MM} \mathbf{y} \\
\frac{\text{d}L}{\text{d}\boldsymbol{\Theta}} &=&  \mathbf{y}^T\mathbf{\hat{K}}^{-1}_{MM} \frac{\text{d}\mathbf{\hat{K}}_{MM}}{\text{d}\boldsymbol{\Theta}}\mathbf{\hat{K}}^{-1}_{MM}\mathbf{y} + \text{Tr}\left(\mathbf{\hat{K}}^{-1}_{MM}\frac{\text{d}\mathbf{\hat{K}}_{MM}}{\text{d}\boldsymbol{\Theta}}  \right)
\end{eqnarray}

Gradient-based optimizers may be used to minimize the negative log marginal likelihood in order to determine the hyperparameter values of the trained GP. After the training phase, the prediction at a new point $\mathbf{x}^*\in \mathbb{R}^d$ is made by using the conditional properties of a multivariate normal distribution (Figure \ref{Kriging_example}):
\begin{equation}
p\left( y^*|\mathbf{x}^*,\mathcal{X}^M,\mathcal{Y}^M,\boldsymbol{\Theta} \right) = \mathcal{N}\left(y^* | \hat{y}^*,\hat{s}^{*2} \right)
\end{equation}
with $\hat{y}^*,\hat{s}^{*2}$ the mean prediction and the associated variance given by:
\begin{eqnarray}
\hat{y}^* &=& m + \mathbf{k}^T_{\mathbf{x}^*}\left({\mathbf{K}}_{MM} + \sigma^2 \mathbf{I} \right)^{-1} \left(\mathbf{y} - \mathbf{1}\mu \right) \\
\hat{s}^{*2}&=& k_{\mathbf{x}^*,\mathbf{x}^*} - \mathbf{k}^T_{\mathbf{x}^*}\left({\mathbf{K}_{MM}} + \sigma^2 \mathbf{I} \right)^{-1}\mathbf{k}_{\mathbf{x}^*}
\end{eqnarray}

\noindent where $k_{\mathbf{x}^*,\mathbf{x}^*} = k(\mathbf{x}^*,\mathbf{x}^*)$ and $\mathbf{k}_{\mathbf{x}^*} = \left[k \left( \mathbf{x}_{(i)}, \mathbf{x}^* \right)\right]_{i=1,\ldots,M}$

In the training phase of GP, the operations that dominate in terms of time complexity are the linear solve $\mathbf{\hat{K}}^{-1}_{MM} \mathbf{y}$, the log determinant $\log \left(|\mathbf{\hat{K}}_{MM}| \right)$ and the trace term $\text{Tr}\left(\mathbf{\hat{K}}^{-1}_{MM}\frac{\text{d}\mathbf{\hat{K}}_{MM}}{\text{d}\boldsymbol{\Theta}}  \right)$. In GP, these quantities are often computed with the Cholesky decomposition of $\mathbf{\hat{K}}_{MM}$ which is computationally intensive and involves $\mathcal{O}\left(M^3 \right)$ operations. In order to reduce the computational cost associated to GP when the set of data is large (due for instance to a high-dimensional space that needs to be covered), sparse GP have been developed \citep{titsias2009variational}.

An example of a 1D function and a GP built based on four observations (samples) is represented in Figure \ref{Kriging_example}. The confidence interval is based on the associated GP variance $\left(\pm 3\hat{s}^{*2} \right)$. The variance is null at the observation locations when no nugget effect is considered. It also increases as the distance from an existing data sample increases. GP is very interesting in the complex system design context because it provides the designer with both the prediction of the model and its estimated uncertainty. 

\section{Multi-fidelity modeling approaches based on Gaussian Processes}\label{Section_MF_GP_methods}
Different approaches based on Gaussian Processes (GP) exist to fuse information provided by models with different fidelities depending on the complexity of the relationships between these fidelities. In engineering design field, linear models such that co-kriging \cite{cressie1992statistics,boyle2005dependent,fricker2013multivariate} based on Linear Model of Coregionalization (LMC) or classical Auto-Regressive (AR1) are implemented most of the time. These approaches are presented in Sections \ref{LMC_section} and \ref{AR1_theory}. Then, more advanced techniques, developed in the machine learning field, are overviewed to account for more complex dependencies between the available fidelities.

\subsection{Co-kriging with Linear Model of Coregionalization (LMC)}\label{LMC_section}

A general linear model may be considered for multi-fidelity modeling using GP considering $s$ fidelity models. It is based on the idea that instead of considering a set of $s$ independent scalar GPs, one may consider a single multi-output GP (also called vector-valued GP) of dimension $s$. Within the context of multi-fidelity, each component of the vector-valued GP corresponds to a fidelity. It is referred to as co-kriging approach.

Let us consider for a fidelity model $t$, observation data (resulting from the model simulation on a Design of Experiment (DoE) for instance) in the input space $\mathcal{X}_t \in \mathbb{R}^d$ and the corresponding model output $\mathcal{Y}_t(\mathcal{X}_t) \in \mathbb{R}$. For $s$ different fidelity models, it is possible to organize the observation data set by increasing fidelities $\mathcal{D}_t = \{\mathcal{X}_t,\mathcal{Y}_t  \}$ for $t=1,\ldots, s$, it may also be written $\mathcal{D} = \{(\mathcal{X}_1,\mathcal{Y}_1),(\mathcal{X}_2,\mathcal{Y}_2),\ldots, (\mathcal{X}_s,\mathcal{Y}_s)  \}$. Therefore, $\mathcal{Y}_1$ corresponds to the cheapest and lowest fidelity model output dataset whereas $\mathcal{Y}_s$ is the output dataset of the most computationally intensive model with the highest fidelity.

\begin{figure}[!h]
\begin{center}
\includegraphics[width=0.3\linewidth]{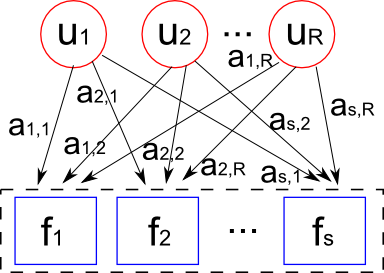}
\caption{Linear Model of Coregionalization schematic view}
\label{LMC_example}
\end{center}
\end{figure}

The underlying idea of co-kriging based on LMC \cite{goovaerts1997geostatistics,alvarez2012kernels} (referred in the following as LMC) is to exploit the existing correlations between the various outputs (here the fidelity models) in order to improve the modeling accuracy with respect to the separate and independent modeling of each output. In the LMC, the outputs are expressed as linear combinations of independent latent functions (Figure \ref{LMC_example}).

The key question for multi-output GP is the definition of the kernel function, the problem resumes to learn an unknown vector-valued function $\mathbf{f}(\cdot)=\left[f_1(\cdot), \cdots, f_s(\cdot)\right]$ between an input space defined on $\mathbb{R}^d$ and an output space defined on $\mathbb{R}^s$.
In this approach, the function $\mathbf{f}(\cdot)$ is assumed to follow a Gaussian Process:
\begin{equation}
\mathbf{f} \sim \mathcal{GP}(\mathbf{m},\mathbf{K})
\end{equation}
with $\mathbf{m} \in \mathbb{R}^s$ a vector corresponding to the mean function of each component $m_t(\mathbf{x})$ for $t=1,\cdots, s$ and $\mathbf{K}$ a matrix valued function defined as follows:
\begin{equation}
\mathbf{K}(\mathcal{X},\mathcal{X}) = \begin{bmatrix}
    K_{1,1}(\mathcal{X}_1,\mathcal{X}_1)        & \dots & K_{1,s}(\mathcal{X}_1,\mathcal{X}_s) \\
    K_{2,1}(\mathcal{X}_2,\mathcal{X}_1)        & \dots & K_{2,s}(\mathcal{X}_2,\mathcal{X}_s) \\
    \hdotsfor{3} \\
    K_{s,1}(\mathcal{X}_s,\mathcal{X}_1)       & \dots & K_{s,s}(\mathcal{X}_s,\mathcal{X}_s)
\end{bmatrix}
\end{equation}
$\mathbf{K}(\mathcal{X},\mathcal{X})$ is a $(M_1+M_2+\cdots+M_s) \times (M_1+M_2+\cdots+M_s)$ matrix with $M_i$ the dataset size for the fidelity $i$ and $s$ the total number of fidelity models and $K_{1,1}(\mathcal{X}_1,\mathcal{X}_1)$ is a $M_1 \times M_1$ matrix.
Therefore, in the covariance matrix $\mathbf{K}(\cdot, \cdot)$, the submatrix $K_{t,t'}(\cdot, \cdot)$ corresponds to the covariance between the outputs of $f_t(\cdot)$ and $f_{t'}(\cdot)$, meaning between fidelity models $t$ and $t'$, representing the degree of correlation between them.
Different families of kernels may be used for vector-valued functions. The most commune approach consists in considering separable kernels. These kernels are formulated as a product between a kernel function for the input space, and a kernel function to account for the interactions
of the outputs. Indeed, each term of the matrix $\mathbf{K}(\cdot,\cdot)$ may be expressed as:
\begin{equation}
K_{t,t'}(\mathbf{x},\mathbf{x}') = k(\mathbf{x},\mathbf{x}')\times k_T(t,t')
\end{equation}
where $k(\cdot,\cdot)$ is a kernel on the input space defined on $\mathbb{R}^d$ and $k_T(\cdot,\cdot)$ is a kernel for the output space defined on $\mathbb{R}^s$.
In this approach, the contribution of the input and output variables are separated. It is then possible to rewrite the matrix $\mathbf{K}(\mathbf{x},\mathbf{x}')$ such that:
\begin{equation}
\mathbf{K}(\mathbf{x},\mathbf{x}') = k(\mathbf{x},\mathbf{x}')\times \mathbf{B}
\end{equation}
where $\mathbf{B}$ is called the matrix of coregionalization of size $s\times s$, it represent the relationship between the outputs, meaning the model fidelities. In case $\mathbf{B}=\mathbf{I}_s$ is the identity matrix, the $s$ outputs are considered as independent. 

It is even possible to generalize this approach to a sum of separable kernels such that:
\begin{equation}
\mathbf{K}(\mathbf{x},\mathbf{x}') = \sum_{r=1}^R k_r(\mathbf{x},\mathbf{x}')\times \mathbf{B}_r
\end{equation}
involving $R$ different kernels. To ensure the validity of the kernel definition for vector-valued function, LMC defines each output component $f_t(\cdot)$ such that:
\begin{equation}
f_t(\mathbf{x}) = \sum_{r=1}^R a_{t,r} \times u_r(\mathbf{x})
\label{LCM_output}
\end{equation}
with $u_r(\cdot)$ a latent function generated by a GP of mean zero and covariance matrix $Cov\left(u_r(\mathbf{x}),u_{r'}(\mathbf{x}') \right)=k_r(\mathbf{x},\mathbf{x}')$ if $r=r'$ and a scalar coefficient $a_{t,r}$. The latent functions $u_r(\cdot)$ for $r=1,\ldots,R$ are independent if $r \neq r'$. Moreover, $u_r(\mathbf{x})$ and $u_{r'}(\mathbf{x'})$ may share the same covariance function $k_r(\mathbf{x},\mathbf{x'})$ while being independent (for instance two different realizations of a single GP). It is therefore possible to rewrite Eq.(\ref{LCM_output}) by regrouping the latent functions that share the same covariance model:
\begin{equation}
f_t(\mathbf{x}) = \sum_{r=1}^R \sum_{i=1}^{C_r} a^i_{t,r} \times u^i_r(\mathbf{x})
\end{equation}

The output function is expressed as a sum of $R$ groups of independent latent functions $u^i_r(\cdot)$ of size $C_r$ and that these functions within each group share the same covariance model $k_r(\mathbf{x},\mathbf{x'})$.  
Therefore, due to the independence of the latent functions $u^i_r(\cdot)$ is is possible to express the covariance function between two outputs $Cov\left( f_t(\mathbf{x}, f_{t'}(\mathbf{x'})\right) = K_{t,t'}(\mathbf{x},\mathbf{x'})$ such that:
\begin{equation}
K_{t,t'}(\mathbf{x},\mathbf{x'}) = \sum_{r=1}^R \sum_{i=1}^{C_r} a^i_{t,r}a^i_{t',r} \times k_r(\mathbf{x},\mathbf{x'})= \sum_{r=1}^R b_{t,t'}^r \times k_r(\mathbf{x},\mathbf{x'})
\end{equation}
with $b_{t,t'}^r= \displaystyle\sum_{i=1}^{C_r} a^i_{t,r}a^i_{t',r} $.

Eventually, the kernel matrix $\mathbf{K}(\mathbf{x},\mathbf{x}')$ may be written:
\begin{equation}
\mathbf{K}(\mathbf{x},\mathbf{x}') = \sum_{r=1}^R \mathbf{B}_r k_r(\mathbf{x},\mathbf{x'})
\end{equation}
with $\mathbf{B}_r$ a coregionalization matrix and its components $b_{t,t'}^r$. The rank of the matrix $\mathbf{B}_r$ is defined by $C_r$ corresponding to the number of independent latent functions that share the same covariance function $k_r(\mathbf{x},\mathbf{x'})$.

Co-kriging with LMC is expressed as a sum of separable kernels decoupling the output treatment from the input variables. To illustrate, considering two fidelity models $f_1(\cdot)$ and $f_2(\cdot)$, two latent functions ($R=2$) with a rank $C_r=1$ for the coregionalization matrices, the output functions may be defined such that:
\begin{eqnarray}
f_1(\mathbf{x}) &=& a_{1,1}^1 u_1^1(\mathbf{x}) + a_{1,2}^1 u_2^1(\mathbf{x}) \\
f_2(\mathbf{x}) &=& a_{2,1}^1 u_1^1(\mathbf{x}) + a_{2,2}^1 u_2^1(\mathbf{x})
\end{eqnarray}
with $u_1^1(\cdot)$ and $u_2^1(\cdot)$ two latent functions sampled from two GPs with different covariance functions. In case a rank $C_r=2$ is considered for the coregionalization matrices (still with $R=2$), the outputs are defined by:
\begin{eqnarray}
f_1(\mathbf{x}) &=& a_{1,1}^1 u_1^1(\mathbf{x}) + a_{1,1}^2 u_1^2(\mathbf{x}) + a_{1,2}^1 u_2^1(\mathbf{x}) + a_{1,2}^2 u_2^2(\mathbf{x}) \\
f_2(\mathbf{x}) &=& a_{2,1}^1 u_1^1(\mathbf{x}) + a_{2,1}^2 u_1^2(\mathbf{x}) + a_{2,2}^1 u_2^1(\mathbf{x}) + a_{2,2}^2 u_2^2(\mathbf{x})
\end{eqnarray}
where $u_1^1(\cdot)$ and $u_1^2(\cdot)$ share the same covariance function whereas $u_2^1(\cdot)$ and $u_2^2(\cdot)$ share another covariance function.

A limitation of co-kriging with LMC for multi-fidelity applications is that it considers all the outputs with the same weight, meaning that they provide the same level of information, it is referred to a symmetrical approach. By treating the outputs equally, symmetric covariance functions are implemented in order to capture the output correlations through the share of useful information across the outputs as much as possible. However, in the multi-fidelity framework, asymmetrical information are available. Indeed, to improve the predictions of the expensive high-fidelity output $f_s(\cdot)$ it tries to transfer information from the inexpensive lower fidelity outputs. The multi-fidelity modeling utilizes the correlated inexpensive lower-fidelity information to enhance the expensive high-fidelity modeling. The GP-based approaches presented in the next sections account for this asymmetrical information.

\subsection{Auto-Regressive model (AR1)}\label{AR1_theory}
\begin{figure}[!h]
\begin{center}
\includegraphics[width=0.4\linewidth]{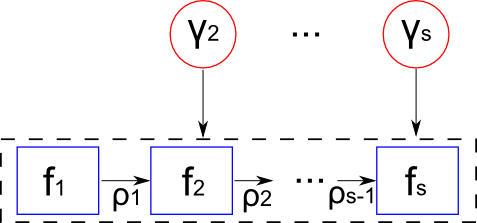}
\caption{AR1 schematic view}
\label{AR1_example}
\end{center}
\end{figure}

The Auto-Regressive (AR1) method (Figure \ref{AR1_example}) is the most classical one for multi-fidelity modeling and the most used one in engineering design problems \cite{laurenceau2008building,kuya2011multifidelity,toal2011efficient,keane2012cokriging,toal2014multifidelity,fernandez2016review,bailly2019multifidelity}. It relies on a linear autoregressive information fusion scheme introduced by Kennedy and O'Hagan \cite{kennedy2000predicting}, assuming a linear dependency between the different model fidelities.

AR1 assigns a GP prior to each fidelity model $t$ where the higher-fidelity model prior $f_t(\cdot)$ is equal to the lower-fidelity prior $f_{t-1}(\cdot)$ multiplied by a scaling factor $\rho(\mathbf{x})$ plus an additive bias function $\gamma_t(\cdot)$:
\begin{equation}\label{AR1_eq}
f_t(\mathbf{x}) = \rho_{t-1}(\mathbf{x}) f_{t-1}(\mathbf{x}) + \gamma_t(\mathbf{x})
\end{equation} 
$\rho_{t-1}(\mathbf{x})$ is a scale factor and quantifies the correlation between the outputs $y_t$ and $y_{t-1}$, and $\gamma_t(\cdot)$ is a GP with mean $\mu_{\gamma_t}$ and covariance function $k_t(\cdot)$, meaning that $\gamma_t \sim \mathcal{GP} \left(\gamma_t | m_{\gamma_t},k_t(\mathbf{x},\mathbf{x'}, \boldsymbol{\theta}_t) \right)$, with $\boldsymbol{\theta}_t$ the hyperparameters of the covariance function $k_t(\cdot)$. $\rho_{t-1}(\cdot)$ is often assumed as a constant function, meaning that:
\begin{equation}
f_t(\mathbf{x}) = \rho_{t-1} f_{t-1}(\mathbf{x}) + \gamma_t(\mathbf{x})
\end{equation} 

Considering this hierarchy of $s$ levels of code (from the less accurate
to the most accurate), for each level the conditional distribution of the GP $f_t(\cdot)$ knowing $f_1(\cdot), \ldots, f_{t-1}(\cdot)$ is only determined by $f_{t-1}(\cdot)$.
To obtain the relationship induced by AR1 model in Eq.(\ref{AR1_eq}), it is assumed that $Cov \left(f_t(\mathbf{x}),f_{t-1}(\mathbf{x}') | f_{t-1}(\mathbf{x}) \right) = 0, \; \forall \mathbf{x} \neq \mathbf{x}' $. It means that if $f_{t-1}(\cdot)$ is known, nothing more can be learnt for $f_t(\cdot)$ from any simulation of the cheaper code $f_{t-1}(\mathbf{x}')$ for $\forall \mathbf{x}' \neq \mathbf{x}$. Two main alternative numerical schemes exist for AR1 GPs inference: a fully coupled one proposed by Kennedy and O'Hagan \cite{kennedy2000predicting} and a recursive inference introduced by Le Gratiet and Garnier \cite{le2014recursive}.

The recursive inference for AR1 assumes that the DoE for the different fidelities $\{\mathcal{D}_1, \mathcal{D}_2, \ldots, \mathcal{D}_s \}$ have a nested structure, meaning that $\mathcal{D}_s \subseteq \mathcal{D}_{s-1} \subseteq \cdots \subseteq \mathcal{D}_1$, the DoE of higher fidelity is a subset of the DoE of lower fidelity. In order to recursively learn the GPs, the GP prior $f_{t-1}(\cdot)$ in Eq.(\ref{AR1_eq}) is replaced by the GP posterior $f^*_{t-1}(\cdot)$ of the previous inference level. This inference scheme is equivalent to the fully coupled one proposed by Kennedy and O'Hagan \cite{kennedy2000predicting} (considering the nested structure of the DoEs), meaning that the GP posterior distribution predicted by the fully coupled scheme is matched by the recursive inference of Le Gratiet and Garnier \cite{le2014recursive}. The advantage of the recursive inference approach is that it results in $s$ standard GP regression and offers a decoupled inference approach, simplifying the learning of the hyperparameters.
By doing so, the multi-fidelity GP posterior distribution $p(f_t|y_t,x_t,f^*_{t-1})$ for $t=1,\dots, s$ is defined by the following predictive mean and variance for each level:
\begin{equation}
\mu^*_{t}(\mathbf{x}) = \rho_{t-1} m^*_{t-1}(\mathbf{x}) + m^*_{\gamma_t} + k_{\mathbf{x}M_t}K^{-1}_t \left(y_t - \rho_{t-1} m^*_{t-1}(x_t) - m^*_{\gamma_t} \right)
\end{equation}
\begin{equation}
\sigma^{*^2}_{t}(\mathbf{x}) = \rho_{t-1} \sigma^{*^2}_{t-1}(\mathbf{x}) + k_{\mathbf{x},\mathbf{x}} - k_{\mathbf{x}M_t}K^{-1}_t k^T_{\mathbf{x}M_t}
\end{equation}
\noindent where $k_{\mathbf{x}M_t}=[k(\mathbf{x},\mathbf{x_{{1}_t}}),\cdots,k(\mathbf{x},\mathbf{x_{{M}_t}})]$ with $\mathbf{x_{{i}_t}}$ the $i^\text{th}$ data point of DoE at fidelity $t$. AR1  has been extended for scalability purpose to account for high dimensional problems (for instance with Proper orthogonal decomposition \cite{xiao2018extended} or Nystrom approximation of sample covariance matrices \cite{zaytsev2017large}). 

As it can be seen in Eq.(\ref{AR1_eq}), AR1 only assumes a certain linear relationship between the fidelities. Moreover, AR1 may be seen as a particular case of co-kriging using LMC for particular value of the coregionalization matrix. This linear mapping between the fidelity may be a limitation for some engineering design problems where this dependence structure is not appropriate. More advanced approaches have been developed to account for non-linear dependencies between the fidelities. Two principal approaches are presented in the next sections.

\subsection{Non-linear Auto-Regressive multi-fidelity Gaussian Process (NARGP)}

In order to generalize the AR1 approach, Perdikaris \textit{et al.} \cite{perdikaris2017nonlinear} proposed a non-linear mapping between the fidelities called NARGP:
\begin{equation}
f_t(\mathbf{x}) = z_{t-1}\left(f_{t-1}(\mathbf{x}) \right) + \gamma_t(\mathbf{x})
\end{equation}
with $z_{t-1}(\cdot)$ a mapping function between two successive fidelity models with an assigned GP prior. As $f_{t-1}(\cdot)$ is a GP, $z_{t-1}\left(f_{t-1}(\cdot) \right)$ is a composition of two GPs and the non-linear mapping of a Gaussian distribution is not analytically tractable in practice. The posterior distribution of $f_t(\cdot)$ is no longer Gaussian.
In order to alleviate this intractability, the authors proposed to follow the same recursive inference strategy proposed by Le Gratiet and Garnier for AR1 \cite{le2014recursive}. It also requires to satisfy the same hypotheses, especially on the nested DoE assumption. In the inference, the GP prior of $f_{t-1}(\cdot)$ is replaced with the GP posterior $f^*_{t-1}(\cdot)$ obtained with the previous fidelity level. Following this assumption, NARGP model may be expressed by:
\begin{equation}
f_t(\mathbf{x}) = g_t \left(\mathbf{x},f^*_{t-1}(\mathbf{x})\right)
\end{equation}  
with $g_t \sim \mathcal{GP}\left(f_t|0,k_t\left( (\mathbf{x},f^*_{t-1}(\mathbf{x})),(\mathbf{x'},f^*_{t-1}(\mathbf{x'}) \right)\right)$.
It is important to note that it follows the same assumption as AR1, meaning that $\gamma_t(\cdot)$ and $z_{t-1}(\cdot)$ are independent. 
NARGP defines a mapping $\mathbb{R}^{M_{t-1}+1} \rightarrow \mathbb{R}$ between the input space of the lower fidelity model $t-1$ plus its corresponding output $M_{t-1}+1$ to the higher fidelity model $t$ output. 

The authors proposed a specific covariance function for $g_t(\cdot)$ that reflects the non-linear structure:
\begin{equation}
k_t(\mathbf{x},\mathbf{x'}) = k_{z_{t-1}}(\mathbf{x},\mathbf{x'}) \times k_{f_{t-1}}\left(f^*_{t-1}(\mathbf{x}),f^*_{t-1}(\mathbf{x'})\right) + k_{g_t}(\mathbf{x},\mathbf{x'})
\end{equation}
Similarly to the kernels presented in Section \ref{LMC_section}, a separable kernel is considered where the treatment of the input variables and the output variables is decoupled. The proposed approach extends the capabilities of AR1 and enables to capture non-linear, non-functional and space-dependent cross-correlations between the low and high-fidelity models.

Due to the intractability and the recursive inference strategy, for $t\geq 2$, the posterior distribution $f^*_{t-1}(\mathbf{x})$ at a location $\mathbf{x}$ is no longer a Gaussian distribution and in the training process, it is necessary to approximate it by uncertainty propagation along each recursive step. Therefore, the posterior distribution for fidelity model $t$ is estimated by:
\begin{equation}
p\left(f^*_{t}(\mathbf{x})\right) = \int p\left( f^*_{t}(\mathbf{x},f^*_{t-1}(\mathbf{x}))\right) p\left(f^*_{t-1}(\mathbf{x})\right) \text{d}\mathbf{x}
\end{equation}
where $p\left(f^*_{t-1}(\mathbf{x})\right)$ is the posterior distribution at the previous level.

NARGP resumes to a disjointed architecture in which a GP for each fidelity is fitted in an isolated hierarchical manner. Therefore GPs at lower fidelities are not updated once they have been fitted. To avoid this limitation, Cutajar \textit{et al.} \cite{cutajar2019deep} proposed to extend NARGP to a fully Deep Gaussian Process (DGP) by training all the layers in a coupled fashion.

\subsection{Multi-Fidelity Deep Gaussian Process (MF-DGP)}

\begin{figure}[!h]
\begin{center}
\includegraphics[width=0.8\linewidth]{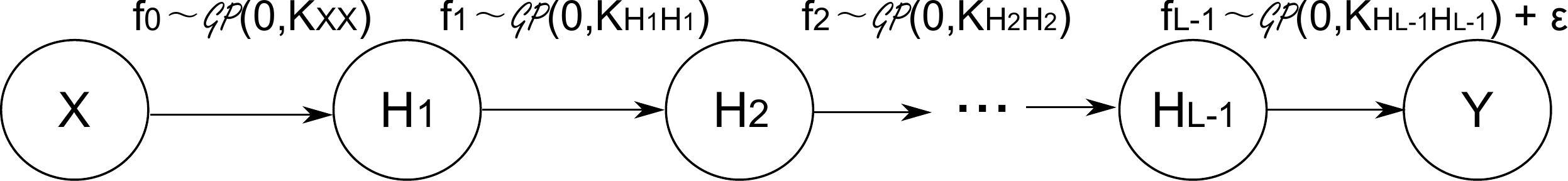}
\caption{DGP schematic view}
\label{DGP_schematic}
\end{center}
\end{figure}

Deep Gaussian Processes (DGP) have been introduced \cite{damianou2013deep} as a nested structure of GPs representing the mapping between the inputs and the output as a functional composition of GPs (Figure \ref{DGP_schematic}):
\begin{equation}
y(\mathbf{x}) = f_{L-1} \left( \cdots f_0(\mathbf{x})) \right) + \epsilon
\end{equation}

\begin{figure}[!h]
\begin{center}
\includegraphics[width=0.4\linewidth]{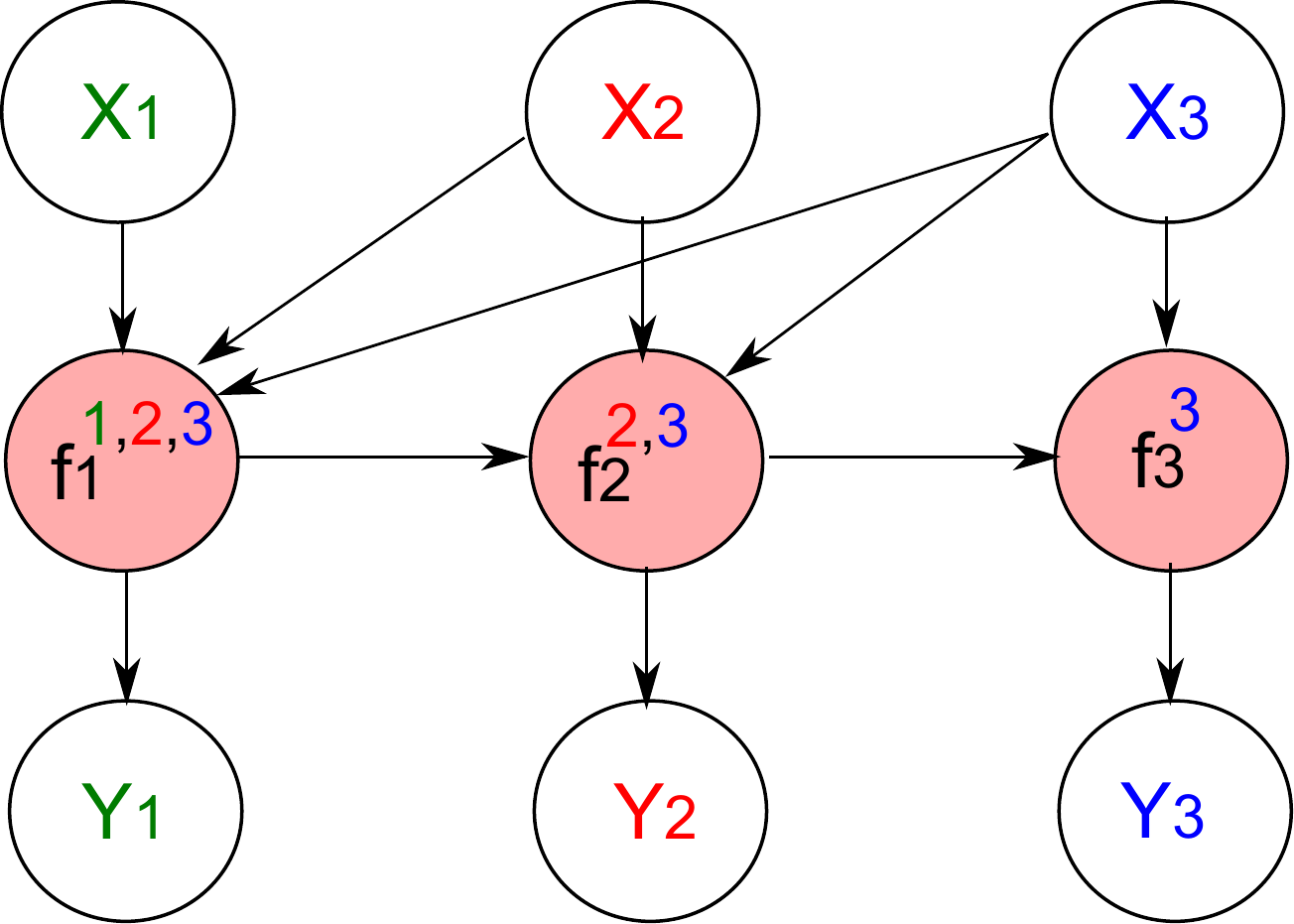}
\caption{MF DGP schematic view}
\label{MF_DGP_schematic}
\end{center}
\end{figure}

\noindent with $L$ the number of layers. Each layer corresponds to a multi-output GP between input node $H_{l-1}$ and an output node $H_l$. Cutajar \textit{et al.} \cite{cutajar2019deep} proposed an adaptation of DGP for multi-fidelity modeling in which each layer represents a fidelity (Figure \ref{MF_DGP_schematic}). Similarly to NARGP, the inference of DGP is not analytically tractable due to the marginal likelihood computing: 
\begin{equation}
p(y|\mathbf{x}) = \int p(H_L|H_{L-1}) \dots p(H_1|\mathbf{x})\text{d}H_1\dots\text{d}H_{L-1}
\end{equation}

\begin{figure}[!h]
\begin{center}
\includegraphics[width=0.4\linewidth]{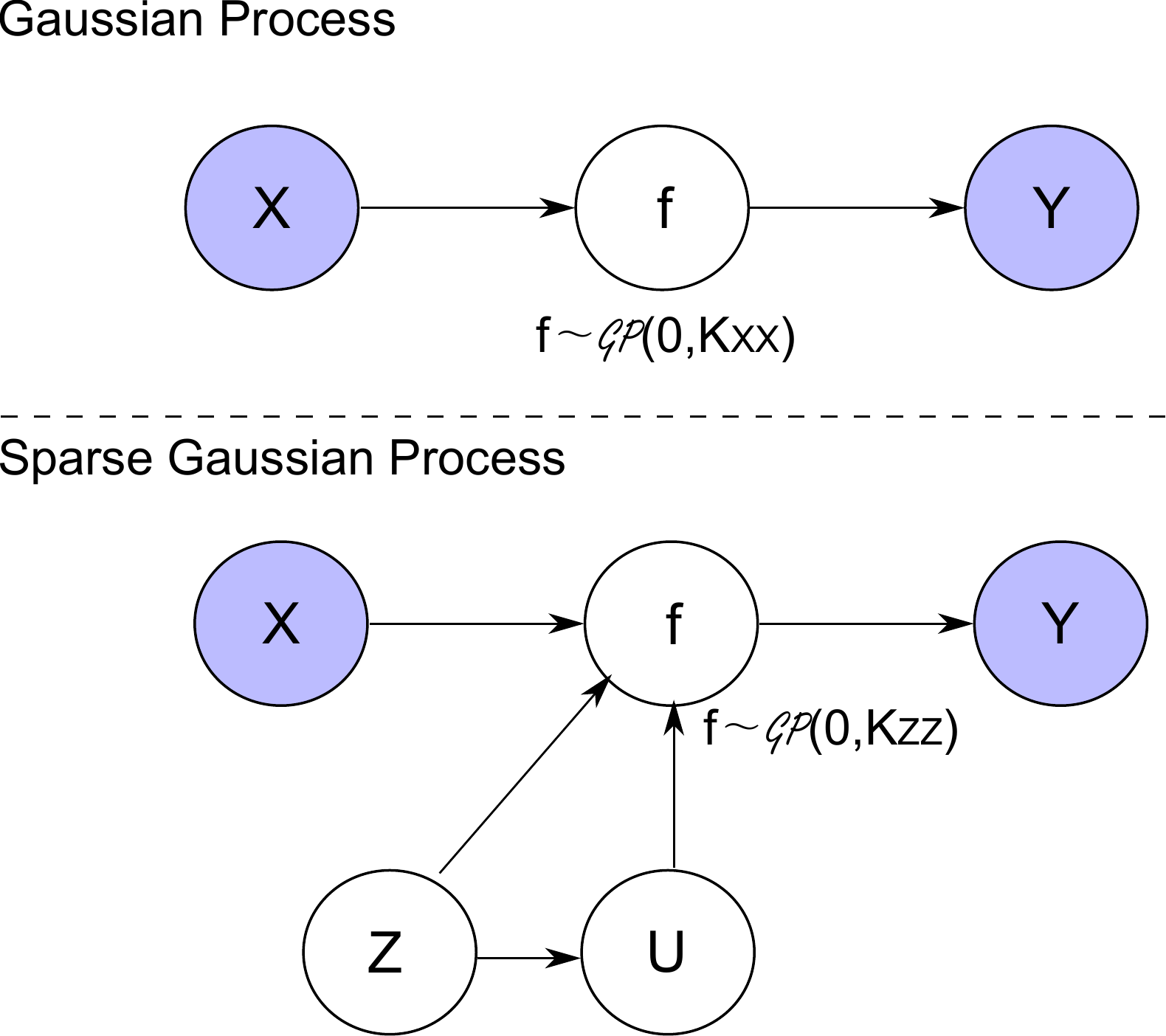}
\caption{Sparse GP schematic view, $(X,Y)$ is the available dataset}
\label{Sparse_GP_schematic}
\end{center}
\end{figure}

The GP at each layer is conditioned on the data belonging to that level, as well as the evaluation of that same input data at the preceding fidelity level.
Instead of using a disjointed architecture for the inference, dedicated inference strategies to DGP have been proposed to keep the mapping between the layers \cite{damianou2013deep,wang2016sequential,salimbeni2017doubly} including the doubly stochastic variational inference derived from sparse GP \cite{salimbeni2017doubly}. At each layer for inference, the sparse variational approximation of a GP is considered (Figure \ref{Sparse_GP_schematic}). 

It consists in introducing a set of inducing input variables $\mathbf{Z}_l$ for each layer $l$ and defining them as training optimization variables along with the GP hyperparameters $\boldsymbol{\theta}_l$ and the hyperparameters $(\mathbf{\bar{U}}_l, \boldsymbol{\Sigma}_l)$ of the variational distribution $q_l(\mathbf{U}_l) \sim \mathcal{N}(\mathbf{\bar{U}}_l, \boldsymbol{\Sigma}_l)$ corresponding to the inducing variables responses through the GPs: $\mathbf{U}_l = f_l(\mathbf{Z}_l)$. Therefore, compared to the previous multi-fidelity models (AR1, LMC and, NARGP), in MF-DGP, the number of hyperparameters that have to be trained is greatly increased, from few dozen (for AR1, LMC and, NARGP) to few hundreds or even thousands making more difficult the training of the MF-DGP.

In MF-DGP, for the intermediate layers, the inputs are the combination of the dataset input points $X_l$ along with the corresponding function evaluations from the previous layer $f_{l-1}(X_l)$. The dimension is therefore increased by one dimension in addition to the dimension of $X_l$. Consequently, freely optimizing the inducing inputs is not appropriate. The inducing inputs are fixed during the MF-DGP training based on the available observations at the previous fidelity layer \cite{cutajar2019deep}. Moreover, the optimization of the variational distributions is performed using gradient-based technique (Adam optimizer \cite{kingma2014adam}), which may be inappropriate when optimizing a distribution \cite{amari1998natural}.
An alternative training approach for MF-DGP has been proposed in \cite{hebbal2019multi} in order to alleviate these limitations.
MF-DGP enables to account for complex relationship between the fidelity models while keeping the nested layer architecture and training them in a coupled fashion. However, due to the analytical intractability of the marginal likelihood, a more complex inference process is required.

\vspace{0.2cm}

\begin{figure}[!h]
\begin{center}
\includegraphics[width=0.6\linewidth]{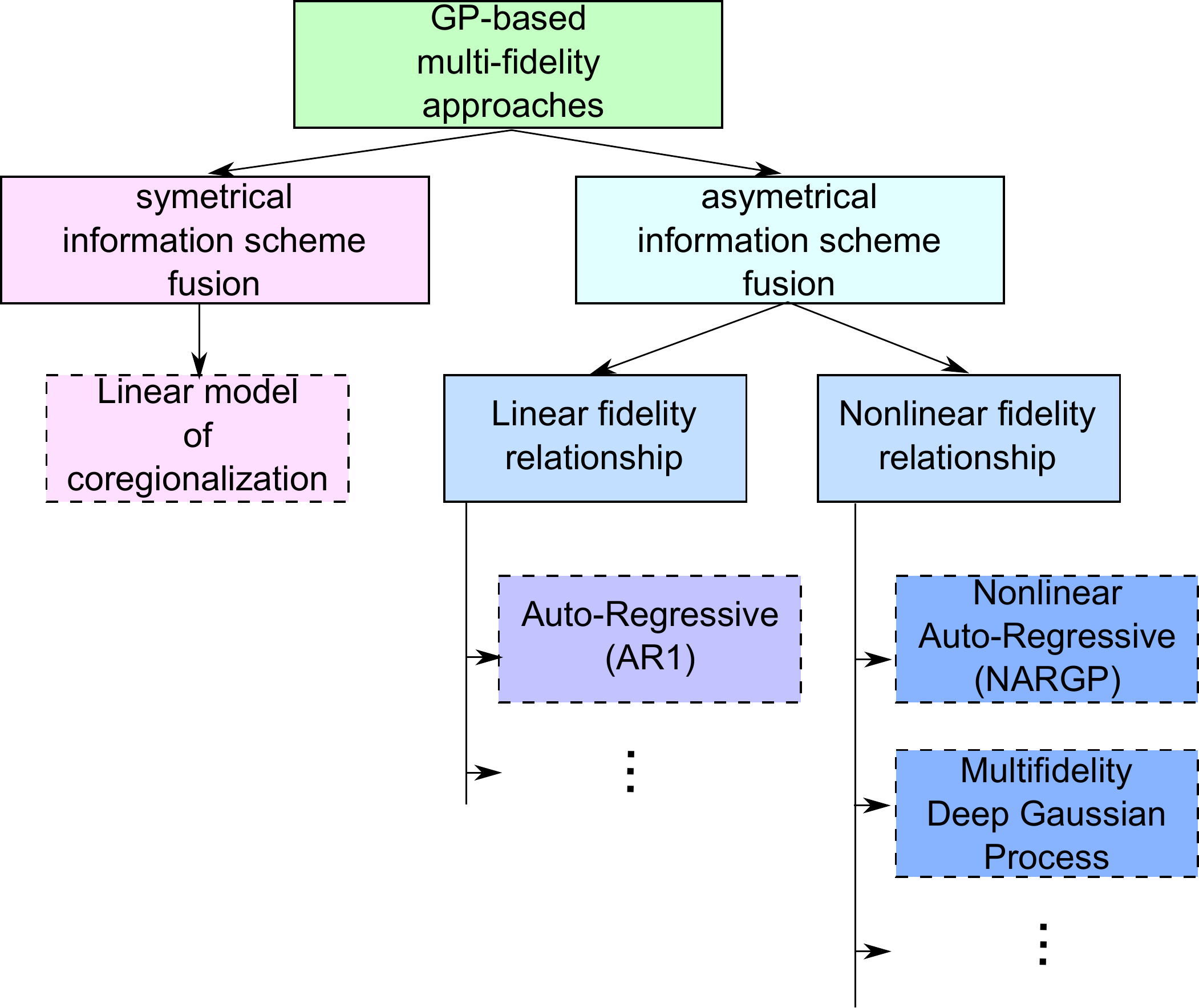}
\caption{Classication of GP-based multi-fidelity approaches}
\label{Classification_GP_multifi}
\end{center}
\end{figure}

For the considered GP-based multi-fidelity approaches described in the previous sections, a classification is illustrated in Figure \ref{Classification_GP_multifi}. The main distinctions correspond to the symmetrical or asymmetrical treatment of the fidelity information and the linear or non-linear relationship between the fidelities.

In the following sections, the different methods described in the previous sections (LMC, AR1, NARGP, MF-DGP) are evaluated and compared on different benchmarks based on analytical functions and aerospace engineering applications. The results of this comparative analysis are presented in the next sections.

\section{Analytical and aerospace applications}\label{Section_analytic}
\subsection{Settings and implementations}
For the benchmark of analytical test cases and aerospace problems, six techniques are compared: a GP using only the HF dataset (GP HF), the auto-regressive model (AR1)  with inference scheme introduced by Kennedy and O'Hagan \cite{kennedy2000predicting}, the co-kriging linear model of corregionalization (LMC), the non-linear auto-regressive multi-fidelity gaussian process without nested DoE (NARGP) and with nested DoE (NARGP*), and multi-fidelity Deep Gaussian Process (DGP).
For the considered problems, several sizes of design experiments are considered for the HF dataset to analyze the influence of HF data quantity. In order to assess the robustness of the methods to the LF and HF datasets, the experimentations are repeated on 20 different DoEs using Latin Hypercube Sampling for each size of the dataset. For the nested DoE of NARGP*, the same DoE as other techniques are considered except that the HF DoE is included in the LF DoE. To have the same number of samples in LF DoE, the same number of HF samples are removed from LF DoE.\\

The multi-fidelity methods are compared with respect to three metrics: the coefficient of determination (R2), the Root Mean Square Error (RMSE) and the Mean Negative test LogLikelihood (MNLL). These metrics are defined as follows. Considering a test set $(\mathbf{x}_\text{test},\mathbf{y}_\text{test})$ of size $n_\text{test}$ and the associated highest multi-fidelity predicted values $\mathbf{\hat{y}}$ and the variance of the prediction $\mathbf{\hat{\sigma}}^2$:
\begin{itemize}
    \item coefficient of determination
$R^2=1-\frac{\sum_{i=1}^{n_{\text{test}}} \left(y_\text{test}^{(i)} - \hat{y}^{(i)} \right)^2}{\sum_{i=1}^{n_{\text{test}}} \left(y_\text{test}^{(i)} - \bar{y} \right)} $ with $\bar{y}$ the mean of the observed data.
\item root mean square error $RMSE = \sqrt{\frac{\sum_{i=1}^{n_{\text{test}}} \left(y_\text{test}^{(i)} - \hat{y}^{(i)} \right)^2}{n_{\text{test}}}}$
\item mean negative test log likelihood (Gaussian case) $MNLL = - \frac{1}{n_{\text{test}}} \sum_{i=1}^{n_{\text{test}}} \log\left(\phi\left(\frac{y_\text{test}^{(i)} - \hat{y}^{(i)}}{\hat{\sigma}^{(i)}} \right) \right)$
\end{itemize}

A large HF test set is used to estimate the metrics. For GP-based multi-fidelity methods, it is important to compare the prediction accuracy metrics (R2 the higher the better and RMSE, the lower the better) and the likelihood of the multi-fidelity model to accurately explain the test set (MNLL, the lower the better) especially with respect to the uncertainty model associated to the prediction. Indeed, the uncertainty model for prediction of GP-based techniques is often used either for surrogate model refinement, uncertainty propagation or optimization. Therefore, the surrogate model has to be accurate both in terms of prediction and of uncertainty model associated to the prediction. The obtained results are presented through numerical tables and boxplot figures. The tables present the mean value and the standard deviation for R2, RMSE and MNLL considering 20 repetitions from different LHS for the training multi-fidelity set. Moreover, an indicator providing the improvement of RMSE of the multi-fidelity techniques with respect to the single fidelity GP HF is added. A negative value means that the mutli-fidelity technique improves the RMSE compared to GP HF by an amount of x\%.\\

The different multi-fidelity methods are implemented using Gpflow \cite{de2017gpflow} and Emukit \cite{emukit2019} (relying on GPy \cite{gpy2014}) tools. Gradient-based optimizers are used to train the models (multi-start BFGS \cite{fletcher2013practical} and Adam optimizer \cite{kingma2014adam}). The optimizer settings have been adapted to the dimensions of the test cases.
All GP-based multi-fidelity techniques are implemented with a squared exponential kernel. Co-kriging with LMC is based on a coregionalization matrix of rank 2 (corresponding to two independent latent functions). 

\subsection{Analytical test-cases}
To highlight interesting features of the different multi-fidelity methods, two different analytical test problems are considered. Two axes are explored with these analytical problems. With the first problem, the impact of the type of relationship between the low and the high fidelities (\textit{e.g.}, linear, non-linear, degree of correlation) is studied. With the second problem, the influence of the input space dimensionality and the number of high-fidelity samples are explored.

\subsubsection{Influence of the  relationship linearity between low and high fidelities}

\begin{figure}[!h]
\begin{center}
\includegraphics[width=.75\linewidth]{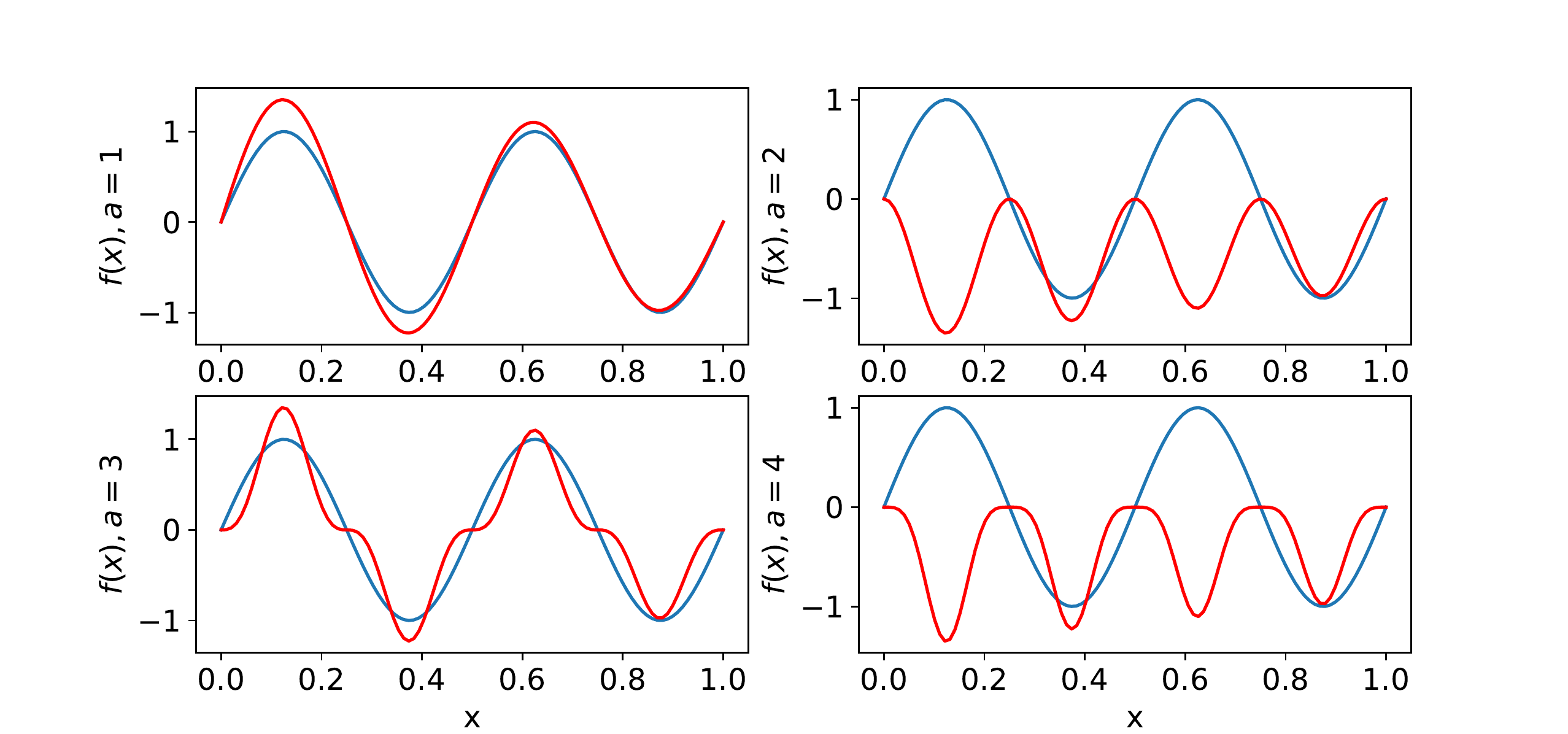}
\caption{Low-fidelity and high-fidelity responses for different values of $a$ (in blue: low-fidelity model, in red: high-fidelity model)}
\label{fig:cutajar}
\end{center}
\end{figure}

\begin{figure}[!h]
\begin{center}
\includegraphics[width=.75\linewidth]{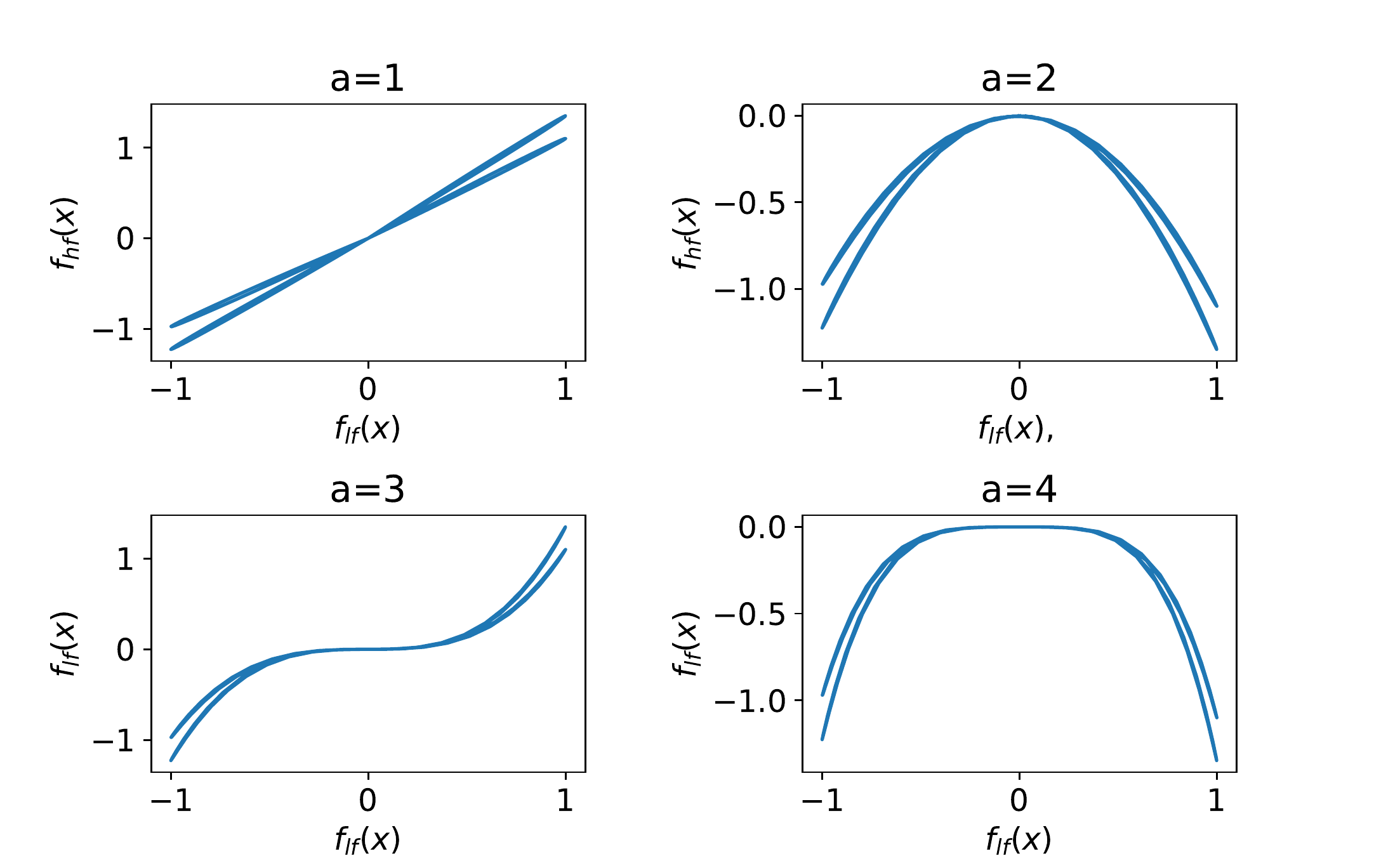}
\caption{Relationship between low and high-fidelity models with different values of  $a$}
\label{fig:cutajar-relationship}
\end{center}
\end{figure}

The first analytical problem is a one dimension problem with a parameter allowing to vary the linearity of the relationship between the low and the high-fidelity models. It is a modified function from  \cite{cutajar2019deep}. It enables to study the influence of the relationship (\textit{e.g.}, linearity, degree of correlation) between the low and high-fidelity data on the different multi-fidelity models. The two fidelity models are defined as follows :
\begin{eqnarray}
f_{hf}(x)& = & \sin(2\pi x)\\
f_{lf}(x)&=&\left(\frac{x}{4}-\sqrt{2}\right)\sin(2\pi x+a \pi)^{a}
\end{eqnarray}

\noindent with $x \in[0,1]$ and $a$ a term that allows to modify the type of relationship between the two models. In this experimentation, four values of $a$ are considered: 1, 2, 3 and 4. The different functions are depicted in Figure \ref{fig:cutajar}. As it can be seen, for $a=1$, the low and high-fidelity models are very similar, the difference being the amplitude of the sinusoidal signal. For $a=3$, the low and high-fidelity models share similarities in terms of oscillations and phases but the high-fidelity model presents more complexity not captured by the low-fidelity model. For $a=2$ and $a=4$, the low and high-fidelity models do not share the same oscillation frequency and amplitude. The high-fidelity models in these cases present a higher oscillation frequency and a smaller amplitude.
The relationships between the low-fidelity (LF) and high-fidelity (HF) responses are depicted in Figure \ref{fig:cutajar-relationship}. For $a=1$ and $a=3$ the models are positively correlated. LF and HF exact models presents a $R2$ of $0.97$ for $a=1$ and $0.87$ for $a=3$ highlighting the strong correlation between the models. The relationship for $a=3$ is non linear whereas for $a=1$ it can be considered as linear. For $a=2$ and $a=4$, the relationship between the two models is very complex (positive and negative correlation with different frequencies). LF and HF exact models present a $R2$ of $-4.93$ for $a=2$ and $-3.90$ for $a=4$ illustrating the weak correlation between the two models.
Regarding the variation of the relationship between the low and high-fidelity responses, this test case aims at evaluating the influence of the linearity or non-linearity between the low and high-fidelity models. It is expected that as the relationship between the low-fidelity and high-fidelity models becomes non-linear, the relative performance of AR1 and LMC will decrease compared to NARGP and DGP.

\begin{figure}[!h]
\begin{center}
\includegraphics[width=1.\linewidth]{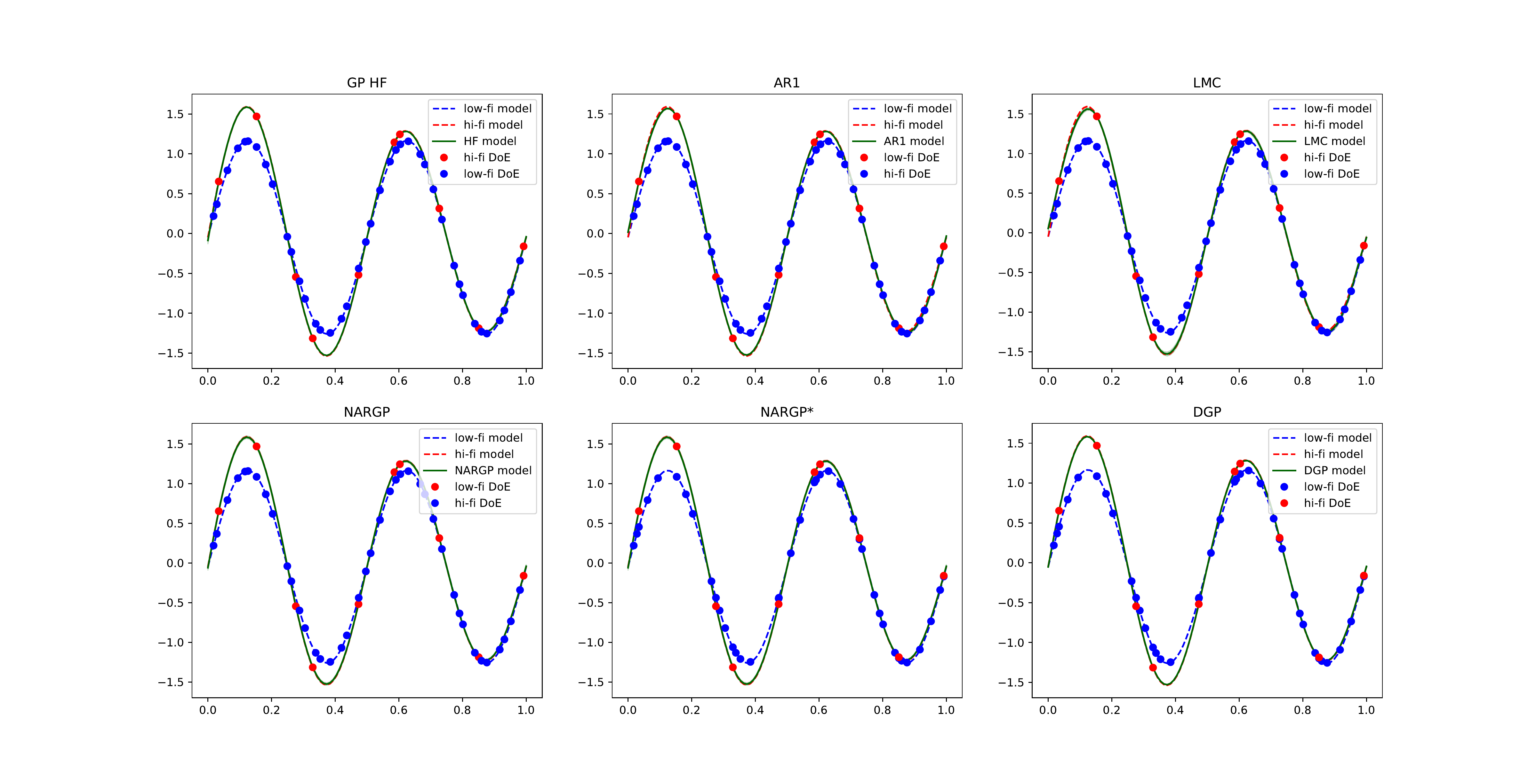}
\caption{Prediction for the 1D problem for the different multi-fidelity techniques with $a=1$ for one repetition}
\label{fig:1D_a=1_plots}
\end{center}
\end{figure}

\begin{figure}[!h]
\begin{center}
\includegraphics[width=1.\linewidth]{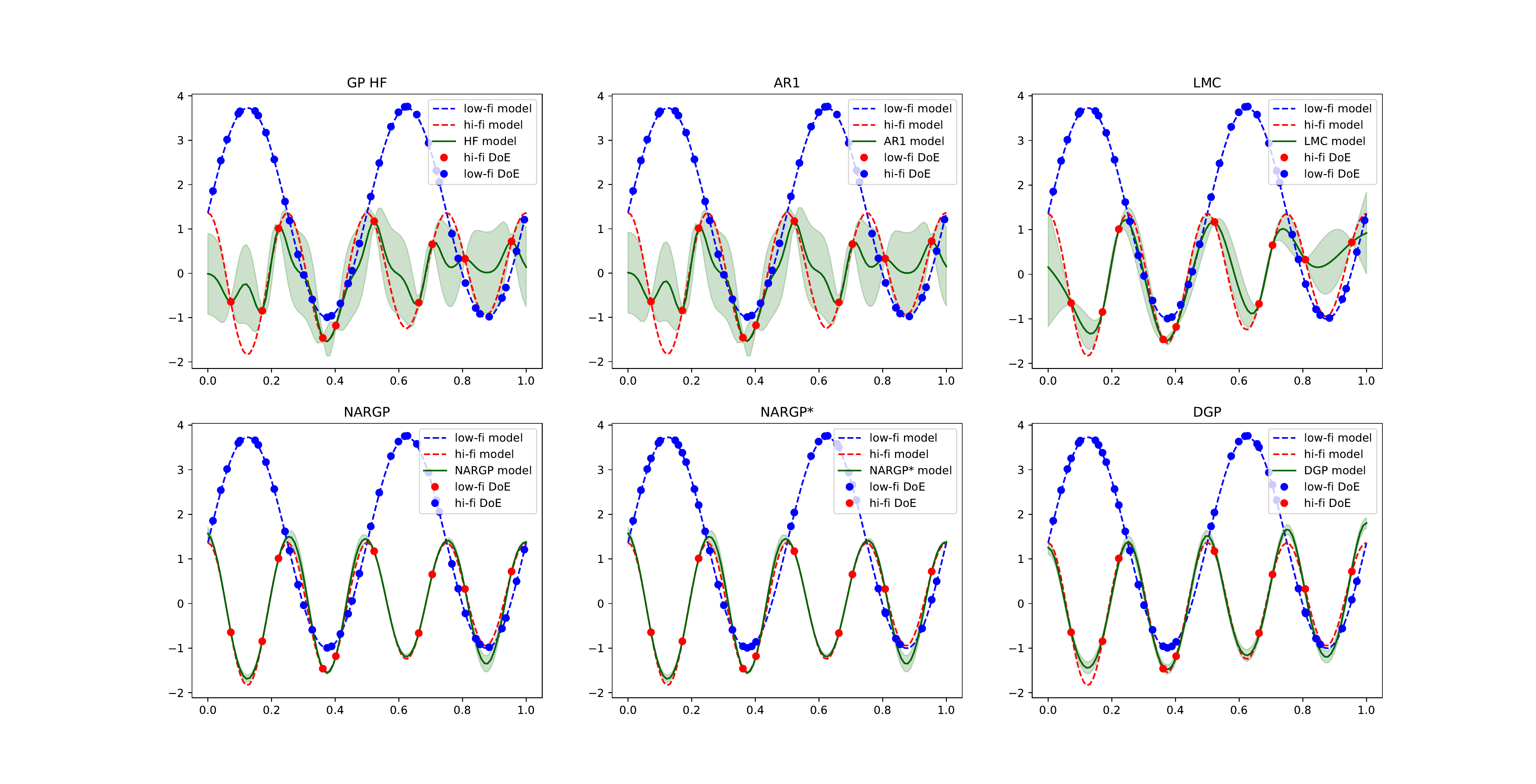}
\caption{Prediction for the 1D problem for the different multi-fidelity techniques with $a=2$ for one repetition}
\label{fig:1D_a=2_plots}
\end{center}
\end{figure}

\begin{figure}[!h]
\begin{center}
\includegraphics[width=1.\linewidth]{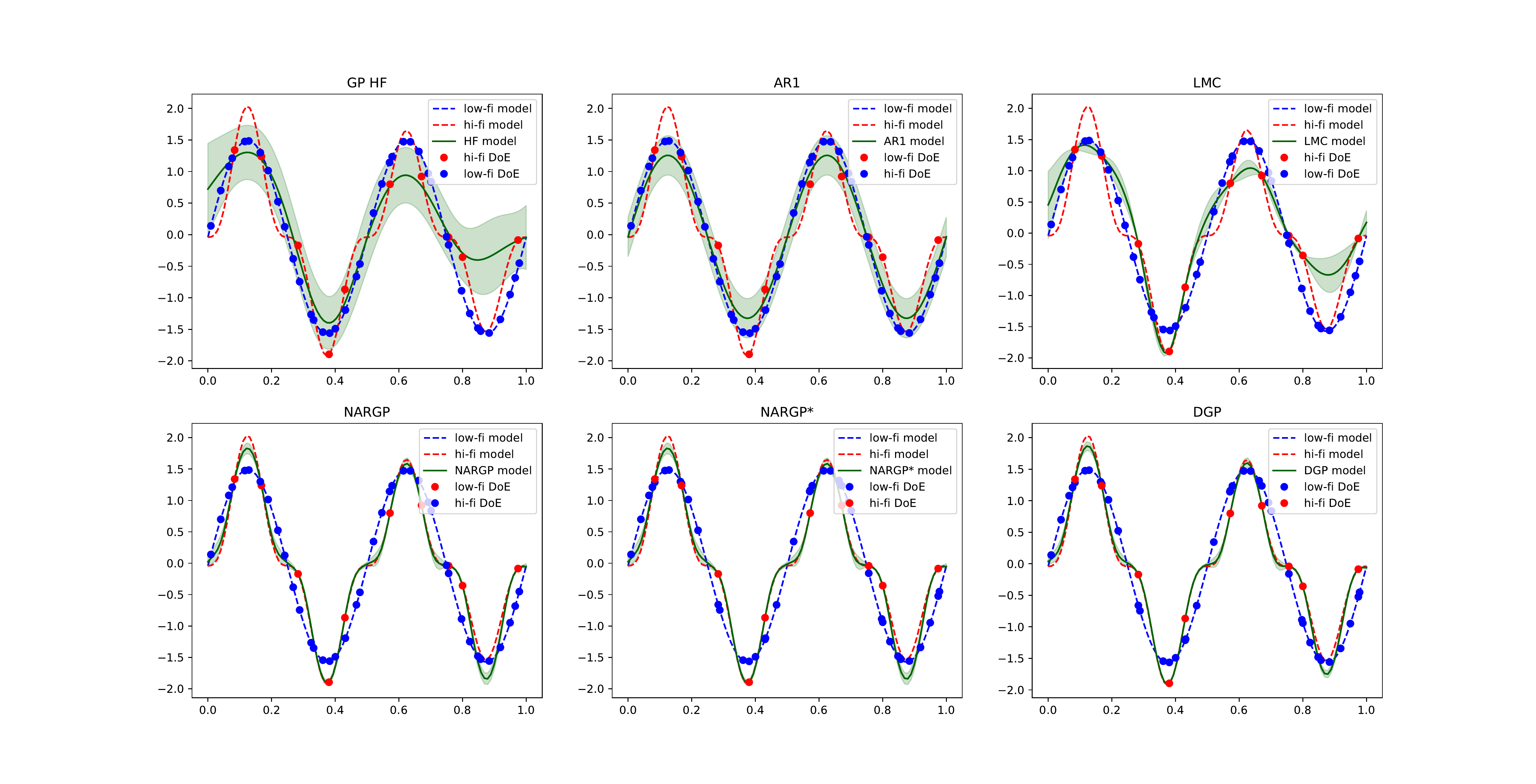}
\caption{Prediction for the 1D problem for the different multi-fidelity techniques with $a=3$ for one repetition}
\label{fig:1D_a=3_plots}
\end{center}
\end{figure}

\begin{figure}[!h]
\begin{center}
\includegraphics[width=1.\linewidth]{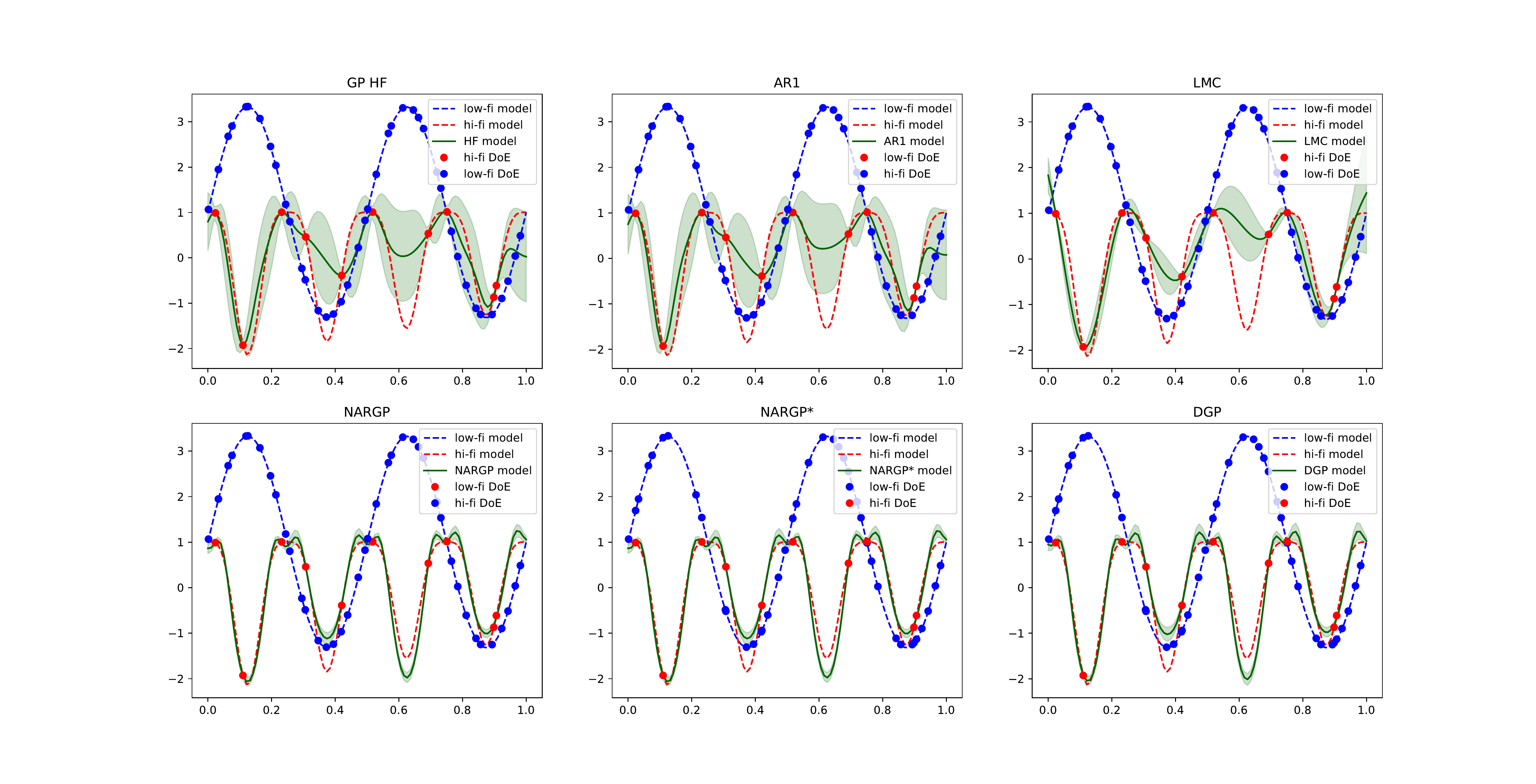}
\caption{Prediction for the 1D problem for the different multi-fidelity techniques with $a=4$ for one repetition}
\label{fig:1D_a=4_plots}
\end{center}
\end{figure}

\begin{figure}[!h]
\begin{center}
\includegraphics[width=0.5\linewidth]{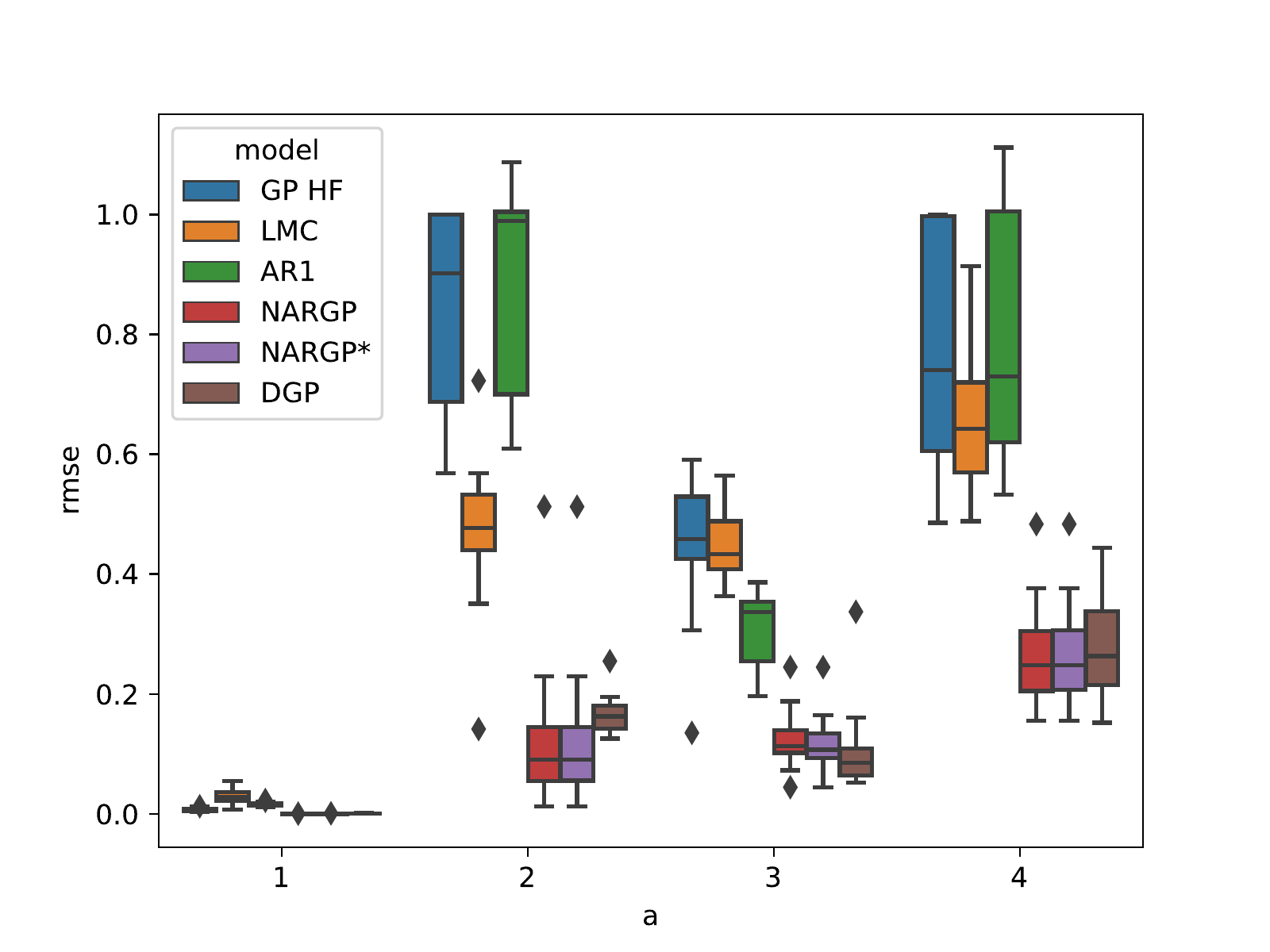}\includegraphics[width=0.5\linewidth]{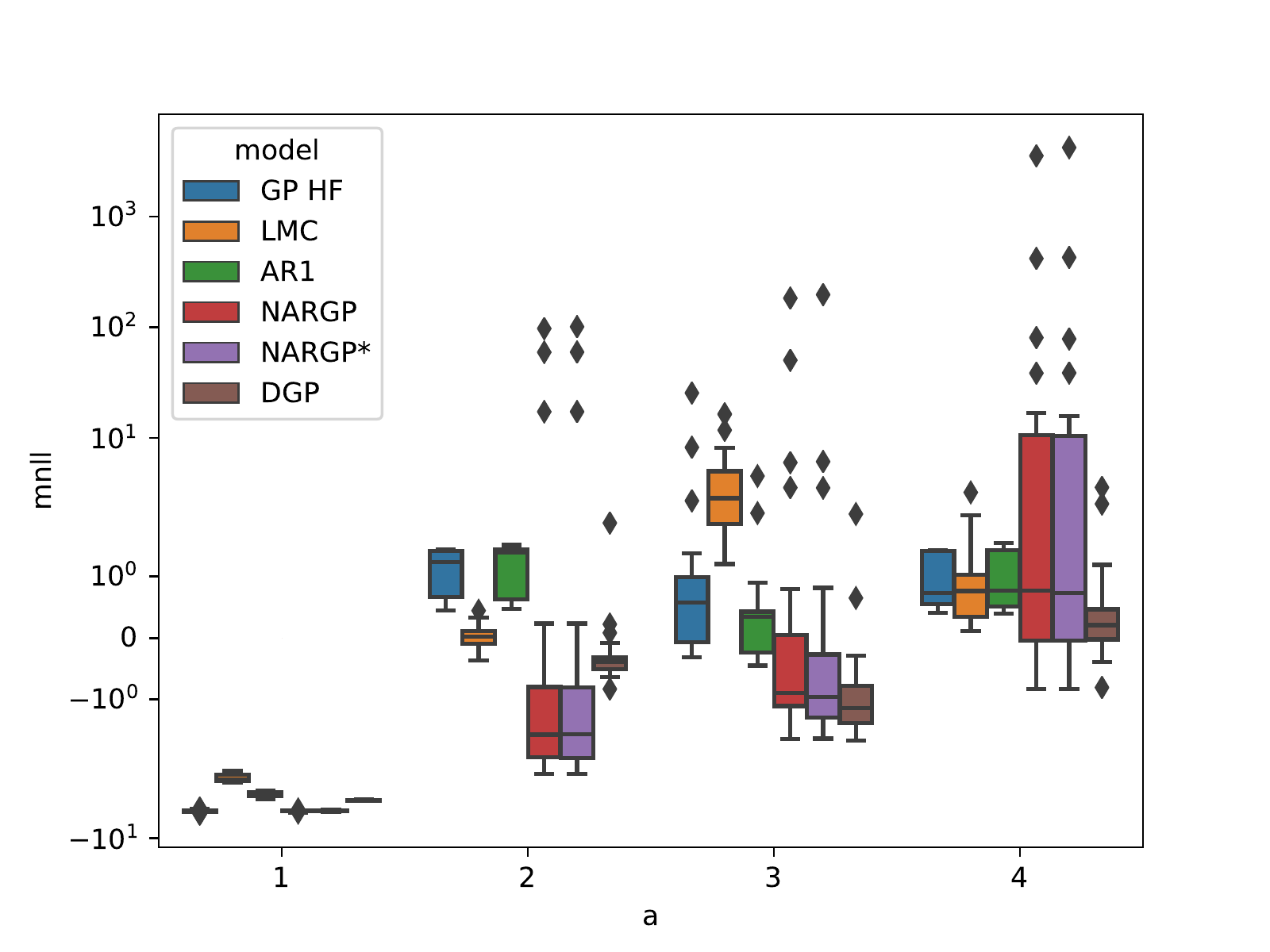}
\caption{Boxplots of RMSE and MNLL for different values of $a$}
\label{cutajar}
\end{center}
\end{figure}

\begin{table}[h!]
\scriptsize
\begin{center}
\caption{Summary of the results obtained on the 1D test case}
\label{1D_results_table}
\begin{tabular}{|c|c|c|c|c|c|}
\hline
\multirow{2}{*}{Function}&\multirow{2}{*}{Method}&\multirow{2}{*}{R2 (std)}&\multirow{2}{*}{RMSE (std)}&\multirow{2}{*}{MNLL (std)}&Evolution of\\
&&&&&RMSE wrt GP HF\\
\hline
\multirow{6}{*}{$a=1$}&GP HF&9.999e-1(4.208e-5)&7.337e-3(2.493e-3)&-5.694(1.849e-1)&-\\
&LMC&9.990e-1(7.257e-4)&3.005e-2(1.129e-2)&-2.864 (2.107e-1)&$+310\%$\\
&AR1&9.997e-1(1.051e-4)&1.643e-2(2.980e-3)&-4.021 (2.070e-1)&$+124\%$\\
&NARGP&1.000(9.942e-8)&2.778e-4(1.309e-4)&-5.684 (8.572e-2)&$-96\%$\\
&NARGP*&1.000(3.107e-7)&3.675e-4(2.581e-4)&-5.667 (7.087e-2)&$-95\%$\\
&DGP&1.000(4.135e-7)&7.228e-4(2.472e-4)&-4.530 (2.688e-2)&$-90\%$\\\cline{2-6}
\hline
\multirow{6}{*}{$a=2$}&GP HF&2.702e-1(2.807e-1)&8.367e-1 (1.727e-1)& 1.062(3.824e-1)&-\\
&LMC&7.670e-1(9.458e-2)&4.707e-1(1.070e-1)&2.892e-2( 1.775e-1)&$-44\%$\\
&AR1&2.190e-1(2.879e-1)&8.669e-1(1.714e-1)&1.072(4.260e-1)&$4\%$\\
&NARGP&9.736e-1(5.646e-2)&1.208e-1(1.087e-1)&7.342(2.479e+1)&$-86\%$\\
&NARGP*&9.736e-1(5.642e-2)&1.208e-1(1.087e-1)&7.537( 2.549e+1)&$-86\%$\\
&DGP&9.719e-1(1.063e-2)&1.650e-1(2.875e-2)&-2.637e-1( 5.418e-1)&$-80\%$\\\cline{2-6}
\hline
\multirow{6}{*}{$a=3$}&GP HF&7.862e-1(7.840e-2)&4.511e-1( 1.013e-1)&2.165e(5.643)&-\\
&LMC&7.997e-1(4.718e-2)&4.446e-1(5.115e-2)&4.787(4.622)&$-1\%$\\
&AR1&9.016e-1(3.287e-2)&3.087e-1(5.539e-2)&4.176e-1(1.105)&$-32\%$\\
&NARGP&9.831e-01(1.271e-2)&1.226e-1(4.371e-2)& 1.145e+1 (4.096e+1)&$-73\%$\\
&NARGP*&9.846e-01(1.223e-2)&1.166e-1(4.249e-2)&9.573(4.312e+1)&$-74\%$\\
&DGP&9.856e-1(2.369e-2)&1.027e-1(6.209e-2)&-8.789e-1( 8.397e-1)&$-77\%$\\\cline{2-6}
\hline
\multirow{6}{*}{$a=4$}&GP HF&3.727e-1(2.832e-1)&7.712e-1( 1.805e-1)&8.902e-1(4.086e-1)&-\\
&LMC&5.512e-1(1.651e-1)&6.593e-1(1.188e-1)&8.732e-1( 7.251e-1)&$-15\%$\\
&AR1&3.287e-1(3.116e-1)&7.970e-1(1.898e-1)&9.146e-1( 4.336e-1)&$+3\%$\\
&NARGP&9.253e-1(4.735e-2)&2.624e-1(7.632e-2)&2.061e+2( 7.727e+2)&$-66\%$\\
&NARGP*&9.251e-1(4.740e-2)&2.628e-1(7.638e-2)&2.397e+2( 9.168e+2)&$-66\%$\\
&DGP&9.164e-1(4.814e-2)&2.776e-1(8.061e-2)&4.703e-1( 9.770e-1)&$-64\%$\\\cline{2-6}
\hline
\end{tabular}
\end{center}
\end{table}

Considering the results obtained in Table \ref{1D_results_table}, for the case $a=1$, all the multi-fidelity methods provide similar results, they all capture almost perfectly ($R2 \approx 1.0$ and $RMSE \leq 1.6\times 10^{-2}$) the high-fidelity model (see Figure \ref{fig:1D_a=1_plots}). Moreover, GP HF is also able to capture the HF model using only $10$ data points, meaning that too many HF data samples are used in the multi-fidelity framework to have a relative interest with respect to using a surrogate model on the HF data only (see Section \ref{Sect_influence_nb_pts_dim} for a discussion on the influence of the number of HF data points). When considering less HF data, for instance by considering only $5$ HF data points, RMSE increases to $0.72$ for GP HF ($R2$ of $0.44$). Linear multi-fidelity techniques AR1 and LMC enable to improve the RMSE compared to GP HF (RMSE of $0.11$ for AR1 and LMC) while non-linear techniques still improve it but are less efficient than AR1 or LMC (RMSE of $0.20$ for NARGP and NARGP*). 

In the case $a=3$, two categories of results may be distinguished. On one side, the linear multi-fidelity techniques AR1 and LMC present respectively 32\% and 1\% of improvement in terms of RMSE compared to GP HF. On the other side, NARGP and DGP present respectively 74\% and 77\% of improvements for the RMSE compared to GP HF. In Figure \ref{fig:1D_a=3_plots}, it can be seen that NARGP and DGP capture the complexity of the HF model using both LF and HF data whereas, AR1 and LMC are mainly influenced by LF behavior and do not capture the change in the HF oscillations near the zero value of the y-axis. Even if, in this case, the relation between low and high-fidelity models may be approximated by a linear tendency, the non-linear multi-fidelity methods (NARGP and DGP) provide more accurate results (RMSE in the order of $1.1 \times 10^{-1}$ compared to $3.1 \times 10^{-1}$ for AR1 and $4.7 \times10^{-1}$ for LMC). 

For $a=2$ and $a=4$, once again, two categories of results may be distinguished with LMC and AR1 on one side and NARGP and DGP on the other. Indeed, as expected, due to the strong non-linear relationship between low and high-fidelity models (Figure \ref{fig:cutajar-relationship}), AR1 and LMC present difficulties to appropriately catch HF model (Figures \ref{fig:1D_a=2_plots} and \ref{fig:1D_a=4_plots}). For $a=2$, AR1 presents a degradation of 4\% in terms of RMSE compared to GP HF and LMC improves by 44\% GP HF RMSE. NARGP and DGP enable to improve by respectively 86\% and 80\% the RMSE compared to GP HF, despite the weak degree of correlation between LF and HF models in this case.

Another interesting aspect is the trade-off between the prediction accuracy (characterized by R2 and RMSE) and the likelihood of explaining the HF model using a multi-fidelity technique (measured by MNLL). Compared to R2 or RMSE, MNLL also takes into account the uncertainty model quality associated to the prediction. In case $a=4$, NARGP tends to provides a better prediction accuracy (with a mean RMSE of $2.63 \times 10^{-1}$) compared to DGP (mean RMSE of $2.77 \times 10^{-1}$) but DGP provides a better MNLL indicator ($4.7\times 10^{-1}$ compared to $2.1\times 10^{2}$), meaning that NARGP underestimates (in this case, as it can be seen in Figure \ref{fig:1D_a=4_plots}) the variance associated to the prediction. The variance associated to the prediction is a key element in many applications using GP-based modeling such as Bayesian optimization \cite{jones1998efficient} , reliability analysis \cite{balesdent2013kriging,echard2013combined} or multi-fidelity model refinement \cite{le2013multi}. Therefore, depending on the use of the GP-based multi-fidelity model, a less accurate surrogate but with a more precise uncertainty model might be preferred. 

The ability of the GP-based multi-fidelity techniques to model HF behavior is dependent on the relation between the considered model fidelities (linearity, degree of correlation, \textit{etc}.) and also on the number of samples available. For instance, due to a strong correlation between LF and HF models in case $a=1$ ($R2$ of $0.97$) and the simplicity of the HF behavior (sinusoidal oscillations), the multi-fidelity models are able to perfectly capture the HF model ($R2$ of at least $0.999$ on average). In case $a=3$, using the same number of samples, the best multi-fidelity model (in this case DGP), presents a $R2$ of $0.986$ on average, due to a lower correlation between LF and HF and a more complex HF behavior. This tendency even increases for the most complex case $a=4$ where the best multi-fidelity model (in this case NARGP) has a $R2$ of $0.925$ on average. This translates  the need of more HF data when the relationship between the fidelities are complex (non-linear, weak correlation).  
To study the influence of the number of HF samples and the problem dimension, an analytical test case with varying dimension is presented in the next section.

\subsubsection{Influence of the dimension and of the number of high-fidelity samples}\label{Sect_influence_nb_pts_dim}

\begin{figure}[!h]
\begin{center}
\includegraphics[width=1\linewidth]{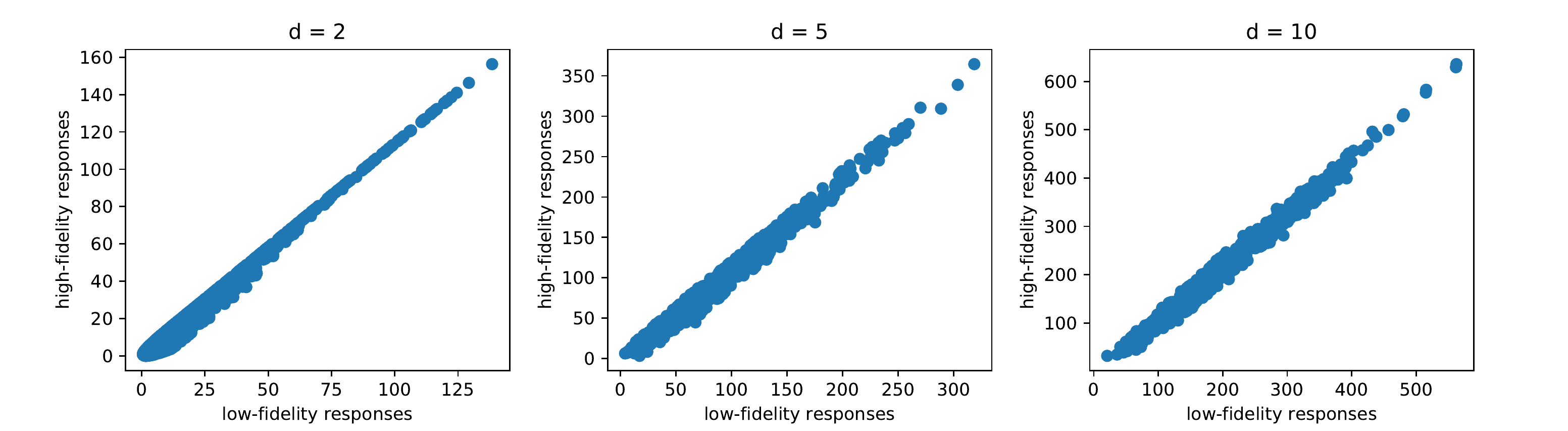}
\caption{Scatter plot of low-fidelity and high-fidelity responses for different dimensions}
\label{fig:10D}
\end{center}
\end{figure}

\begin{figure}[!h]
\begin{center}
\includegraphics[width=.75\linewidth]{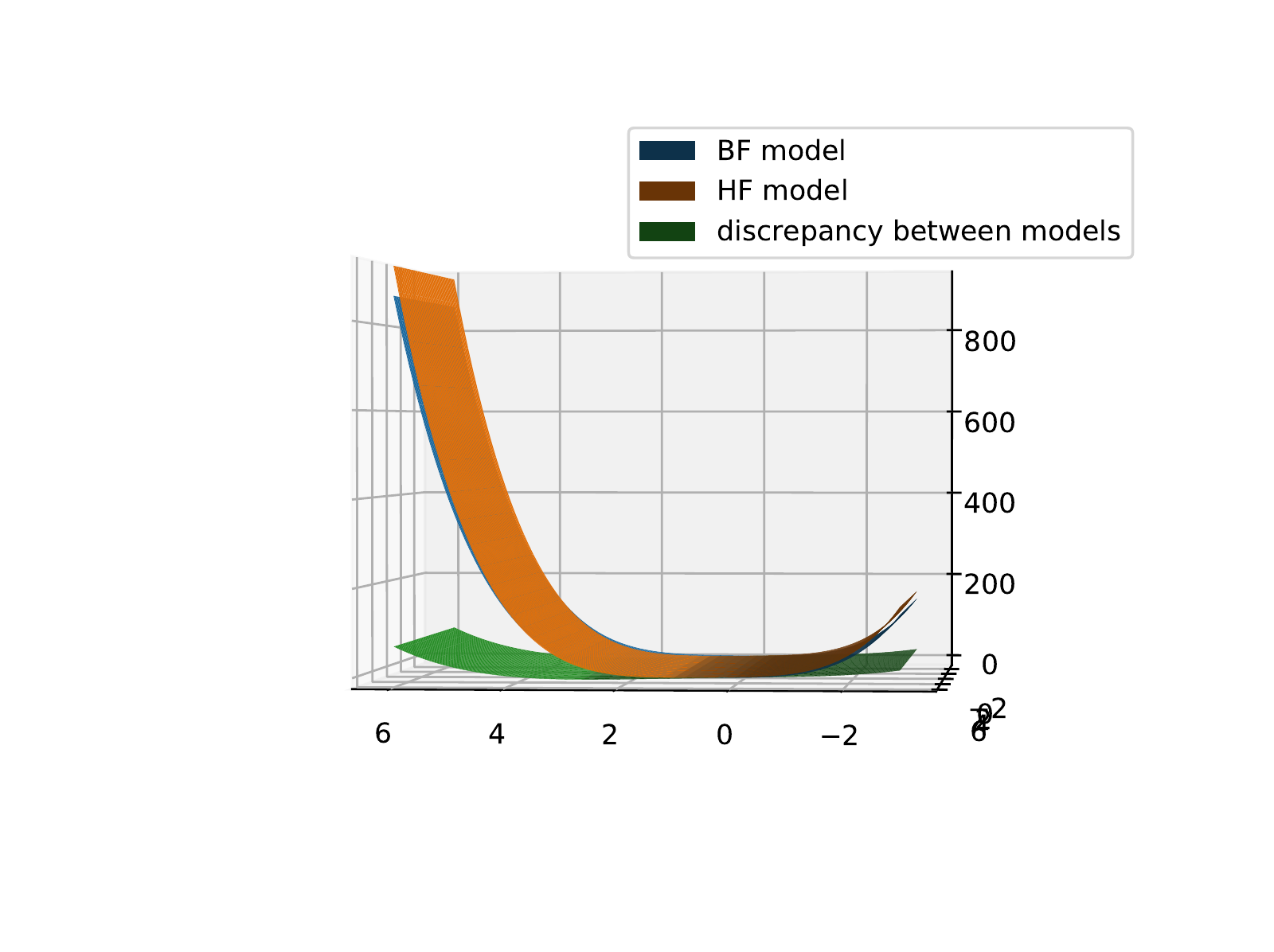}
\caption{BF responses (blue), HF responses (orange) and discrepancy between the two models (green) for the 2D case}
\label{fig:responses_models}
\end{center}
\end{figure}

The second considered test case is derived from \cite{cai2017metamodeling} and is a multidimensional problem defined as follows:
\begin{eqnarray}
f_{hf}(\mathbf{x}) &=&\displaystyle \sum_{i=1}^{d-1}\left((x_{i+1}^2-x_i)^2+(x_i-1)^2\right)\\
f_{lf}(\mathbf{x})&=&\displaystyle \sum_{i=1}^{d-1}\left(0.9x_{i+1}^4+2.2x_i^2-1.8x_ix_{i+1}^2+0.5\right)
\end{eqnarray}
with $d$ the dimension and $x_i \in[-3,3]$, $i=1,\dots,d$. This test case has been used to assess the performance of the different models when varying the dimension of the problem and the number of available HF samples. For that purpose, 2D problem, 5D problem and 10D problem have been considered. The figure \ref{fig:10D} describes a scatter plot of the relationship between the LF and the HF models obtained for the different dimensional problems. As it can be seen, the relationship may be approximated by a linear tendency between the two models. Moreover, the two models are quite close all over the domain of definition as depicted in Figure \ref{fig:responses_models} ($R2$ of approximately 0.95 between the models for all the different dimensions of the input space).

\begin{table}[h!]
\scriptsize
\begin{center}
\caption{Summary of the results obtained on the varying dimension test case}
\label{Table_results_varying_dim}
\begin{tabular}{|c|c|c|c|c|c|c|}
\hline
\multirow{2}{*}{Function}&\multirow{2}{*}{Method}&R2 (std)&\multirow{2}{*}{RMSE (std)}&\multirow{2}{*}{MNLL (std)}&Evolution of &DOE size\\
&&&&&RMSE wrt GP HF& (LF, HF)\\
\hline
\multirow{12}{*}{Var dim (2)}&\multirow{2}{*}{GP HF}&-0.181(0.597)&1.058(0.247)&6.653e4 (1.518e5)&-&40, 4\\
&&0.307(0.295)&0.815 (0.170)&1.994e3(6.686e3)&-&40, 8\\\cline{2-6}
&\multirow{2}{*}{LMC}&0.837(0.378)&0.241(0.324)&50.5 (94.0)&$-77\%$&40, 4\\
&&
0.950(0.176)&1.218e-1(0.186)&1.095e+2(3.100e+2)&$-85\%$&40, 8\\\cline{2-6}
&\multirow{2}{*}{AR1}&0.986(0.014)&0.095(0.071)&2.834e3 (1.098e4)&$-91\%$&40, 4\\
&&1.000(0.000)&3.299e-4(1.700e-4)&-7.071(0.447)&$-99.95\%$&40, 8\\\cline{2-6}
&\multirow{2}{*}{NARGP}&0.760(0.229)&0.411(0.265)& 8.697e2 (2.124e3)&$-61\%$&40, 4\\
&&0.999(0.005)&8.762e-3(3.407e-2)&-5.083(1.341)&$-98.9\%$&40, 8\\\cline{2-6}
&\multirow{2}{*}{NARGP*}&0.758(0.224)&0.418(0.258)&9.855e2  (2.588e3)&$-60\%$&40, 4\\
&&0.999(0.002)&9.951e-3(2.618e-2)&-5.065(1.161)&$-98.8\%$&40, 8\\\cline{2-6}
&\multirow{2}{*}{DGP}&0.907(0.195)&0.214(0.218)&8.706 (3.937e+1)&$-80\%$&40, 4\\
&&1.000( 0.000)&1.531e-2(8.209e-3)&-2.807 (0.3152)&$-98.2\%$&40, 8\\\cline{2-6}
\hline
\multirow{18}{*}{Var dim (5)}&\multirow{3}{*}{GP HF}&-0.302(0.312)&1.133(1.339e-1)&1.479e+4(3.743e+4)&-&100, 5\\
&&-0.287(0.394)&1.124(1.563e-1)&1.134e+2(1.744e+2)&-&100, 10\\\
&&-0.150(0.260)&1.065(1.200e-1)&2.950e+1(7.267e+1)&-&100, 20\\\cline{2-6}
&\multirow{3}{*}{LMC}&0.800(0.081)&4.390e-1(8.656e-2)&8.560e-1(5.291e-1)&$-61\%$&100, 5\\
&&0.861(0.021)&3.721e-1(2.834e-2)&6.066e-1(1.978e-1)&$-67\%$&100, 10\\\
&&0.881(0.030)&3.425e-1(4.273e-2)&5.855e-1(3.193e-1)& $-68\%$&100, 20\\\cline{2-6}
&\multirow{3}{*}{AR1}&0.791(0.088)&4.495e-1(8.449e-2)&3.313(6.088)& $-60\%$&100, 5\\
&&0.856(0.029)&3.777e-1(3.812e-2)&1.164(6.617e-1)&$-66\%$&100, 10\\\
&&0.871(0.045)&3.551e-1(5.750e-2)&7.309e-1(6.224e-1)&$-66\%$&100, 20\\\cline{2-6}
&\multirow{3}{*}{NARGP}&0.229(0.399)&8.519e-1(2.121e-1)&6.492e+3(2.820e+4)&$-25\%$&100, 5\\
&&0.575(0.179)&6.378e-1(1.364e-1)&3.068(4.484)&$-43\%$&100, 10\\
&&0.565(0.199)&6.443e-1(1.395e-1)&2.263(1.728)&$-40\%$&100, 20\\\cline{2-6}
&\multirow{3}{*}{NARGP*}&0.658(0.146)&5.729e-1(1.174e-1)&2.215(2.037)& $-49\%$&100, 5\\
&&0.780(0.059)&4.653e-1(5.766e-2)&7.594e-1(2.675e-1)&$-58\%$&100, 10\\\
&&0.797(0.037)&4.484e-1(4.077e-2)&6.912e-1(2.334e-1)&$-58\%$&100, 20\\\cline{2-6}
&\multirow{3}{*}{DGP}&0.297(0.447)&7.965e-1(2.619e-1)& 2.802e+1(1.043e+2)&$-30\%$&100, 5\\
&&0.563(0.331)&6.227e-1(2.210e-1)&2.072(4.555)&$-44\%$&100,  10\\\
&&0.744(0.134)&4.905e-1(1.227e-1)&8.727e-1(8.748e-1)&$-54\%$&100, 20\\\cline{2-6}
\hline
\multirow{18}{*}{Var dim (10)}&\multirow{3}{*}{GP HF}&-0.514(0.237)&1.227(9.475e-2)&1.219e+4 (3.648e+4)&-&200, 10\\
&&-0.377(0.267)&1.168(0.1115)&7.623(3.817)&-&200, 20\\
&&-0.211(0.165)&1.098(7.286e-2)&3.189(1.035)&-&200, 40\\\cline{2-6}
&\multirow{3}{*}{LMC}&0.773(0.070)&0.4726(6.328e-2)&0.7221 (0.1781)&$-61\%$&200, 10\\
&&0.758(0.097)&0.4848(8.383e-2) &0.8239(0.2542)&$-58\%$&200, 20\\
&&0.809(0.018)&0.4367(2.077e-2)& 0.7735(0.1054)&$-60\%$&200, 40\\\cline{2-6}
&\multirow{3}{*}{AR1}&0.744(0.050)&0.5038(4.683e-2)& 0.9754 (0.3635)&$-59\%$&200, 10\\
&&0.754(0.035)&0.4943(3.494e-2)&0.8866(0.2235)&$-58\%$&200, 20\\
&&0.778(0.025)&0.4702(2.699e-2)&0.8136(0.1341)&$-57\%$&200, 40\\\cline{2-6}
&\multirow{3}{*}{NARGP}&0.308(0.367)&0.8108(0.1858)&4.844e+1 (2.002e+2)&$-33\%$&200, 10\\
&&0.432(0.142)&0.7479(9.284e-2)&1.894(8.369e-1)&$-36\%$&200, 20\\
&&0.443(0.169)&0.7383(0.1071)&1.712(0.6647)&$-32\%$&200, 40\\\cline{2-6}
&\multirow{3}{*}{NARGP*}&0.695(0.033)&0.5514(2.965e-2) &0.9676 (0.2276)&$-55\%$&200, 10\\
&&0.708(0.030)&0.5399(2.769e-2)&0.7918(8.719e-2)&$-53\%$&200, 20\\
&&0.704(0.042)&0.5424(3.674e-2)&0.8245(0.1204)&$-50\%$&200, 40\\\cline{2-6}
&\multirow{3}{*}{DGP}&0.122(0.440)&0.9085(0.2301)&6.641 (8.387)&$-26\%$&200, 10\\
&&0.236(0.433)&0.8466 (0.2182)&2.019(2.052)&$-28\%$&200, 20\\
&&0.574(0.129)&0.6.453(9.562e-2)&0.9781(0.1397)&$-41\%$&200,40\\
\hline
\end{tabular}
\end{center}
\end{table}

A first analysis of the results for the varying dimension test case (Table  \ref{Table_results_varying_dim}) highlights that by increasing the dimension, even if the number of data samples for the LF and HF models are kept proportional to the dimension, improvement of the RMSE for the multi-fidelity methods with respect to the GP HF tends to decrease as the dimension increases. This is due to the classical curse of dimensionality, meaning that when the problem dimension increases, the volume of the design space increases so fast that the available LF and HF data become sparse, resulting in a difficulty to capture the variations of the HF model. For instance, in dimension $2$, all the methods improve by at least $85$\% the RMSE compared to GP HF in case $8$ HF data samples are considered. In dimension $10$, the maximal improvement of the RMSE compared to GP HF is only of $61$\% even if the function behavior is similar for all the dimension problems. 
Similarly, the add of HF data samples (still proportional to the dimension) has a stronger impact for low dimension problems than for high dimension problems. Indeed, in dimension $2$, by doubling the number of HF data, LMC improves from 77\% to 85\% the RMSE compared to GP HF which results in 8\% of difference, whereas in dimension $5$ the difference decreases to 6\% , and it even deteriorates by -3\% in dimension $10$. This phenomenon is observed for all the multi-fidelity techniques and is even stronger for non-linear multi-fidelity methods. For instance for DGP, by doubling the number of HF data, in terms of RMSE improvement compared to GP HF, in dimension $2$ it results in a difference of $18$\%, whereas it decreases to $14$\% in dimension $5$ and to $2$\% in dimension 10.

\begin{figure}[!h]
\begin{center}
\includegraphics[width=.5\linewidth]{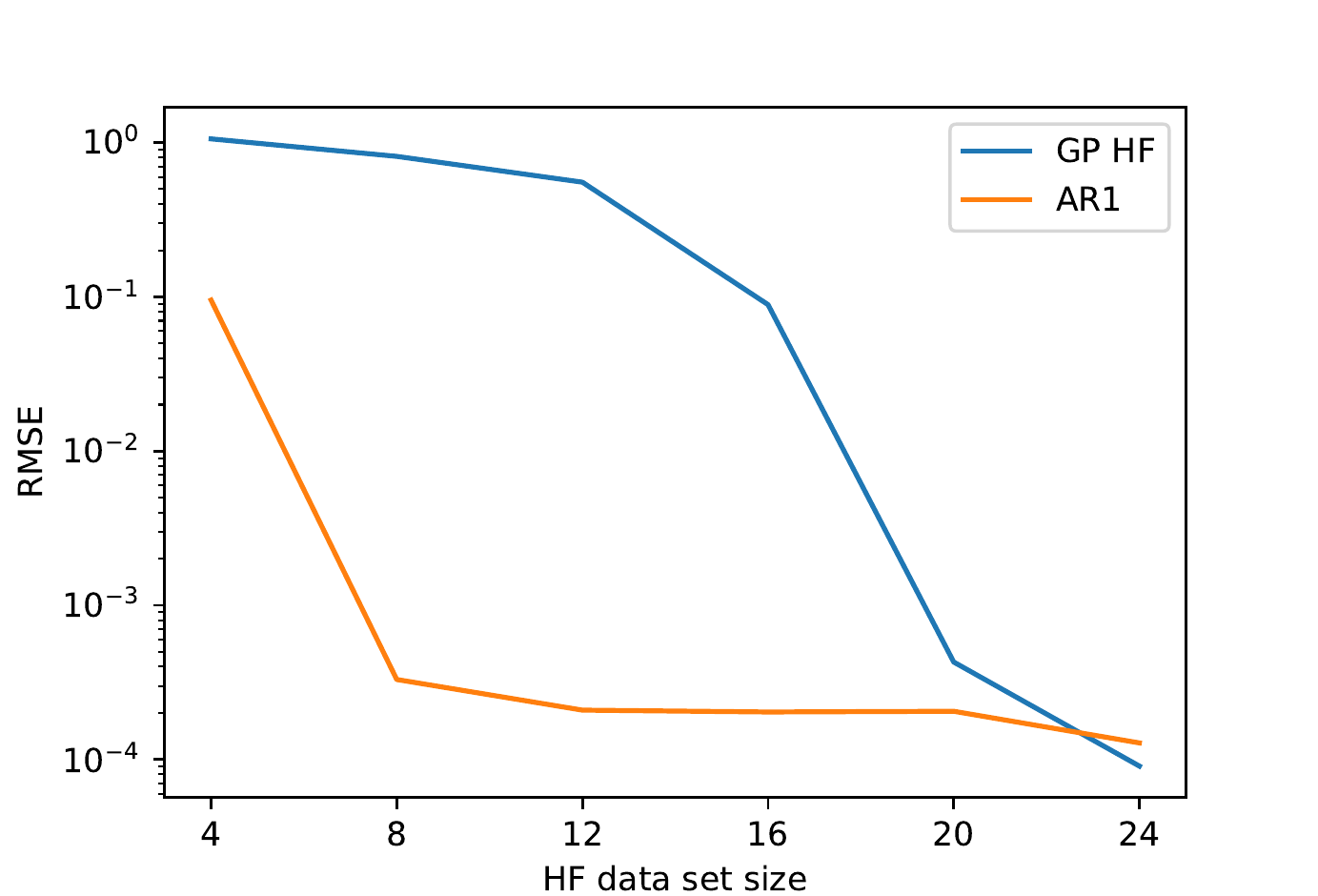}
\caption{Variation of RMSE for GP HF and AR1 models with respect to the size of HF dataset, in case $dim = 2$ (mean over 20 repetitions)}
\label{fig:variation_rmse}
\end{center}
\end{figure}

In this test case, independently of the dimension and the number of HF samples (1, 2 or 4 times the problem dimension), all the multi-fidelity techniques improve the prediction accuracy compared to the only use of HF data with GP HF. This is due to the limited number of HF data samples and the use of LF information to improve the prediction of the behavior of the HF model. However, when the number of HF samples increases, the relative improvements of the RMSE of the multi-fidelity techniques with respect to GP HF decreases. For instance, in Figure \ref{fig:variation_rmse}, the variations of RMSE for GP HF and AR1 models as a function of the number of HF dataset is represented in the case of $dim=2$. It can be seen that, as long as less than 23 HF data samples are used, AR1 method provides a better RMSE than GP HF, however the addition of more HF data samples do not allow to futher improve the multi-fidelity model with respect to single GP model of the HF dataset. Therefore, when it is possible to get enough high-fidelity samples
to construct an accurate single fidelity surrogate model, the question of whether a multi-fidelity surrogate will offer a substantial cost reduction for comparable accuracy is essential. For more information on that subject, please refer to \cite{giselle2019issues}. 

In addition, for the varying dimension test case, independently of the dimension, AR1 and LMC tend to outperform NARGP and DGP. This is due to the fact that the linear relationship that exists between the LF and the HF models is better captured by the linear multi-fidelity techniques. The difference of improvements between linear and non-linear multi-fidelity techniques increases with the dimension of the problem. For instance, in case $dim=2$, the difference between AR1 and NARGP for RMSE improvements compared to GP is of $1.15\%$ (improvements of $99.95$\% for AR1 and of $98.8$\% for NARGP* for $10$ samples in HF data). The gap increases to $5.0$\% in dimension $10$ between LMC and NARGP* (improvements of $58$\% for AR1 compared to $53$\% for NARGP*). 
Except for the case $dim=2$, AR1 and LMC tends to provide similar results in prediction accuracy with LMC having better MNLL metric results.

With the increase of dimensions and of the number of multi-fidelity samples, the training time increases especially for DGP technique. Indeed, to keep a joint nested GP architecture during the inference, a challenging optimization problem has to be solved involving thousand of hyperparameters compared to the alternative mutli-fidelity models (in the order of the dozen of hyperparameters) resulting in a more time consuming task. For example, for the 2D test case with  40 samples in the LF dataset and 8 samples in the HF dataset, the number of hyperparameters that have to be trained is $4$ for GP HF, $9$ for AR1, $10$ for LMC, $13$ for NARGP and $1074$ for MF-DGP. One can notice that even for this low dimensional test case, the DGP model involves much more hyperparameters to be trained with respect to the other models. One can notice that this number of hyperparameters directly depend on size of the DoE  for this model (induced variables). It is not the case for the other models.

%

\subsection{Structural design problem}\label{Section_cantilever}
\subsubsection{Problem definition}
\begin{figure}[!h]
\begin{center}
\includegraphics[width=0.5\linewidth]{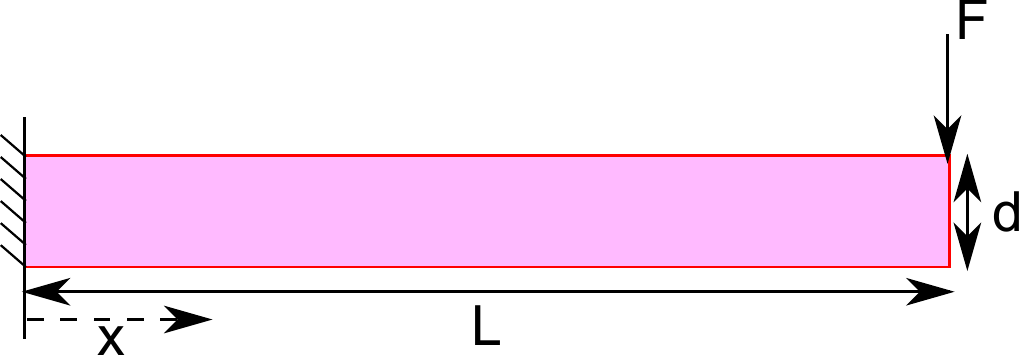}
\caption{Cantilever beam analytical problem, constraint and force}
\label{Cantilever_analytical_constraints_forces}
\end{center}
\end{figure}

\begin{figure}[!h]
\begin{center}
\includegraphics[width=0.5\linewidth]{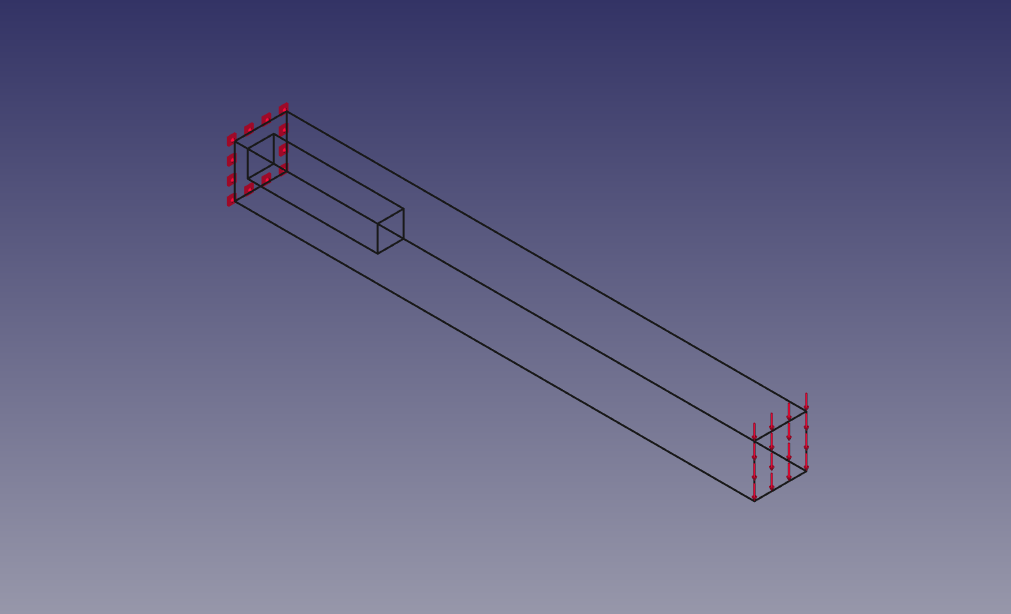}
\caption{Cantilever beam with a hole, constraint and force}
\label{Cantilever_constraints_forces}
\end{center}
\end{figure}

A structural design problem derived from the classical cantilever beam problem is considered as a first engineering application. It consists of a two-fidelity problem, in which the lower fidelity is an analytical estimation of the maximal von Mises (VM) constraint within the cantilever beam (Figure \ref{Cantilever_analytical_constraints_forces}) and the high-fidelity model is the estimation of the maximal VM using a Finite Element Method (FEM) of a modified cantilever beam which includes a rectangular hole at its basis (Figure \ref{Cantilever_constraints_forces}).

A generic steel material is considered with a Young Modulus of $210$ GPa, a Poisson ratio of $0.3$ and a density of $7900 kg/m^3$. Three main parameters are considered (Figure \ref{Cantilever_analytical_constraints_forces}), the length of the beam (L), the length of the square cross-section (d) and the force applied to the extremity of the beam (F). The VM stress is given by the following equation:
\begin{equation}
\sigma_{VM} = \sqrt{(\sigma_{ax} + \sigma_b)^2 + 3\tau_{sh}^2}
\end{equation}
where $\sigma_{ax}$ is the axial stress, $\sigma_b$ the bending stress and $\tau_{sh}$ the shear stress. For the classical cantilever beam problem, the maximal VM is reached at the basis of the beam (meaning at $x=0$ on Figure \ref{Cantilever_analytical_constraints_forces}). At the basis, the axial stress is null, the shear stress is given by $\tau_{sh} = \frac{F}{d^2}$ and the bending stress is equal to $\sigma_b = \frac{6F\times L}{d^3}$.
Therefore, given the parameters $F,\; L$ and $d$, it is possible to easily estimate the maximal VM within the beam.

\begin{figure}[!h]
\begin{center}
\includegraphics[width=0.5\linewidth]{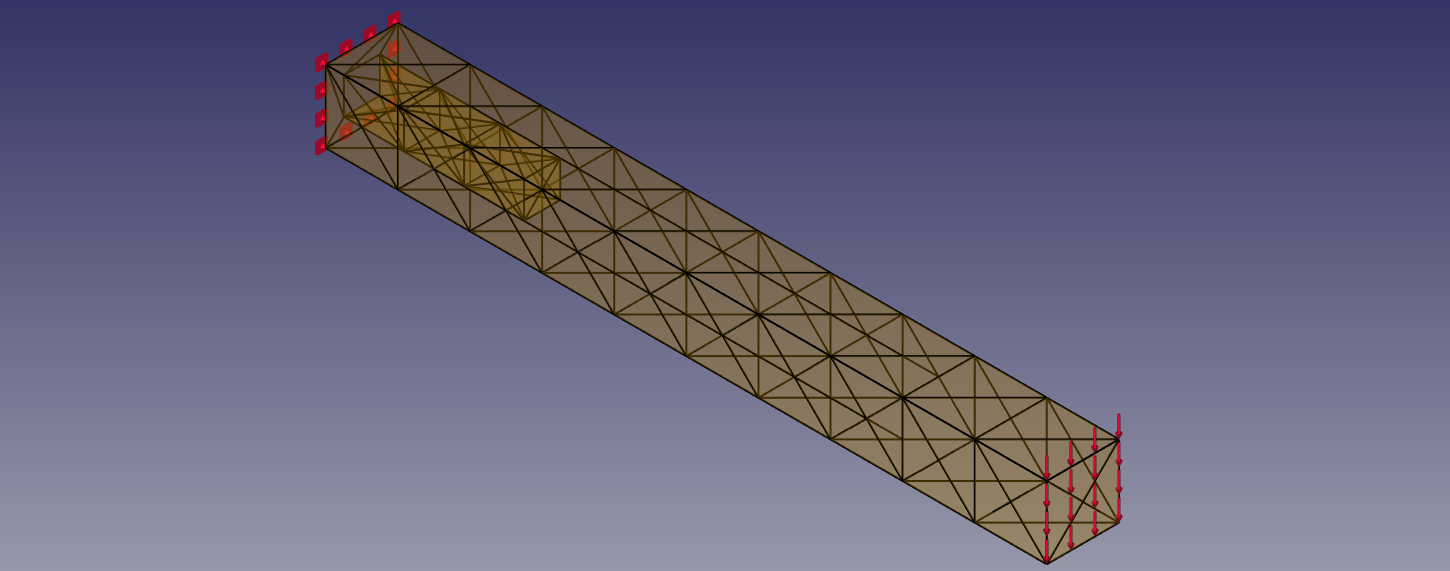}
\caption{Cantilever beam with a hole, associated mesh}
\label{Cantilever_constraints_forces_mesh}
\end{center}
\end{figure}

\begin{figure}[!h]
\begin{center}
\includegraphics[width=0.5\linewidth]{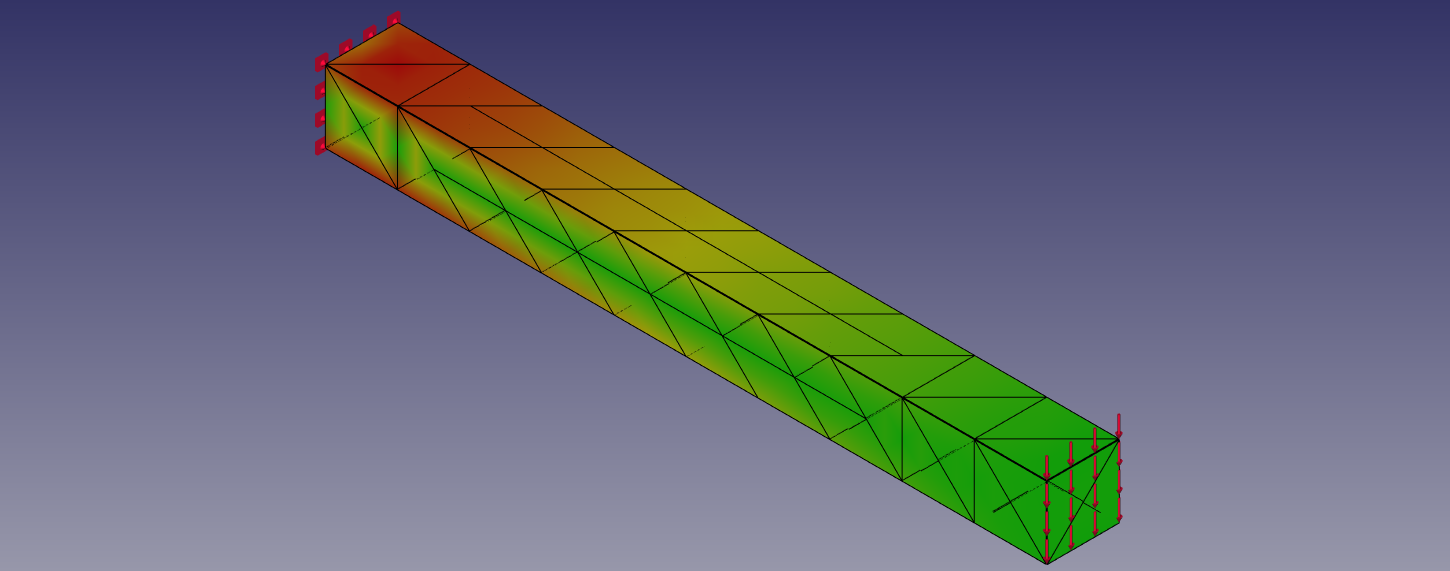}
\caption{Cantilever beam with a hole, von Misses constraints}
\label{Cantilever_constraints_forces_VM_results}
\end{center}
\end{figure}

For the high fidelity model, a rectangular cantilever beam with a rectangular hole at the basis is considered (Figure \ref{Cantilever_constraints_forces}). No analytical expression is available for such a case and a FEM is used. A 3D Freecad model \cite{falck2012freecad} is constructed defining the beam, the rectangular hole, the constrained face, and the surface force applied to the extremity face (Figure \ref{Cantilever_constraints_forces_VM_results}). Then, using GMSH meshing tool \cite{geuzaine2009gmsh}, a mesh of the cantilever is defined for FEM calculation (Figure \ref{Cantilever_constraints_forces_mesh}).

Finally, using CalculiX solver \cite{wittig2013calculix}, the maximal von Mises stress is estimated over the defined mesh (Figure \ref{Cantilever_constraints_forces_VM_results}).

The multi-fidelity modeling problem consists in using a fusion information scheme of the two available fidelity models to approximate the maximal VM within the beam with a rectangular hole using both the analytical LF model and the FEM HF model considering three input design variables: the force, the beam length and the length of the square-cross section. Table \ref{tab:Cantilever_input_design_vb} summarizes the input design variables and their domain of definition.  

\vspace{0.2cm}
\begin{minipage}{\linewidth}
\centering
\captionof{table}{Cantilever beam input design variable definition} \label{tab:Cantilever_input_design_vb} 
\begin{tabular}{|l|c|}
  \hline
  Input variables & Domain of definition  \\
  \hline
  Force (F) & [850., 950.]kN \\
  Beam length (L) & [2., 3.]m \\
  Length of the square-cross section (d) & [0.25, 0.4]m \\
  \hline
\end{tabular}
\end{minipage}

\subsubsection{Results}
\begin{figure}[!h]
\begin{center}
\includegraphics[width=0.5\linewidth]{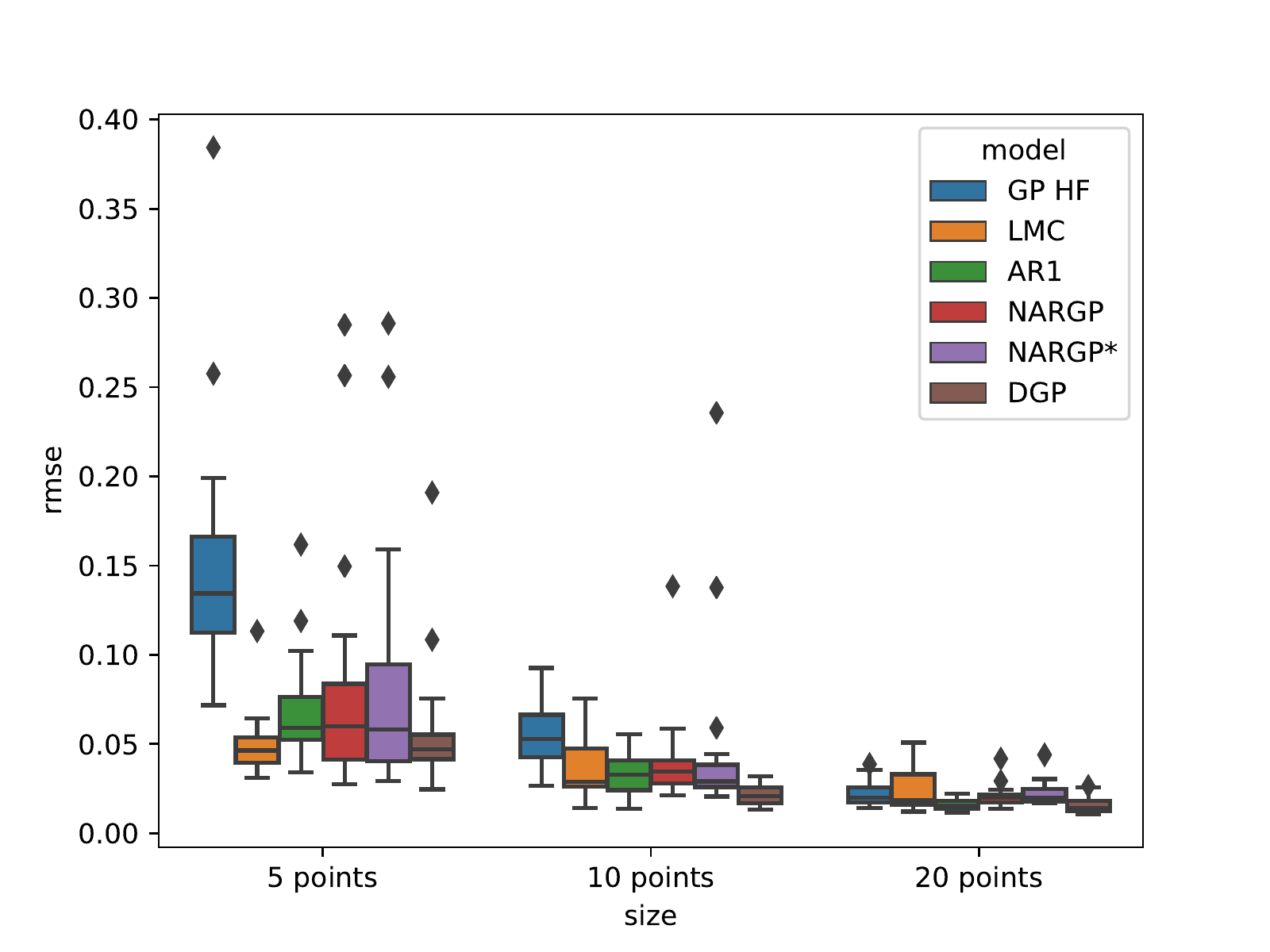}\includegraphics[width=0.5\linewidth]{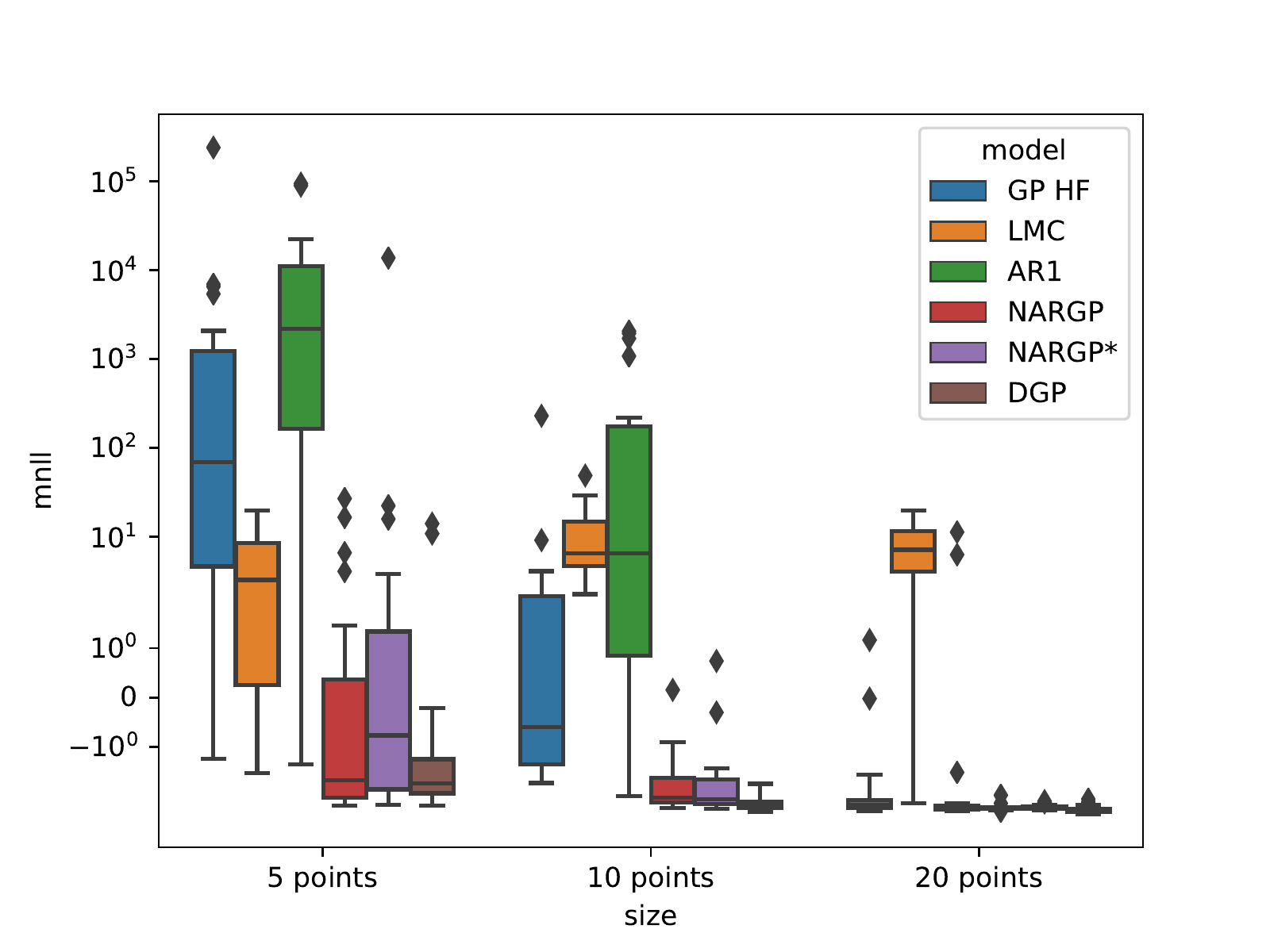}
\caption{Boxplots of RMSE and MNLL for cantilever problem}
\label{Cantilever_boxplot}
\end{center}
\end{figure}

For the cantilever beam problem, the results for the different DoE sizes for the HF dataset in Table \ref{tab:res_cantilever} and in Figure \ref{Cantilever_boxplot}.  The size of the LF dataset is 60. For HF sample of size 5 and 10, except for NARGP* with 10 samples in HF, the multi-fidelity methods improve the prediction compared to GP HF, meaning the LF information enables to improve the prediction capability. For HF sample of size 5, R2 increases from $0.742$ for GP HF to $0.975$ for LMC which is a substantial improvement. Moreover, the robustness to the experimentation repetitions is increased for all the methods compared to GP HF thanks to the information given by the LF dataset. For instance, for $5$ HF samples, GP HF standard deviation of $R2$ is $0.288$, while for multi-fidelity techniques they are all lower from $0.023$ for LMC to $0.202$ for NARGP*. This is due to the fact that compared to GP HF, multi-fidelity techniques use LF information and therefore are less sensitive to the position of HF samples.

\begin{table}[h!]
\scriptsize
\caption{Summary of the results obtained on the Cantilever beam problem}
\begin{center}
\begin{tabular}{|c|c|c|c|c|c|c|}
\hline
\multirow{2}{*}{Function}&\multirow{2}{*}{Method}&\multirow{2}{*}{R2  (std)}&\multirow{2}{*}{RMSE (std)}&\multirow{2}{*}{MNLL (std)}&Evolution of&DOE size\\
&&&&&RMSE wrt GP HF& (LF, HF)\\
\hline
\multirow{18}{*}{Cantilever Beam}&\multirow{3}{*}{GP HF}&0.742(0.288)&1.515e-1(6.749e-2)& 1.315e+4(5.210e+4)&-&60, 5 \\
&&0.970(0.018)&5.440e-2(1.685e-2)&1.197e+1(5.003e+1)&-&60, 10\\
&&0.995(0.003)&2.220e-2(6.925e-3)&-2.149(9.966e-1)&-&60, 20\\\cline{2-6}
&\multirow{3}{*}{LMC}&0.975(0.023)&4.900e-2(1.756e-2)&4.854(5.582)&$-67\%$&60, 5 \\
&&0.986(0.013)&3.568e-2(1.613e-2)&1.215e+1(1.154e+1)& $-34\%$&60, 10\\
&&0.993(0.007)&2.451e-2(1.161e-2)&7.511(6.070)&$10\%$&60, 20\\\cline{2-6}
&\multirow{3}{*}{AR1}&0.947(0.054)&6.830e-2(3.055e-2)&1.405e+4(2.681e+4)& $-55\%$&60, 5 \\
&&0.989(0.007)&3.290e-2(1.113e-2)&3.713e+2(6.861e+2)&$-40\%$&60, 10\\
&&0.998(0.001)&1.596e-2(2.804e-3)&-1.519(3.535)&$-28\%$&60, 20\\\cline{2-6}
&\multirow{3}{*}{NARGP}&0.890(0.200)&8.271e-2(6.941e-2)&1.528(7.293)&$-45\%$&60, 5 \\
&&0.979(0.037)&4.000e-2(2.422e-2)&-1.908(6.527e-1)&$-26\%$&60, 10\\
&&0.996(0.003)&2.078e-2(5.964e-3)&-2.694(1.992e-1)&$-6\%$&60, 20\\\cline{2-6}
&\multirow{3}{*}{NARGP*}&0.879(0.202)&8.822e-2(7.148e-2)&6.886e+2(2.996e+3)& $-42\%$&60, 5 \\
&&0.956(0.116)&4.693e-2(4.987e-2)&-1.928(8.274e-1)&$-14\%$&60, 10\\
&&0.994(0.005)&2.316e-2(8.022e-3)&-2.666e(1.579e-1)&$4\%$&60, 20\\\cline{2-6}
&\multirow{3}{*}{DGP}&0.958(0.072)&5.700e-2(3.548e-2)&-2.859e-1(4.309&$-62\%$&60, 5\\
&&0.995(0.003)&2.146e-2(5.928e-3)&-2.463(3.680e-1)&$-60\%$&60, 10\\
&&0.997(0.002)&1.562e-2(4.719e-3)&-2.849(2.963e-1)&$-30\%$&60, 20\\\cline{2-6}
\hline
\end{tabular}
\end{center}
\label{tab:res_cantilever}
\end{table}

It appears that for small HF sample size, linear multi-fidelity approaches perform better than non-linear methods (R2 of $0.975$ for LMC with respect to  $0.958$ for DGP). This is due to the lack of data to appropriately learn the non-linear mapping between the fidelities. However, for 10 and 20 sample size, DGP presents better results in terms of R2, RMSE and MNLL. Indeed, for HF sample size of 20, AR1 and DGP share the same prediction accuracy (R2 of $0.998$, or improvements of RMSE compared to GP HF of respectively $28$\% and $30$\%), however the uncertainty model associated to the prediction is more accurate for DGP (MNLL of $-2.85$ for DGP compared to $-1.52$ for AR1) as it translates that the likelihood of explaining HF model with DGP is higher than with AR1. Through all the experimentations on the cantiveler beam problem, DGP provides the best MNLL metric which is a valuable capability for a GP-based surrogate model.

\subsection{SSTO trajectory simulation}
\subsubsection{Problem definition}

\begin{figure}[!h]
\begin{center}
\includegraphics[width=0.5\linewidth]{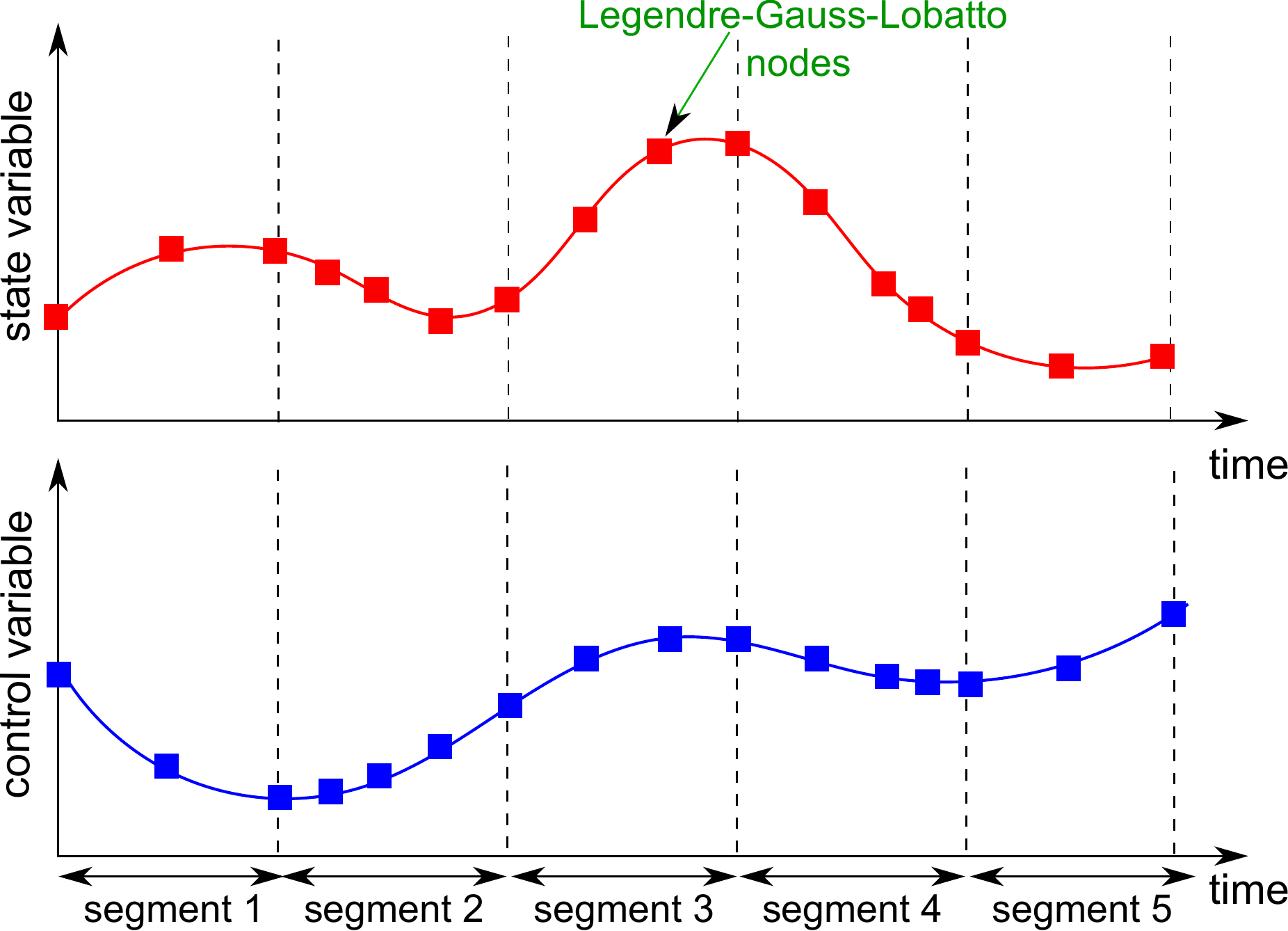}
\caption{Gauss-Lobatto collocation method}
\label{Gauss_lobatto}
\end{center}
\end{figure}

This trajectory problem is based on the "Time-Optimal Launch of a Titan II" example defined by Longuski \textit{et al.} \cite{longuski2014optimal}. It is an optimal control problem which consists in finding the pitch angle profile for a Single-Stage-To-Orbit (SSTO) launch vehicle that minimizes the time required to reach orbit injection under considering a constant thrust. A 2D Cartesian simulation with a planar trajectory, non rotating Earth is considered. The problem is defined by the following equations of motion:
\begin{eqnarray}
\frac{dx}{dt} &=& v_x \\
\frac{dy}{dt} &=& v_y \\
\frac{dv_x}{dt} &=& \frac{T \cos(\theta)- D\cos(\gamma)}{m} \\
\frac{dv_y}{dt} &=& \frac{T \sin(\theta)- D\sin(\gamma)}{m} - g \\
\frac{m}{dt} &=& -\frac{T}{g Isp}
\end{eqnarray}
with $T$ the launch vehicle thrust, $D$ the drag force, $\theta$ the pitch angle, $\gamma$ the flight path angle, $m$ the mass of the launch vehicle, $g$ the gravity acceleration, $Isp$ the engine specific impulse and $v_x, v_y$ the launch vehicle velocity components in the Cartesian space. The target final orbit is a circular orbit at the altitude of 185km. 
The trajectory simulation is carried out using Dymos \cite{falck2019optimal} which is an open-source tool for solving optimal control problems involving multidisciplinary systems. It is built on top of the OpenMDAO framework \cite{gray2019openmdao}. A high order Gauss-Lobatto collocation method \cite{herman1996direct} is used to solve this optimization problem. Gauss-Lobatto is a generalization of the Hermite-Simpson optimization scheme developed by Herman and Conway \cite{herman1996direct}. In this approach, for solving optimal control problem, polynomials are considered to represent the state variable time history over segments (subintervals) of the total time of interest. The polynomial family follows the Gauss-Lobatto rules. Each segment is discretized according to the Legendre-Gauss-Lobatto polynomial nodes (Figure \ref{Gauss_lobatto}). The value of each state variable and each control variable at each state discretization node is a design variable. The higher the number of segments, the higher the accuracy of the optimal control solving but the higher the number of design variables in the optimization problem and therefore the associated computational cost.

For the SSTO problem, two fidelity models are considered. The input space is composed of five design variables: the thrust, the specific impulse, the diameter of the launch vehicle, the initial mass of the vehicle and the coefficient of drag (Table \ref{tab:SSTO_input_design_vb}). The considered output is the fuel burnt mass during the flight.

\vspace{0.2cm}
\begin{minipage}{\linewidth}
\centering
\captionof{table}{SSTO input design variable definition} \label{tab:SSTO_input_design_vb} 
\begin{tabular}{|l|c|}
  \hline
  Input variables & Domain of definition  \\
  \hline
  Thrust (T) & [1800, 2400]kN \\
  Specific impulse (Isp) & [210, 330]s \\
  Launch vehicle diameter (d) & [2.5, 4.4]m \\
  Launch vehicle initial mass ($m_0$) & [120, 124]t \\
  Coefficient of drag ($C_d$) & [0.1, 0.9] \\
  \hline
\end{tabular}
\end{minipage}
\vspace{0.2cm}

The two fidelities are distinguished by the number of segments of the Gauss-Lobatto collocation. The LF model assumes a low number of segments $num_{segments}=4$ corresponding to a discretization scheme enabling fast optimal control solving but limited simulation accuracy. The HF model assumes a higher number of segments $num_{segments}=15$ providing a high accuracy for the trajectory simulation but a more complex and more computationally intensive optimal control problem to be solved.

%

\begin{figure}[h]
\begin{subfigure}{0.5\textwidth}
\includegraphics[width=0.95\linewidth]{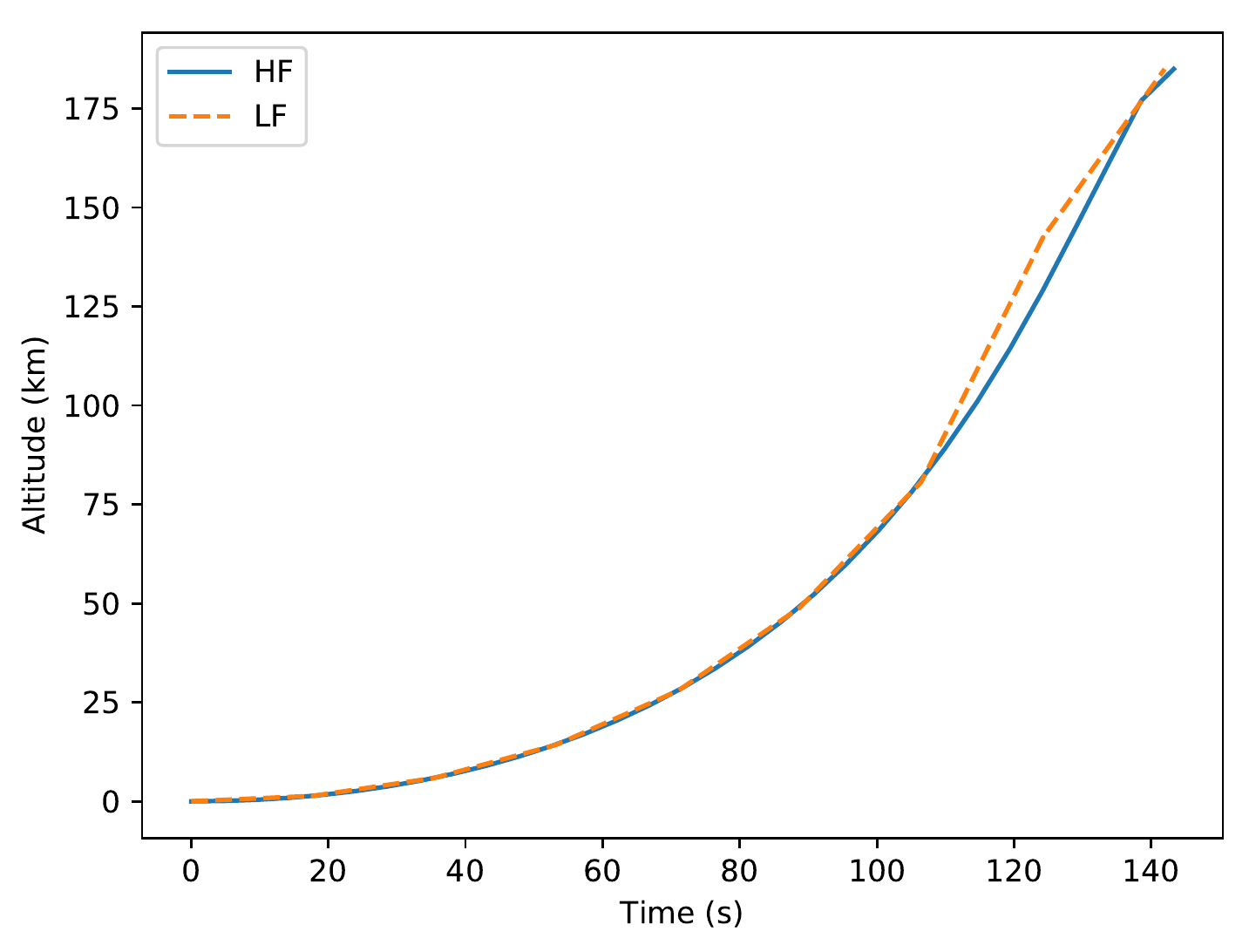} 
\caption{Altitude as a function of time}
\label{SSTO_alt_example}
\end{subfigure}
\begin{subfigure}{0.5\textwidth}
\includegraphics[width=0.95\linewidth, height=5cm]{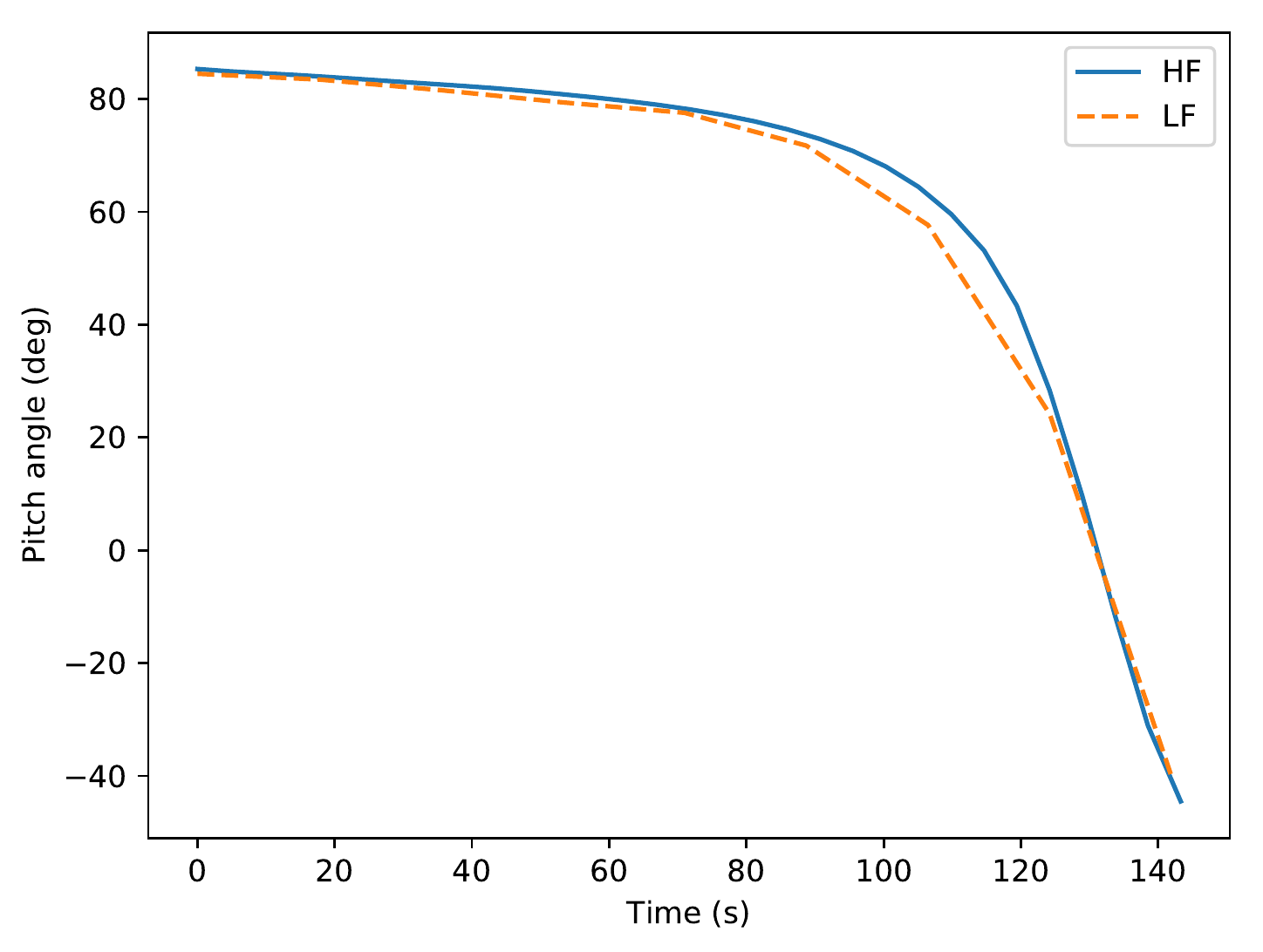}
\caption{Pitch angle as a function of time}
\label{Pitch_SSTO_example}
\end{subfigure}
\caption{Illustrations for SSTO trajectory with Low Fidelity (LF) and High Fidelity (HF) models}
\label{SSTO_traj1}
\end{figure}

\begin{figure}[!h]
\begin{center}
\includegraphics[width=0.5\linewidth]{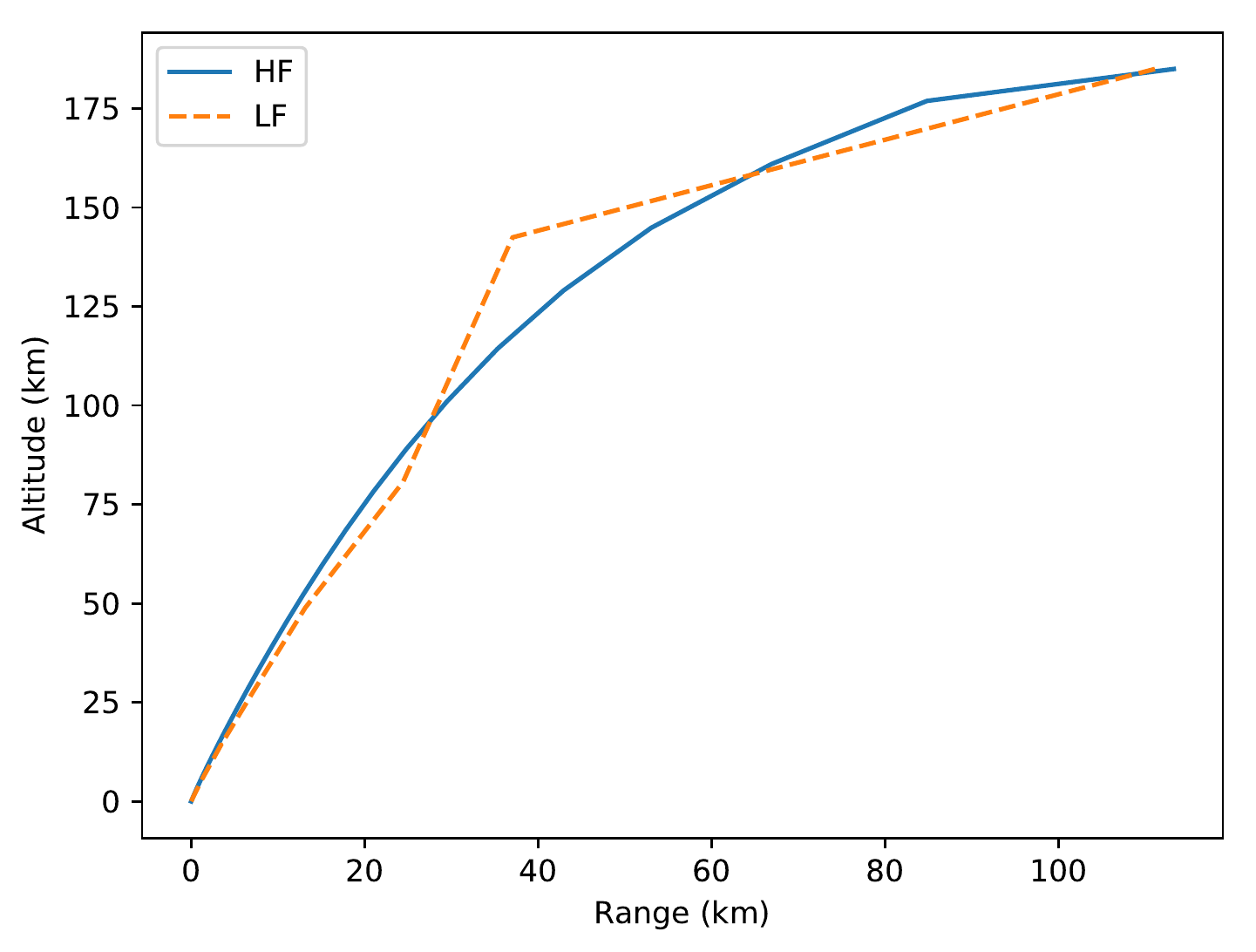}
\caption{Altitude as a function of range for the SSTO trajectory with Low Fidelity (LF) and High Fidelity (HF) models}
\label{SSTO_range_example}
\end{center}
\end{figure}

The difference between LF and HF models are illustrated on Figures \ref{SSTO_alt_example}, \ref{Pitch_SSTO_example} and \ref{SSTO_range_example} representing the altitude as a function of time, the altitude as a function of range and the pitch angle as a function of time. The LF model provides a reasonable approximation of the HF model but with substantial simplification in the trajectory. The limitation on the number of segments is particularly visible for the plot of the altitude as a function of range (Figure \ref{SSTO_range_example}). 
 
\subsubsection{Results}

\begin{figure}[!h]
\begin{center}
\includegraphics[width=0.5\linewidth]{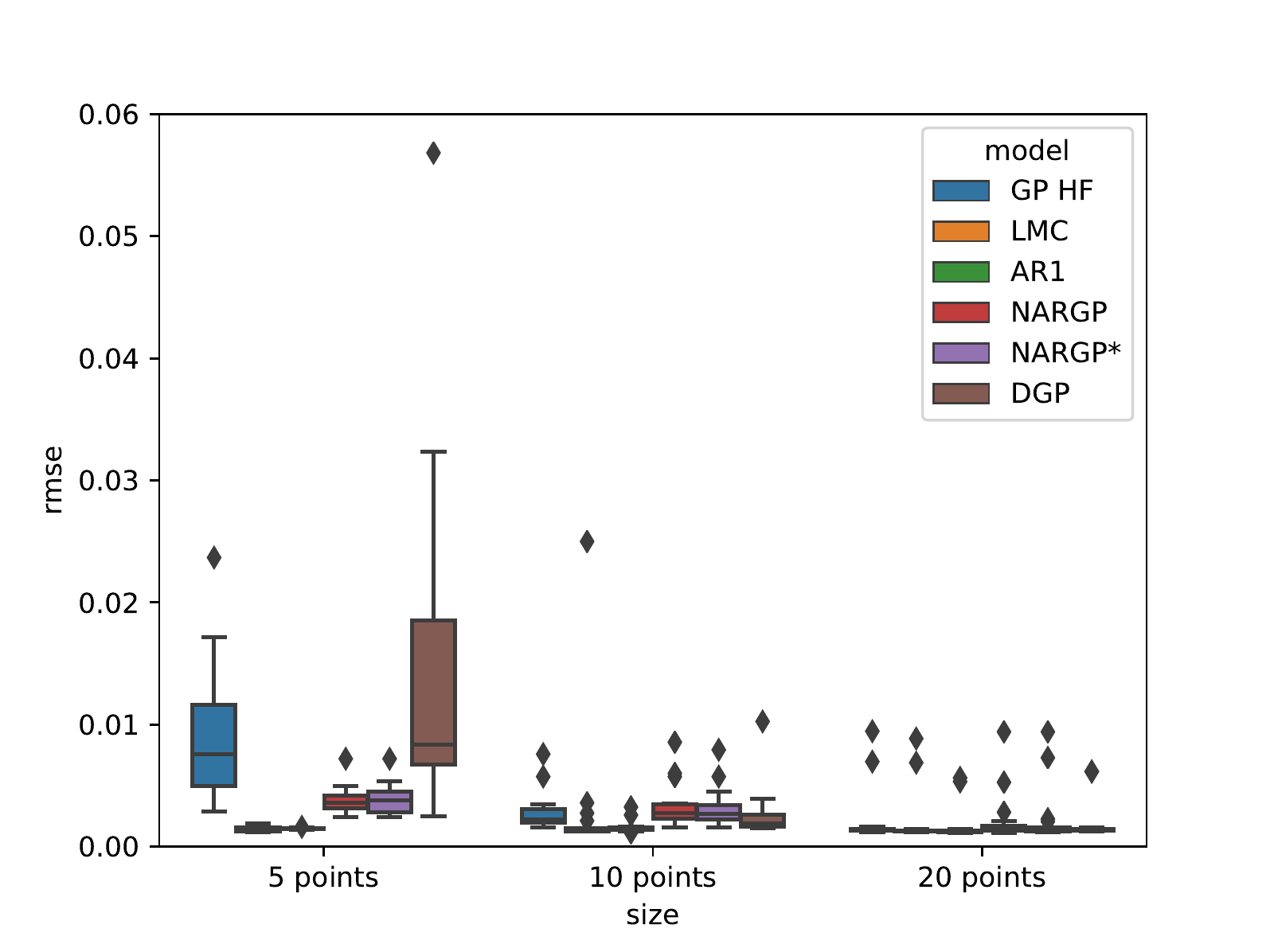}\includegraphics[width=0.5\linewidth]{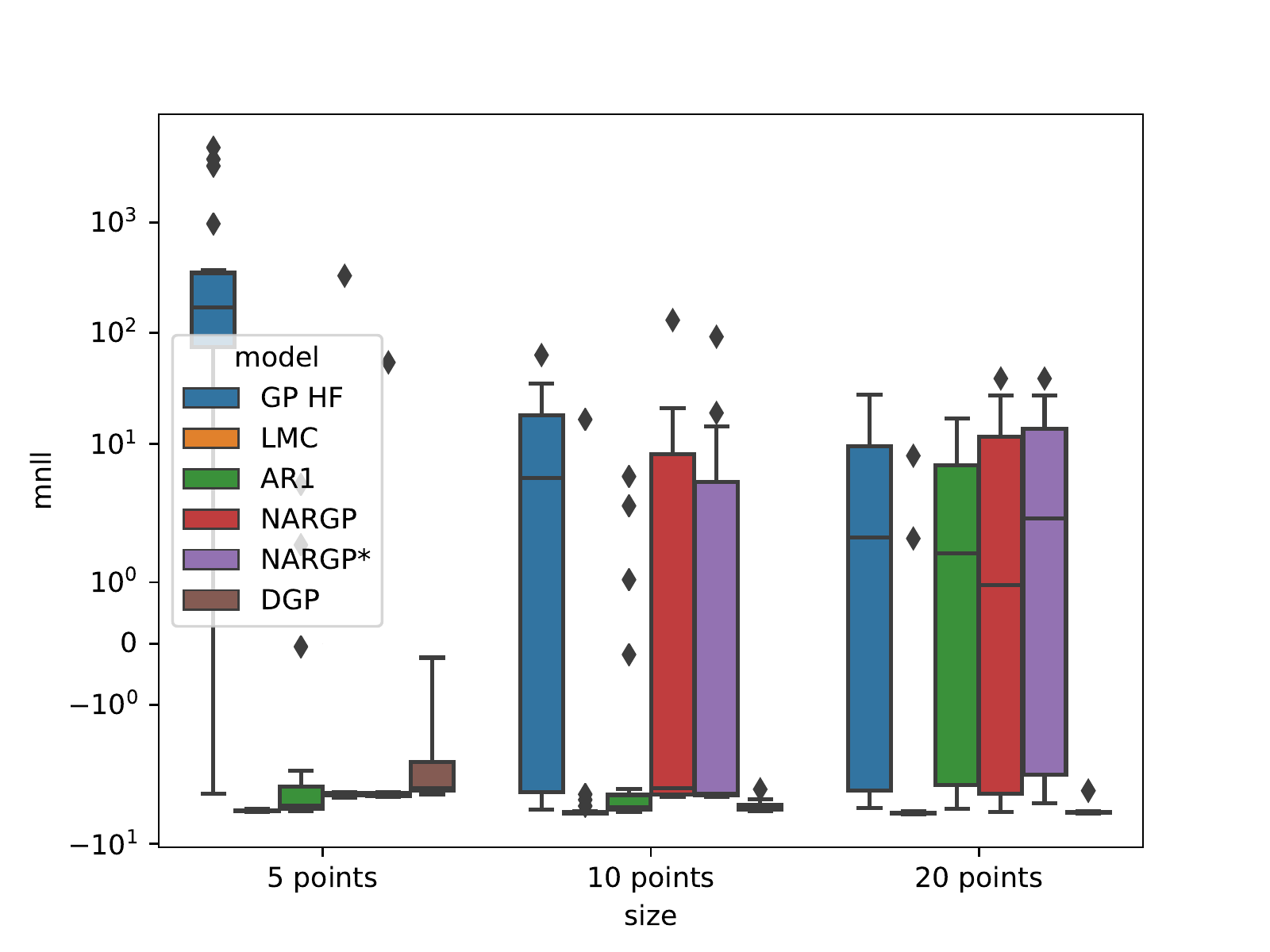}
\caption{Boxplots of RMSE and MNLL for SSTO problem}
\label{SSTO_boxplot}
\end{center}
\end{figure}

Boxplots illustrating the results for SSTO problem are displayed in Figure \ref{SSTO_boxplot}. In addition, results including the different comparison metrics are provided in Table \ref{tab:res_SSTO}. Similarly to the cantilever beam problem, for the SSTO test case, non-linear multi-fidelity techniques perform less accurately compared to linear approaches for a small number of samples in the HF DoE. For instance, for a HF sample size of 5 points, AR1 and LMC provide the same prediction accuracy (R2 of $0.993$) and LMC  provides the best model of prediction uncertainty (MNLL of $-5.03$ for LMC compared to $-3.33$ for AR1). Moreover, DGP approach performs poorly with a R2 of $-7.865$. For HF sample size of $10$ and $20$, both NARGP and NARGP* degrade the RMSE performance compare to GP HF. Furthermore, once enough HF samples are available for DGP, it provides comparable prediction accuracy as linear approaches (R2 of $0.986$ for AR1 compared to $0.982$ for DGP) but with a better likelihood of explaining the HF model thanks to a better uncertainty model (MNLL of $-5.01$ for DGP compared to $2.92$ for AR1). Considering the results with a DoE for HF of 20 samples, this test case is a representative illustration of the trade-off between the prediction accuracy and the quality of the uncertainty model for the prediction. AR1 tends to provide a better prediction against the HF test set, however, the quality of the uncertainty model associated to DGP is better and therefore future use of such a model for optimization, uncertainty propagation or refinement strategies might present some advantages.
Eventually, considering the best multi-fidelity model for each size of HF samples, the addition of HF samples reaches a limit in terms of RMSE improvement compared to GP HF, as for $5$ HF samples the best improvements is of $83$\% while for $20$ HF samples it decreases to $19$\%.

\begin{table}[h!]
\scriptsize
\caption{Summary of the results obtained on the SSTO problem}
\begin{center}
\begin{tabular}{|c|c|c|c|c|c|c|}
\hline
\multirow{2}{*}{Function}&\multirow{2}{*}{Method}&\multirow{2}{*}{R2  (std)}&\multirow{2}{*}{RMSE (std)}&\multirow{2}{*}{MNLL (std)}&Evolution of&DOE size\\
&&&&&RMSE wrt GP HF& (LF, HF)\\
\hline
\multirow{18}{*}{SSTO}&\multirow{3}{*}{GP HF}&0.642(0.425)&9.123e-3(5.248e-3)&7.566e+2(1.367e+3)&-&100, 5 \\
&&0.969(0.041)&2.757e-3(1.442e-3)&1.064e+1(1.748e+1)&-&100, 10\\
&&0.972(0.068)&2.036e-3(2.095e-3)&4.044(8.781)&-&100, 20\\\cline{2-6}
&\multirow{3}{*}{LMC}&0.993(0.002)&1.464e-3(2.187e-4)&-5.027(9.996e-2)&$-83\%$&100, 5 \\
&&0.891(0.438)&2.727e-3(5.141e-3)&-3.973(4.730)&$-1\%$&100, 10\\
&&0.975(0.061)&1.925e-3(2.004e-3)&-4.266(3.153)&$-5\%$&100, 20\\\cline{2-6}
&\multirow{3}{*}{AR1}&0.993(0.001)&1.462e-3(5.668e-5)&-3.326(2.460)&$-83\%$&100, 5 \\
&&0.991(0.007)&1.583e-3(4.667e-4)&-3.227(2.874)&$-42\%$&100, 10\\
&&0.986(0.028)&1.639e-3(1.283e-3)&2.919(6.476)&$-19\%$&100, 20\\\cline{2-6}
&\multirow{3}{*}{NARGP}&0.951(0.032)&3.754e-3(1.051e-3)&1.312e+1(7.274e+1)&$-58\%$&100, 5 \\
&&0.958(0.052)&3.227e-3(1.655e-3)&8.761(2.934e+1)&$+17\%$&100, 10\\
&&0.971(0.069)&2.127e-3(2.115e-3)&7.208(1.143e+1)&$+4\%$&100, 20\\\cline{2-6}
&\multirow{3}{*}{NARGP*}&0.949(0.033)&3.808e-3(1.134e-3)&-7.133e-1(1.262e+1)&$-58\%$&100, 5 \\
&&0.961(0.044)&3.129e-3(1.462e-3)&4.939(2.128e+1)&$+13\%$&100, 10\\
&&0.974(0.062)&2.102e-3(1.896e-3)&5.533(1.192e+1)&$+3\%$&100, 20\\\cline{2-6}
&\multirow{3}{*}{DGP}&-7.865(28.921)&2.651e-2(4.518e-2)&-2.675(9.035e-1)&$190\%$&100, 5 \\
&&0.968(0.071)&2.521e-3(1.893e-3)&-4.640(4.315e-1)&$-8\%$&100, 10\\
&&0.982(0.035)&1.876e-3(1.426e-3)&-5.014(5.717e-1)&$-8\%$&100, 20\\\cline{2-6}
\hline
\end{tabular}
\end{center}
\label{tab:res_SSTO}
\end{table}

\subsection{SSBJ multidisciplinary problem}
\subsubsection{Problem definition}

\begin{figure}[!h]
\begin{center}
\includegraphics[width=0.85\linewidth]{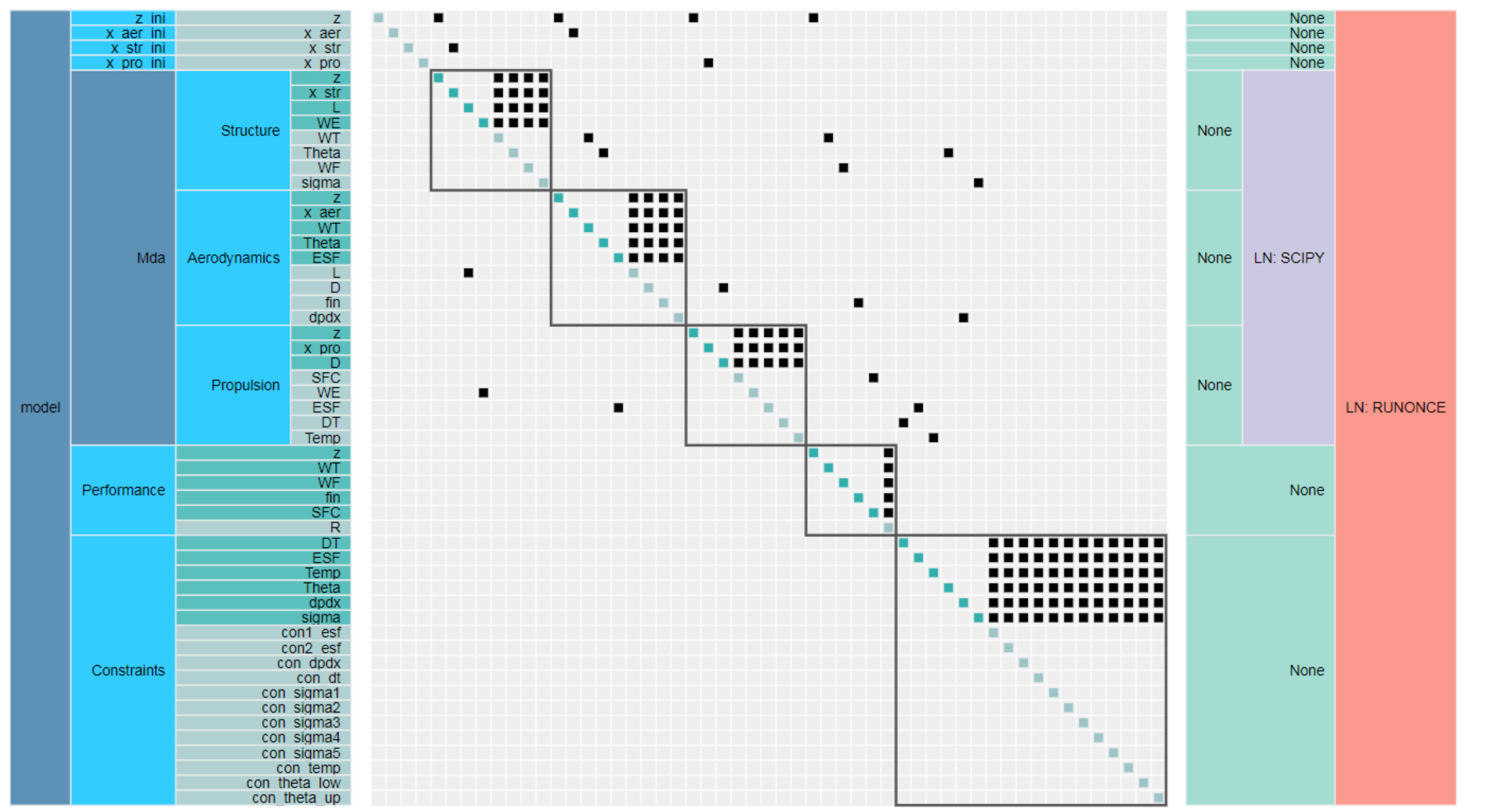}
\caption{N2 chart of the MDA for SSBJ problem}
\label{N2_chart_SSBJ}
\end{center}
\end{figure}

For the third aerospace design application, a multidisciplinary design is considered of a SuperSonic Business Jet (SSBJ) based on the problem defined by Sobieszczanski \textit{et al.} \cite{langley1998bi}. The multidisciplinary analysis is composed of four disciplinary modules: structures, aerodynamics, propulsion and range estimation. All the disciplines are modeled with an analysis level typical for an early conceptual design stage. The aircraft simulation allows to estimate its range through the Breguet range equation. Each discipline implements early design models (analytical formula). The structure discipline computes the stresses undertaken by the wings of the aircraft and the mass of the different components of the vehicle (\textit{e.g.}, fuselage, wing, fuel). It takes as inputs the  definition of the characteristics of the wings (thickness to chord ratio, aspect ratio, sweep angle), the lift coefficient (from the aerodynamics discipline) and the engine mass (from the propulsion discipline). The aerodynamics discipline computes the lift and drag of the vehicle. It takes as inputs the wing characteristics, flight conditions and the size of engine from the other disciplines. The propulsion discipline aims at defining the dimension, mass and consumption of the engine from the flight conditions and drag of the vehicle. Finally, the performance discipline computes the range of the vehicle (R) from the outputs of the other disciplines: the lift over drag ratio (L/D), the engine consumption (SFC), the cruise Mach number (M), the altitude (h) and the weights of the aircraft ($W_T$ and $W_F$) from :
\begin{equation}
R= \frac{M (L/D)661\sqrt{\theta(h)}}{SFC}\text{ln}\left( \frac{W_T}{W_T-W_F}\right)
\end{equation}
This value is considered as the output of the design process for the training of the multi-fidelity surrogate model. For more details on SSBJ simulation, refer to \cite{langley1998bi}. 
The SSBJ problem is simulated using OpenMDAO framework \cite{gray2019openmdao}. 
As the SSBJ is a multidisciplinary problem, it requires a multidisciplinary analysis (MDA) in order to satisfy the coupling consistency between the different disciplines. A N2 chart of the MDA for the SSBJ problem is illustrated in Figure \ref{N2_chart_SSBJ}. This MDA can be performed using Fixed-Point-Iteration, that is an iterative process between the different disciplines. This process is considered as converged when the discrepancy of the output disciplines between two iterations is less than a given tolerance $\epsilon$. The less the tolerance, the higher the accuracy of the response but the higher the duration of the MDA. For that context, two tolerances $\epsilon_{lf} > \epsilon_{hf}$ have been considered to define the two fidelities of the design process. The low-fidelity considers a coarse convergence of the MDA (only one iteration) whereas the high-fidelity considers a very restricted tolerance and requires a dozen of iterations between the disciplines.
The design input parameters are defined in Table \ref{tab:SSBJ_input_design_vb}.

\vspace{0.2cm}
\begin{minipage}{\linewidth}
\centering
\captionof{table}{SSBJ input design variable definition} \label{tab:SSBJ_input_design_vb} 
\begin{tabular}{|l|c|}
  \hline
  Input variables & Domain of definition  \\
  \hline
  Thickness to chord ratio & [0.025, 0.085] \\
  Altitude & [20, 50]km \\
  Mach number & [1.0, 2.0] \\
  Aspect ratio & [1.5, 6.0] \\
  Wing sweep & [20, 70]deg \\
  Wing surface area & [1000, 1750]$m^2$ \\
  \hline
\end{tabular}
\end{minipage}

\subsubsection{Results}

\begin{figure}[!h]
\begin{center}
\includegraphics[width=.5\linewidth]{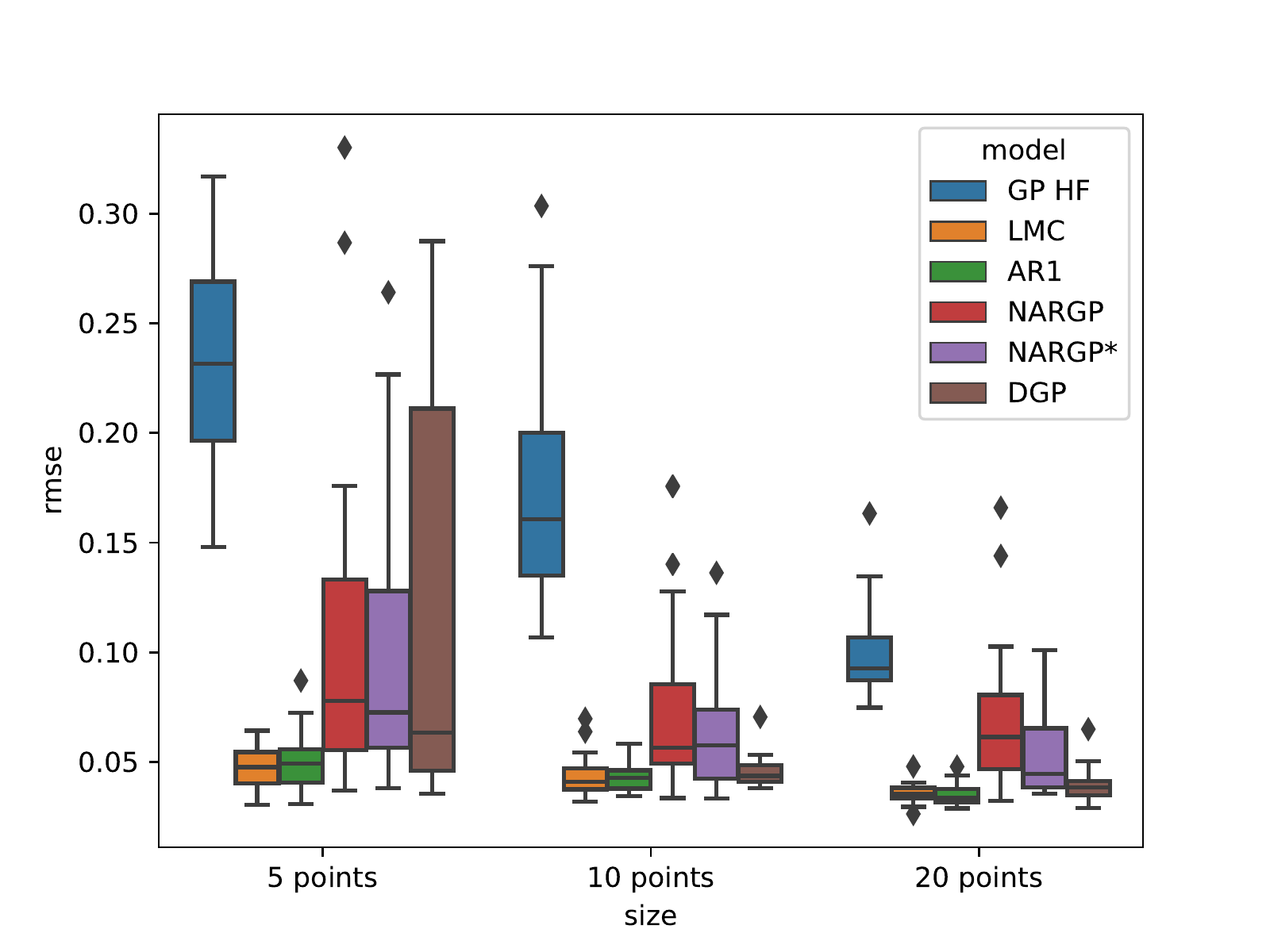}\includegraphics[width=.5\linewidth]{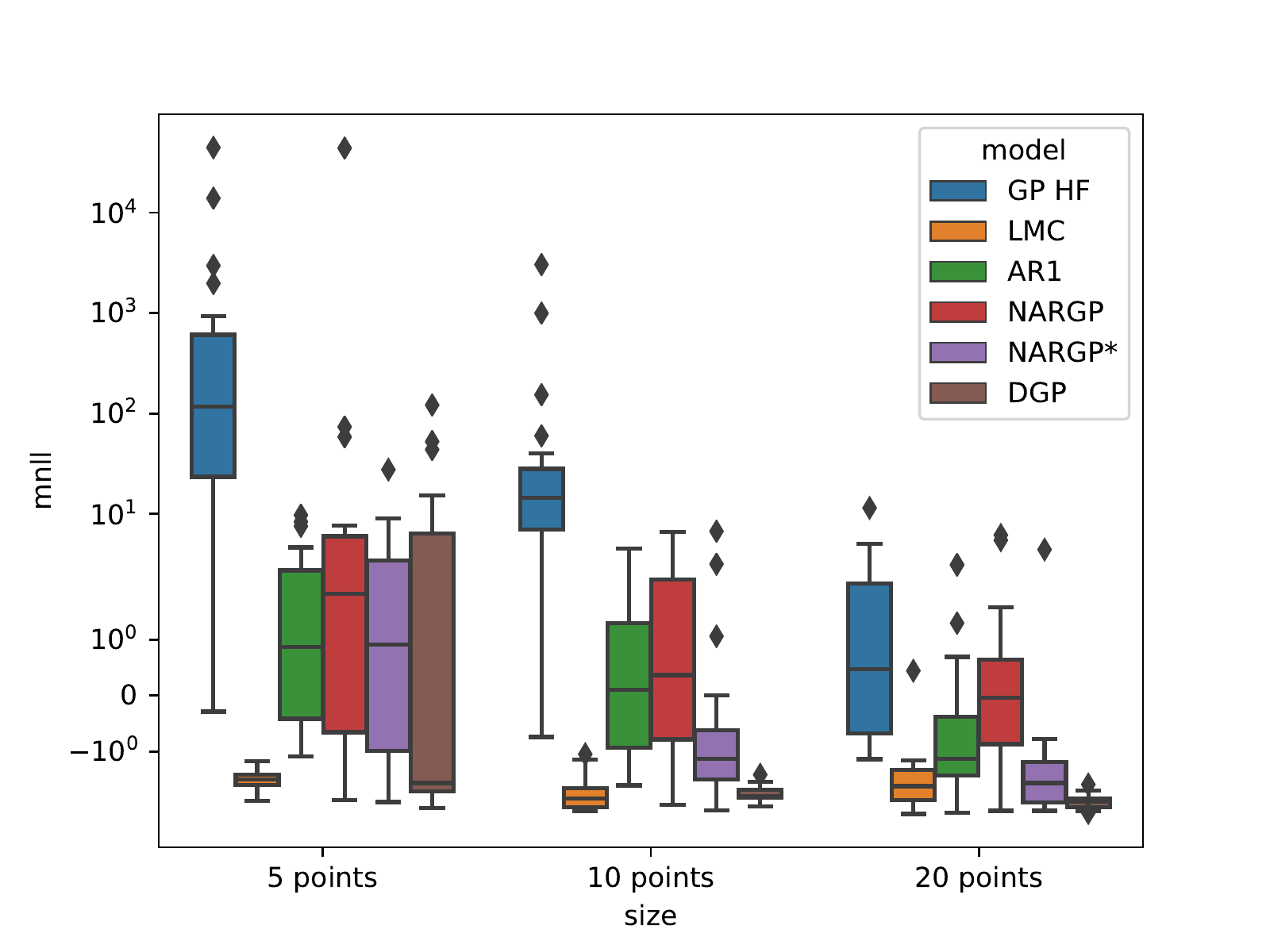}
\caption{Boxplots of RMSE and MNLL for SSBJ problem}
\label{SSBJ_boxplot}
\end{center}
\end{figure}

Boxplots illustrating the results for SSBJ test case are displayed in Figure \ref{SSBJ_boxplot}. Furthermore, numerical results including the comparison metrics are provided in Table \ref{tab:res_SSBJ}.
Similarly to the two previous test cases, linear approaches (AR1 and LMC) provide more accurate results considering limited HF sample size (for 5 points, R2 of $0.963$ for LMC compared to $0.679$ for DGP). However, by slightly increasing the number of HF samples from $5$ to $10$, the prediction accuracy of DGP becomes comparable to LMC and AR1 (RMSE improvements compared to GP HF in average of 74\% for LMC compared to 73\% for DGP). Furthermore, DGP presents a better likelihood of modeling the HF model through a better modeling of the prediction uncertainty (for 20 samples, MNLL of $-1.93$ for DGP compared to $-1.55$ for LMC).

\begin{table}[h!]
\scriptsize
\begin{center}
\caption{Summary of the results obtained on the SSBJ problem}
\label{tab:res_SSBJ}
\begin{tabular}{|c|c|c|c|c|c|c|}
\hline
\multirow{2}{*}{Function}&\multirow{2}{*}{Method}&\multirow{2}{*}{R2  (std)}&\multirow{2}{*}{RMSE (std)}&\multirow{2}{*}{MNLL (std)}&Evolution of&DOE size\\
&&&&&RMSE wrt GP HF& (LF, HF)\\
\hline
\multirow{18}{*}{SSBJ}&\multirow{3}{*}{GP HF}&0.131(0.336)&2.300e-1(4.591e-2)& 3.298e+3(9.908e+3)&-&100, 5 \\
&&0.48(0.329)&1.727e-1(5.192e-2)&2.223e+2(6.797e+2)&-&100, 10\\
&&0.836(0.074)&9.980e-2(2.000e-2)&1.404(2.947)&-&100, 20\\\cline{2-6}
&\multirow{3}{*}{LMC}&0.963(0.014)&4.753e-2(9.139e-3)&-1.514(1.762e-1)&$-79\%$&100, 5 \\
&&0.968(0.015)&4.367e-2(9.459e-3)&-1.762(3.060e-1)&$-74\%$&100, 10\\
&&0.980t0.005)&3.541e-2(4.494e-3)&-1.550(5.549e-1)&$-64\%$&100, 20\\\cline{2-6}
&\multirow{3}{*}{AR1}&0.957(0.024)&5.067e-2(1.334e-2)&1.953(3.185)&$-77\%$&100, 5 \\
&&0.970(0.008)&4.298e-2(5.946e-3)&3.726e-1(1.647)&$-75\%$&100, 10\\
&&0.980(0.006)&3.513e-2(4.680e-3)&-5.478e-1(1.467)&$-64\%$&100, 20\\\cline{2-6}
&\multirow{3}{*}{NARGP}&0.716(0.433)&1.093e-1(7.774e-2)&2.201e+3(9.555e+3)&$-52\%$&100, 5 \\
&&0.875(0.143)&7.791e-2(4.308e-2)&9.731e-1(2.269)&$-54\%$&100, 10\\
&&0.904(0.105)&7.003e-2(3.430e-2)&2.588e-1(2.109e+00)&$-29\%$&100, 20\\\cline{2-6}
&\multirow{3}{*}{NARGP*}&0.791(0.273)&9.819e-2(5.988e-2)&2.791(6.395)&$-57\%$&100, 5 \\
&&0.921(0.073)&6.499e-2(2.835e-2)&-4.889e-1(2.018)&$-62\%$&100, 10\\
&&0.950(0.039)&5.296e-2(1.911e-2)&-1.257(1.358)&$-47\%$&100, 20\\\cline{2-6}
&\multirow{3}{*}{DGP}&0.679(0.433)&1.110e-1(8.951e-2)&1.129e+1(2.936e+1)&$-51\%$&100, 5 \\
&&0.966(0.012)&4.569e-2(7.180e-3)&-1.750(1.303e-1)&$-73\%$&100, 10\\
&&0.974(0.012)&3.945e-2(8.025e-3)&-1.931(1.525e-1)&$-60\%$&100, 20\\\cline{2-6}
\hline
\end{tabular}
\end{center}
\end{table}

It is interesting to notice that LMC tends to perform as well as AR1 technique in terms of prediction accuracy but presents better results regarding the uncertainty model. The difference between the two approaches are in the symmetrical (LMC) and asymmetrical (AR1) fusion schemes. Multi-fidelity problems are asymmetrical by nature (information provided by HF are more accurate than by LF) so AR1 should be more suited for such a type of problems. However, it appears that LMC provides robustness to DoE and accurate predictions that are similar to AR1 or even better, but also provides an accurate uncertainty model for the prediction.

\newpage
\subsection{Aerostructural problem}
\subsubsection{Problem definition}

The aerostructural problem is based on OpenAeroStruct \cite{Jasa2018a} which is a tool that performs aerostructural simulation and optimization using OpenMDAO \cite{gray2019openmdao}. It couples a vortex-lattice method (VLM) \cite{VLM1991} and a finite-element method (FEM) using six degree-of-freedom spatial beam elements with axial, bending, and torsional stiffness to simulate aerodynamic and structural analyses using lifting surfaces \cite{Jasa2018a}. The aerodynamics submodel involves VLM to estimate the aerodynamic loads acting on the lifting surfaces. Considering a structured mesh defining a lifting surface, the aerodynamic properties are estimated using the circulation distribution. The lifting surface is modeled using horseshoe vortices to represent the vortex system of a wing. A vortex filament implies a flow field in the surrounding space. The strength of a vortex filament is its circulation, which induces lift on a surface.

For the structural submodel, a FEM technique is involved that uses spatial beam elements, resulting in six degree-of-freedom per node. The spatial beam element is a combination of beam, torsion and truss elements, therefore it simultaneously carries axial, bending, and torsional loads.

In OpenAeroStruct, the structures and aerodynamics are two separate submodels that receive inputs and compute outputs. The aerodynamics submodel takes as input a mesh and outputs aerodynamic loads, whereas the structural group takes as input aerodynamic loads and outputs structural displacements. The load and displacement exchange is simplified as the same spanwise discretization is used for the aerodynamic and structural submodels. A Gauss-Seidel algorithm \cite{salkuyeh2007generalized} is used to solve the multidisciplinary analyses and satisfy the interdisciplinary couplings.

\begin{figure}[!h]
\begin{center}
\includegraphics[width=0.8\linewidth]{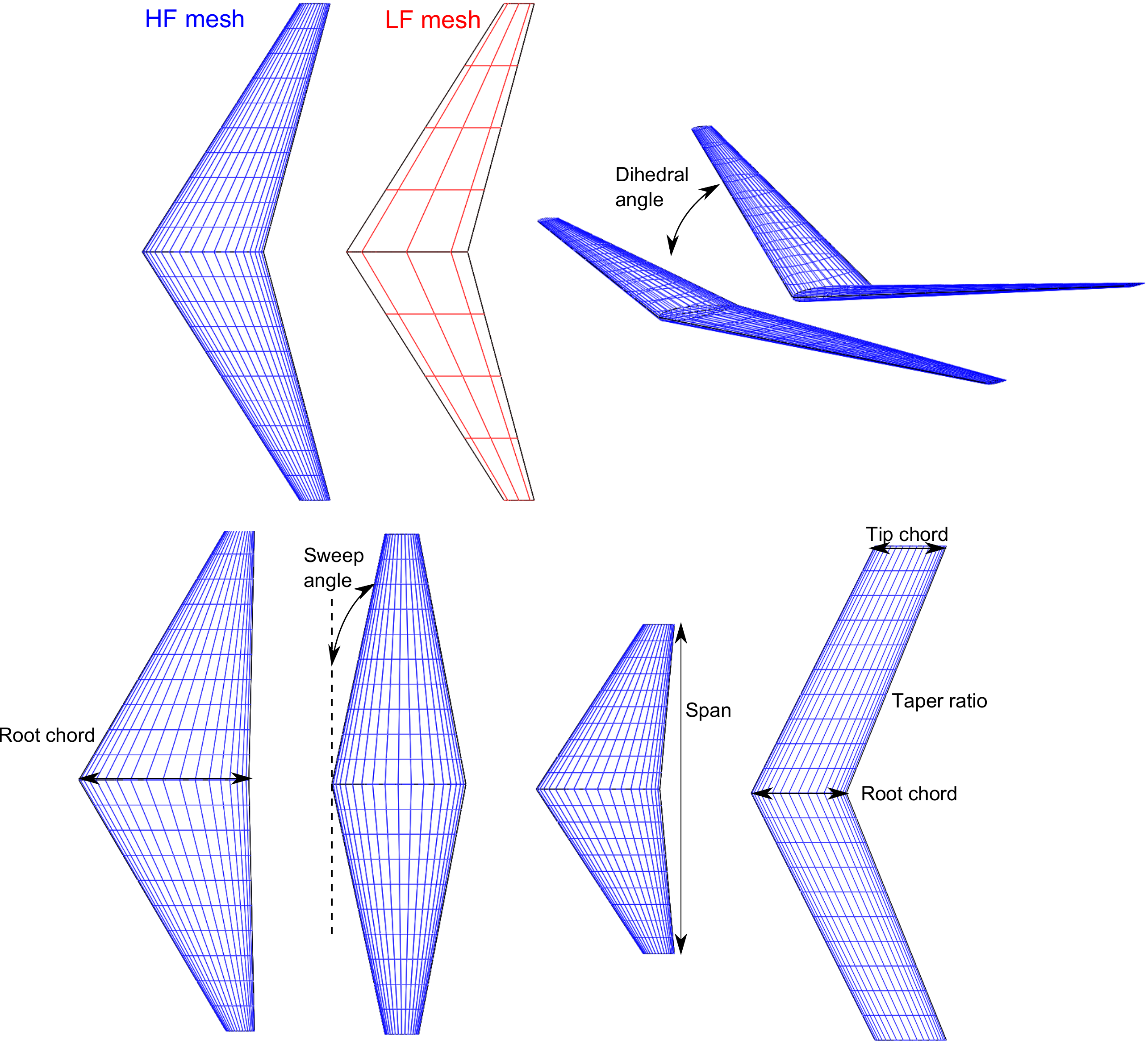}
\caption{Geometrical parameter definition and HF/LF meshes for the aerostructual problem}
\label{OAS_vb_def}
\end{center}
\end{figure}

For the multi-fidelity modeling problem, two fidelities are considered to estimate the lift coefficient CL of a wing. The difference between the models consists in the mesh refinement, a scarce mesh for the LF model and a dense mesh for the HF model (Figure \ref{OAS_vb_def}).

The input space is composed of eight design variables: the angle of attack, the span, the sweep angle, the dihedral angle, the taper ratio, and the root chord at three location along the space  (Table \ref{tab:OAS_input_design_vb}). The geometrical input parameters are illustrated in Figure \ref{OAS_vb_def}.

\begin{minipage}{\linewidth}
\centering
\captionof{table}{Aerostructural input design variable definition} \label{tab:OAS_input_design_vb} 
\begin{tabular}{|l|c|}
  \hline
  Input variables & Domain of definition  \\
  \hline
  Angle of attack & [1.0, 5.0]deg \\
  Span & [5.0, 10.0]m \\
  Sweep angle & [0., 20.]deg \\
  Dihedral angle & [0., 20.]deg \\
  Taper ratio & [0.7, 1.4] \\
  Root chord at three locations & [1.0, 5.0$]^3$m \\
  \hline
\end{tabular}
\end{minipage}

\subsubsection{Results}
\begin{figure}[!h]
\begin{center}
\includegraphics[width=.5\linewidth]{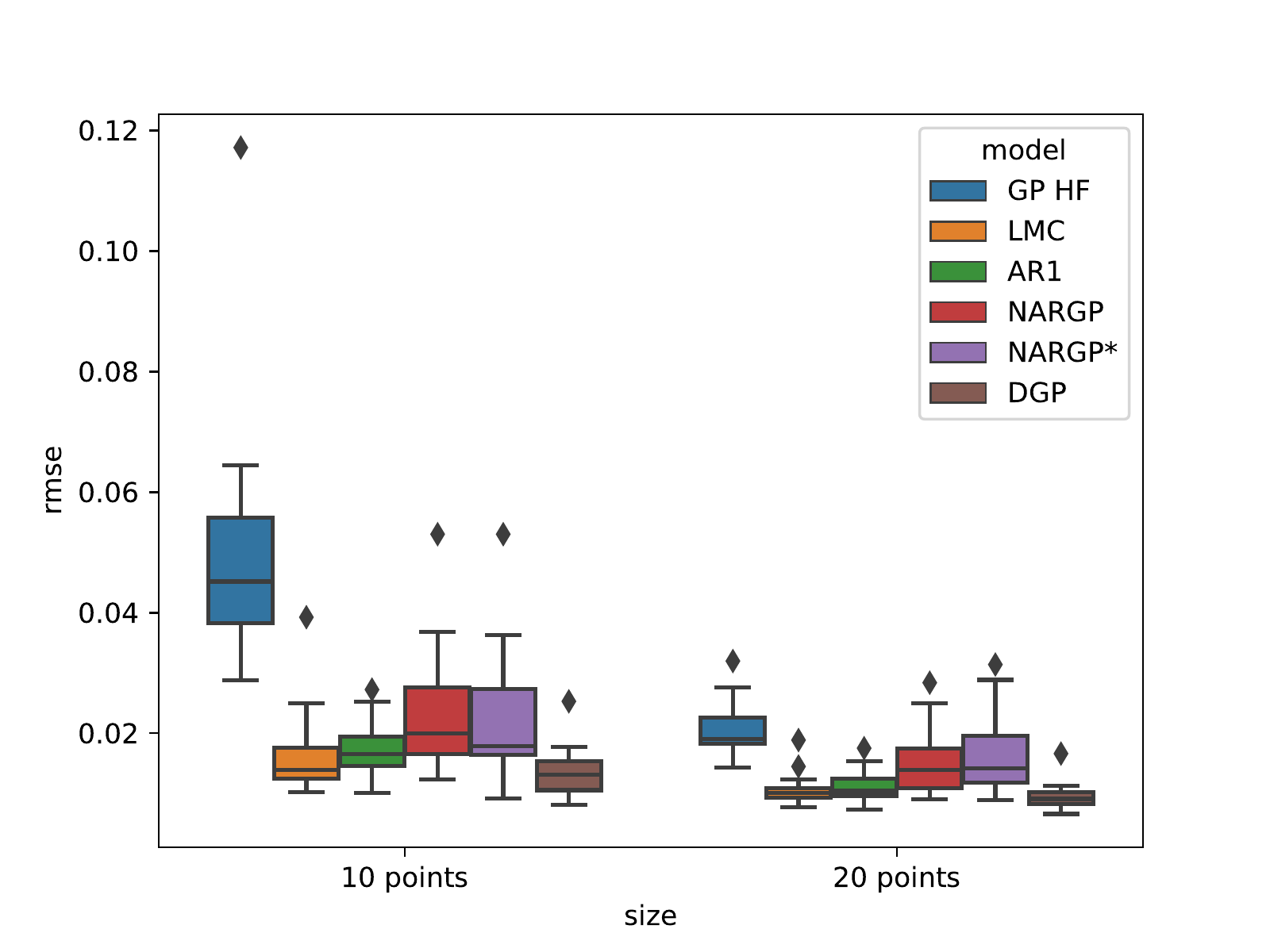}\includegraphics[width=.5\linewidth]{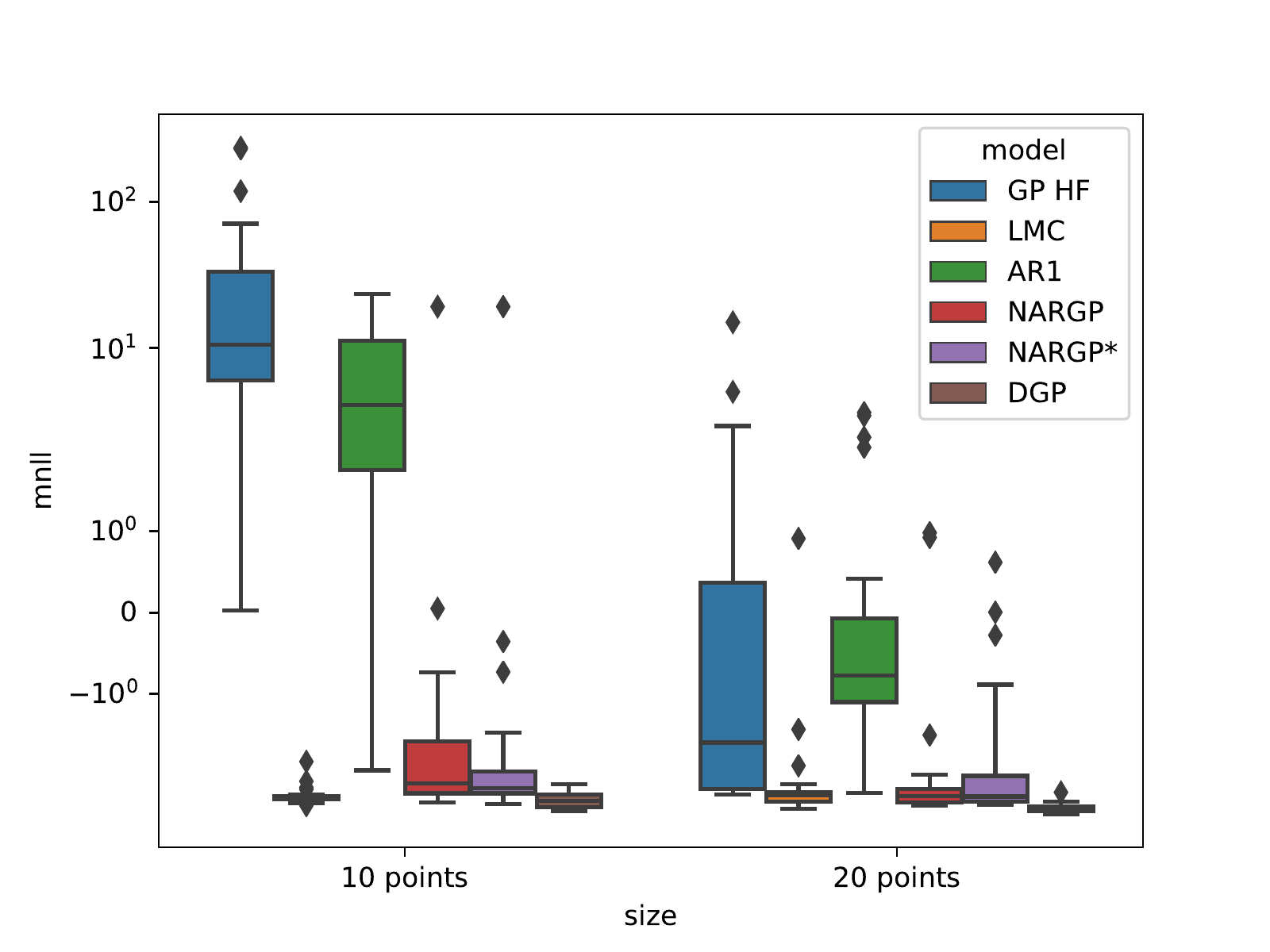}
\caption{Boxplots of RMSE and MNLL for OpenAeroStruct problem}
\label{OAS_boxplot}
\end{center}
\end{figure}

Boxplots illustrating the results for aerostructural test case are displayed in Figure \ref{OAS_boxplot}. Furthermore, numerical results including the comparison metrics are provided in Table \ref{tab:res_OAS}. In these experimentations, DGP outperforms the other multi-fidelity methods both in terms of prediction accuracy (improvement of RMSE compared to GP HF of 72\% for 10 HF samples and of 53\% for 20 HF samples compared to 67\% and 57\% for LMC respectively) and the likelihood of explaining the HF model (MNLL of $-2.99$ for DGP compared to $-2.77$ for LMC). For these two sizes of HF dataset, considering the linear multi-fidelity techniques, LMC provides better results than AR1 method, especially comparing MNLL (for 10 HF samples, $-2.77$ for LMC and $7.48$ for AR1).

\begin{table}[h!]
\scriptsize
\begin{center}
\caption{Summary of the results obtained on the OpenAeroStruct problem}
\begin{tabular}{|c|c|c|c|c|c|c|}
\hline
\multirow{2}{*}{Function}&\multirow{2}{*}{Method}&\multirow{2}{*}{R2  (std)}&\multirow{2}{*}{RMSE (std)}&\multirow{2}{*}{MNLL (std)}&Evolution of&DOE size\\
&&&&&RMSE wrt GP HF& (BF, HF)\\
\hline
\multirow{12}{*}{OAS}&\multirow{2}{*}{GP HF}&0.613(0.379)&4.891e-2(1.842e-2)&4.298e+1(6.880e+1)&-&160, 10 \\
&&0.939(0.028)&2.032e-2(4.372e-3)&-9.421e-3(4.046)&-&160, 20\\\cline{2-6}
&\multirow{2}{*}{LMC}&0.958(0.044)&1.609e-2(6.403e-3)&-2.774(3.113e-1)&$-67\%$&160, 10 \\
&&0.983(0.009)&1.059e-2(2.441e-3)&-2.575(9.259e-1)&$-47\%$&160, 20\\\cline{2-6}
&\multirow{2}{*}{AR1}&0.956(0.022)&1.721e-2(4.192e-3)&7.482(8.182)&$-64\%$&160, 10 \\
&&0.982(0.008)&1.102e-2(2.346e-3)&-2.040e-1(1.712)&$-45\%$&160, 20\\\cline{2-6}
&\multirow{2}{*}{NARGP}&0.914(0.085)&2.272e-2(9.700e-3)&-1.004(4.699)&$-53\%$&160, 10 \\
&&0.963(0.026)&1.534e-2(5.092e-3)&-2.379(1.190)&$-25\%$&160, 20\\\cline{2-6}
&\multirow{2}{*}{NARGP*}&0.921(0.084)&2.147e-2(9.738e-3)&-1.221(4.731)&$-56\%$&160, 10 \\
&&0.956(0.036)&1.635e-2(6.425e-3)&-2.258(1.149)&$-19\%$&160, 20\\\cline{2-6}
&\multirow{2}{*}{DGP}&0.973(0.018)&1.331e-2(3.934e-3)&-2.986(3.612e-1)&$-72\%$&160, 10 \\
&&0.987(0.007)&9.453e-3(2.008e-3)&-3.354(2.363e-1)&$-53\%$&160, 20\\\cline{2-6}
\hline
\end{tabular}
\label{tab:res_OAS}
\end{center}
\end{table}

Moreover, considering NARGP and NARGP*, they seem to be dominated by DGP due to its coupled architecture, there is a clear difference between NARGP and DGP in terms of RMSE improvements (25\% for NARGP compared to 53\% for DGP for 20 HF samples). In addition, nested DoE is not a guarantee for NARGP to perform better. For instance, for 20 samples in HF, non nested NARGP presents more accurate results than nested NARGP* (R2 of $0.963$ compared to $0.956$).

\section{Results synthesis and concluding remarks}
The results presented in this benchmark of analytical and aerospace design problems are a direct illustration of the "no free lunch theorem" \cite{ho2002simple}, no multi-fidelity method performs better for all analytical and engineering problems, for all HF sample size and for all repetitions. The purpose of the paper was not to provide a single method for such a type of modeling problem but to highlight the interests of alternative GP-based multi-fidelity techniques to the classical AR1 method that is often used in aerospace multi-fidelity modeling problems. 
Four multi-fidelity GP-based alternative approaches (AR1, co-kriging with LMC, NARGP and MF-DGP) have been briefly presented outlining the difference in terms of fusion scheme (symmetrical or asymmetrical) and in terms of fidelity relationship (linear or non-linear). These techniques have been evaluated over different analytical test cases and with four aerospace design problems (structure, trajectory optimal control, multidisciplinary simulation of a business jet and aerostructural wing problem). 

Some general tendencies can be drawn about the multi-fidelity GP-based alternatives studied in this paper. When a limited number of HF samples is available due to the computational cost associated to such models, multi-fidelity techniques allow to reduce the prediction error compared to a single high-fidelity GP model. When the number of available HF samples augments, the relative improvement of multi-fidelity methods compared to single fidelity approach decreases up to a point where low-fidelity information do not offer improvement to the prediction accuracy and therefore a single fidelity model is sufficient.  

Moreover, when a very limited number of HF samples with respect to the problem dimension is considered, linear mapping between fidelities (AR1 and LMC) tends to provide better results than non-linear mapping approaches (NARGP and DGP) which are more difficult to train when not enough HF information is available to model this relationship. Indeed, the non-linear multi-fidelity techniques, due to their higher complexity of definition (nested composition of Gaussian processes), offer higher capability of modeling but with a higher number of hyperparameters to be tuned. Therefore, to fit complex relationship between the fidelities a higher number of high-fidelity samples is required. Among the presented GP-based alternatives, DGP presents the highest computational cost in terms of inference due to the nested GP structure that requires an adequate training process involving few hundreds or even thousands of hyperparameters compared to few dozen for AR1, LMC and NARGP (depending on the problem dimension).
Even if theoretically, NARGP should be trained over nested dataset (which is not always possible) the difference in prediction performance along the different benchmark problems is not significant between nested and non-nested NARGP. DGP tends to be the best non-linear method over all the test cases, especially in terms of likelihood of explaining the HF model (MNLL), due to a better description of the uncertainty associated to the prediction. 

The difference between symmetrical and asymmetrical information fusion scheme (AR1 and co-kriging with LMC) for multi-fidelity modeling problem is not as strong as one could have expected. Indeed, even if, mutli-fidelity modeling problem is an asymmetrical information scheme fusion type of problem (HF information is more valuable than LF information), through the different results of the benchmark problems, AR1 does not outperform co-kriging with LMC. Once again, between the two methods, none performs better for all the analytical and engineering problems.  

One key aspect discussed in this paper is the influence of the relationship between LF and HF models. The linearity of the  relationship along with the degree of correlation between the low and the high-fidelity models has an important impact. When it is highly non-linear or weakly correlated, AR1 and LMC tends to be limited in their capabilities to catch such behavior as illustrated by the 1D analytical problem, therefore non-linear techniques should be prefer for such problems. For complex and computationally intensive models, it is difficult to know in advance the type of relationship between the fidelity models, however, as illustrated through the different benchmark problems, it is worthwhile to try different alternative multi-fidelity modeling techniques as the training time compared to the evaluation of the high-fidelity model is often negligible.  

Another important trend is linked to the curse of dimensionality. The improvement resulting by adding HF samples in the dataset is more sensitive for low dimensional test case than for high dimension test cases. Therefore, to keep the same level of accuracy when the dimension increases, a simple rule proportional to the dimension for the number of HF sample is not sufficient. It should take into account the fact that the volume of the design space increases so fast that the LF and HF data become sparse for high dimensional problems. By increasing the dimension, linear multi-fidelity techniques tend to perform better in terms of prediction accuracy as illustrated with the dimension variation problem. Even with limited HF samples, they are able to catch the general behavior of the HF model. 

Eventually, a trade-off between prediction accuracy and the quality of the model of prediction uncertainty is often required. Indeed, for GP-based surrogate models, the model of prediction uncertainty is exploited in various contexts (optimization, uncertainty propagation and refinement strategy) and therefore it is also an important component along with the prediction accuracy. In some applications, it might be interesting to have a slightly less accurate prediction but a considerably improved uncertainty prediction model, therefore this component has to be taken into account in the trade-off.

\section*{Acknowledgments}

This work is funded by the ONERA (Office National d'Etudes et de Recherches Aérospatiales - The French Aerospace Lab) project MUFIN (multidisciplinary and multifidelity under uncertainty for the study of new aerospace concepts, 2019-2021). The PhD thesis of A. Hebbal is funded by ONERA and the University of Lille. 
\bibliography{test}

\end{document}